\documentclass[11pt]{article} 
\usepackage[a4paper, margin=1.6cm]{geometry}

\usepackage{amssymb,amsmath,amsthm}
\usepackage{bm}

\usepackage{natbib}
\usepackage[colorlinks,linkcolor=black,citecolor=blue,urlcolor=blue]{hyperref}
\usepackage[dvips]{graphics}
\usepackage{graphicx}
\usepackage{subcaption}
\usepackage{fontenc}
\usepackage{pifont}
\usepackage{thmtools,thm-restate}

\usepackage[utf8]{inputenc}
\usepackage[colorlinks,linkcolor=black,citecolor=blue,urlcolor=blue]{hyperref}
\usepackage{graphicx}
\usepackage{todonotes}
\usepackage{authblk}

\usepackage{appendix}
\usepackage{chngcntr}
\usepackage{etoolbox}
\usepackage{lipsum}

\newtheorem{theorem}{Theorem}
\newtheorem{corollary}{Corollary}
\newtheorem{lemma}{Lemma}
\newtheorem{proposition}{Proposition}

\theoremstyle{definition}

\newtheorem{assumption}{Assumption}

\newtheorem{remark}{Remark}

\definecolor{darkblue}{rgb}{0., 0., 0.}
\newenvironment{darkblue}{\par\color{darkblue}}{\par}

\DeclareMathOperator*{\argmin}{argmin}

\newcommand{\Beginproof}[1]{\begin{proof}{(Proof{#1}).}}
\newcommand{\Endproof}{\end{proof}}

\title{Exponential Concentration in Stochastic Approximation }
\author[1]{Kody J.H. Law}
\author[2]{Neil Walton} 
\author[1]{Shangda Yang}
\affil[1]{University of Manchester}
\affil[2]{Durham University}

\begin{document}

\maketitle

\abstract{
We analyze the behavior of stochastic approximation algorithms where iterates, in expectation, progress towards an objective at each step. 
When progress is proportional to the step size of the algorithm, we prove exponential concentration bounds. These tail-bounds contrast asymptotic normality results, which are more frequently associated with stochastic approximation.
The methods that we develop rely on a geometric ergodicity proof. This extends a result on Markov chains due to Hajek (1982) to the area of stochastic approximation algorithms. We apply our results to several different 
Stochastic Approximation algorithms, specifically 
Projected Stochastic Gradient Descent, Kiefer-Wolfowitz and Stochastic Frank-Wolfe algorithms.
When applicable, 
our results prove faster $O(1/t)$ and linear convergence rates for Projected Stochastic Gradient Descent with a non-vanishing gradient.
}

\section{Introduction}
\begin{darkblue}
We consider stochastic approximation algorithms where the expected progress toward the optimum is proportional to the algorithm's step size. 
For instance, a stochastic gradient descent algorithm applied to a convex function will satisfy this property when bounded away from the optimum. However, this property can continue to hold as an algorithm approaches the optimum. 
   As we will discuss, this is true when the convex objective function is \emph{sharp}. Stated informally, these objectives have a {\fontfamily{phv}\selectfont V}-shape at the optimum rather than a quadratic {\fontfamily{phv}\selectfont U}-shape. The latter case is extensively studied. The asymptotic error has a normal distribution; see \cite{Chung,fabian1968asymptotic}. However, in the former case, little is known about the limit distribution of the error. 
In these settings, we will show that the error for algorithms such as Projected Stochastic Gradient Descent, Kiefer-Wolfowitz, and Frank-Wolfe have an exponential concentration and a faster rate of convergence than would be anticipated by standard results for stochastic optimization with a smooth objective. 
We develop methods which are typically used in probability to analyze random walks or in applied probability to analyze queueing networks. For stochastic approximation, our results establish new exponential concentration bounds.  
\end{darkblue}

We now summarize the background and problems where our results apply. 
\smallskip

\noindent \textit{Stochastic Gradient Descent: Standard Asymptotic Results.} Due to its applicability in machine learning, there is now a vast literature on stochastic gradient descent \citep{bottou2018optimization}.
The rate of convergence found to the optimal point for a (projected) stochastic gradient descent procedure on a convex objective has order $O({1}/{\sqrt{t}})$ of the optimum after $t$-iterations of the algorithm  \citep{nemirovski2009robust,moulines2011non,bottou2018optimization}. 
In the paper, we find conditions under which the improved $O(1/t)$ convergence rate holds, and developing on the work of \cite{davis2019stochastic}, we also find linear convergence results. 
Our results apply to optimization problems where the gradient does not vanish as we approach the optimum.
A critical feature of our analysis is an exponential concentration bound.
\smallskip

\noindent \textit{Asymptotic Normality, Exponential Bounds, and Reflected Random Walks.}
For stochastic approximation, the normal distribution has long been known to characterize the limiting behavior of a stochastic approximation procedure. See
\cite{fabian1968asymptotic} and Chapter 10 of
\cite{kushner2003stochastic}. 
{\darkblue
Such theories are statistically efficient for smooth optimization problems with and without constraints. See \cite{duchi2021asymptotic}, \cite{davis2023asymptotic} and \cite{moulines2011non} respectively, and results are motivated by the asymptotic normality results for maximum likelihood estimators (MLE) of \cite{le1953some} and \cite{hajek1972local}.
However, one should note that such asymptotics may not always lead to a Gaussian limit. 
For example, the MLE of a uniform distribution is not asymptotically normal but is instead exponentially distributed. (See Section \ref{sec:uniformMLE} of the E-companion for a proof.) } The stochastic optimization algorithms considered in the paper are settings where the normal distribution is not asymptotically optimal. 

While asymptotic normality has a long history, the exponential bounds found here are not well-understood and do not appear in stochastic approximation literature.
We argue that when the objective's gradient is non-vanishing as we approach the optimum, the normal approximation will not hold, and an exponential concentration bound is more appropriate. 

To construct exponential concentration bounds. We establish these bounds using a geometric Lyapunov bound. These arguments are commonly employed to establish the exponential ergodicity of Markov chains. See Kendall's Renewal Theorem in Chapter 15 of \cite{meyn2012markov}.  \cite{hajek1982hitting}, in particular, provides a proof that converts a drift condition into an exponential Martingale that establishes fast convergence rates for ergodic Markov chains. A key contribution of this paper is to extend this argument to stochastic approximation.

These bounds are typically applied to queueing networks  \citep{kingman1964martingale,bertsimas2001performance} because many queueing processes are random walks with constraints and non-zero drift. These conditions lead to exponential distribution bounds \citep{harrison1987multidimensional}.
 \cite{kushner2003stochastic} discusses these connections when analyzing the diffusion approximation of stochastic approximation procedures with constraints. Nonetheless, as we will discuss, diffusion analysis does not fully recover the required exponential concentration. 
The concentration results proven here are, to the best of our knowledge, new in the context of stochastic approximation.

\noindent \textit{Constrained Stochastic Gradient Descent, Sharp Functions, and Geometric Convergence.}
Our results are applicable when the gradient of the function does not vanish.
In particular, our results can be applied to constrained stochastic approximation when the optimum lies on the boundary.
The text \cite{kushner2012stochastic} analyses the convergence of stochastic approximation algorithms on constrained regions.
\cite{buche2002rate} prove convergence rates. These authors observe that analysis typically applied to analyze unconstrained stochastic approximation does not readily apply to the constrained case. 

\begin{darkblue}
Boundary constraints are not a requirement of our analysis. Our results apply under a non-vanishing gradient condition. This closely relates to the property of a function being \emph{sharp}.  \cite{davis2019stochastic} presents a variety of machine learning tasks for which the objective is sharp. We show that our non-vanishing gradient condition is equivalent to sharpness for convex functions.
Our exponential concentration bounds are tighter than Gaussian concentration bounds. Applying this concentration bound to the work of \cite{davis2019stochastic}
leads to an improved linear convergence rate for projected stochastic gradient descent.   
    Recent work \cite{davis2023asymptotic} analyses the asymptotic normality of stochastic gradient descent algorithms, which exhibit sharpness away from a smooth manifold around the optimum. 
\end{darkblue}

\begin{darkblue}
    \noindent \textit{Further Stochastic Approximation Algorithms.}
    The main result of the paper considers a generic stochastic algorithm with non-vanishing drift and sub-exponential noise (Conditions \eqref{fcond:1} and \eqref{fcond:2}). For this reason, our results hold for other mainstream stochastic approximation algorithms. We consider the Kiefer-Wolfowitz and the Frank-Wolfe (or conditional gradient algorithm) as examples. See \cite{kiefer1952stochastic} and \cite{frank1956algorithm}.

    Kiefer-Wolfowitz is the primary alternative to Robbins and Monro's stochastic gradient descent algorithm. Here, gradient estimates are replaced by a finite difference approximation. We prove that exponential concentration and linear convergence hold for Kiefer-Wolfowitz under a non-vanishing drift condition. 

 Frank-Wolfe is a popular projection-free alternative to projected gradient descent algorithms. See \cite{jaggi2013revisiting,hazan2012projection}. The stochastic Frank-Wolfe algorithm is proposed and analyzed in \cite{hazan2016variance}. We provide conditions analogous to sharpness that ensure exponential concentration for the stochastic Frank-Wolfe algorithm. Linear convergence analogous to \cite{davis2019stochastic} can also occur for these algorithms. 
\end{darkblue}

\medskip

\noindent The results as a whole establish a sequence of connections between stochastic modeling bounds used in queueing and stochastic approximation. 
These exponential tail bounds differ from Gaussian concentration bounds typically analyzed in unconstrained stochastic approximation. Moreover, these results lead to faster convergence rates than standard stochastic approximation results. 

\subsection{Organization.}

This article is structured as follows.
Section \ref{sec:Notation} gives initial notation. (Further notation will be introduced as we present each of our results.) 
Section \ref{sec:Main} presents the paper's main results. 
Section \ref{sec:Informal} provides 
intuition on exponential concentration. 
Section \ref{Lya} presents a generic Lyapunov function result for exponential concentration.
Section \ref{sec:PSGDresult} applies our results to Projected Stochastic Gradient Descent (PSGD). 
In Section \ref{sec:Counter}, we show that, although exponential concentration holds, the exponential distribution is not, in general, the limiting distribution when gradients do not vanish.
In Section \ref{sec:KWAlgorithm}, we provide an exponential concentration bound for the Kiefer-Wolfowitz stochastic approximation algorithm.
In Section \ref{sec:SFW}, we give an exponential concentration bound for the Stochastic Frank-Wolfe algorithm.
A linear convergence result for PSGD is presented in Section \ref{sec:Linear}. 
Proofs for the results are given in Section \ref{sec:Proofs}, with later results deferred to the E-companion. 
Numerical experiments are presented in Section \ref{sec:LP}. 

\begin{darkblue}
\section{Problem setting and initial assumptions}\label{sec:Notation}

We give some basic notation and assumptions which hold throughout this paper. The algorithms and results considered will require some specific assumptions. These will presented in the sections relevant to those results.
\end{darkblue}

\noindent \textit{Basic Notation.} We apply the convention that $\mathbb Z_+=\{ n : n =0,1,2,... \}$ and $\mathbb R_+ = \{ x : x\geq 0\}$. Implied multiplication has precedence over division, that is, $2a/3bc = (2\times a)/(3\times b \times c)$. 
\smallskip

\noindent \textit{Optimization Notation.}  We let $\mathcal X$ denote a nonempty closed bounded convex subset of $\mathbb R^d$. For a continuous function $f: \mathcal X \rightarrow \mathbb R$, we consider the minimization 
\begin{equation}\label{Obj}
\min_{\bm x\in\mathcal X}\, f(\bm x) \, .
\end{equation}
We let $\mathcal X^\star$ be the set of minimizers of the above optimization {problem}. We let $\Pi_{\mathcal X}(\bm x)$ denote the projection of $x$ onto the set $\mathcal X$. That is $\Pi_{\mathcal X}(\bm{x}) := \argmin_{\bm{y} \in \mathcal X} ||\bm{x}-\bm{y} ||^2$, where $|| \cdot ||$ denotes the Euclidean norm. We let $d(\bm{x},\mathcal X)$ denote the distance from a point to its projection. That is $d(\bm{x},\mathcal X):= \min_{\bm{y} \in \mathcal X} ||\bm{x}-\bm{y} ||$. We let $F$ be the gap between the maximum and minimum of $f(\bm x)$ on $\mathcal X$, that is
\[
F := \max_{\bm x\in\mathcal X} f(\bm x) - \min_{\bm x\in\mathcal X} f(\bm x)\, .
\]
\noindent \textit{Stochastic Iterations.} 
We consider a generic stochastic iterative procedure for solving the optimization {problem} \eqref{Obj}. 
Consider a random sequence $\{ \bm{x}_{t} \}_{t=0}^\infty$ adapted to a filtration $\{ \mathcal F_t \}_{t=0}^\infty$ with $\bm x_t \in \mathcal X$ for each $t\in\mathbb Z_+$. The sequence $\{\alpha_t\}_{t=0}^\infty$ determines the distance between successive terms. We define
$\bm c_t := {(\bm x_t - \bm x_{t+1}) }/{ \alpha_t}$
 \text{and thus }   
 \begin{equation}\label{eq:SA}
 \bm x_{t+1} = \bm x_t - \alpha_t \bm c_t\, .
\end{equation}

\section{Main Results}\label{sec:Main}

In this section, we present our main results, as well as intuition and counter-examples.

\subsection{Informal description of the main result} 
\label{sec:Informal}
The normal distribution is typically associated with the dispersion of a random walk. However, when a random walk is constrained, the exponential distribution is the limiting distribution.
So, while the normal approximation is applied in the analysis of smooth stochastic approximation algorithms, for non-smooth problems and for constrained problems, the exponential approximation is correct.

\begin{figure}[ht!]
\centering
\begin{subfigure}{.5\textwidth}
  \centering
  \includegraphics[width=1.\linewidth]{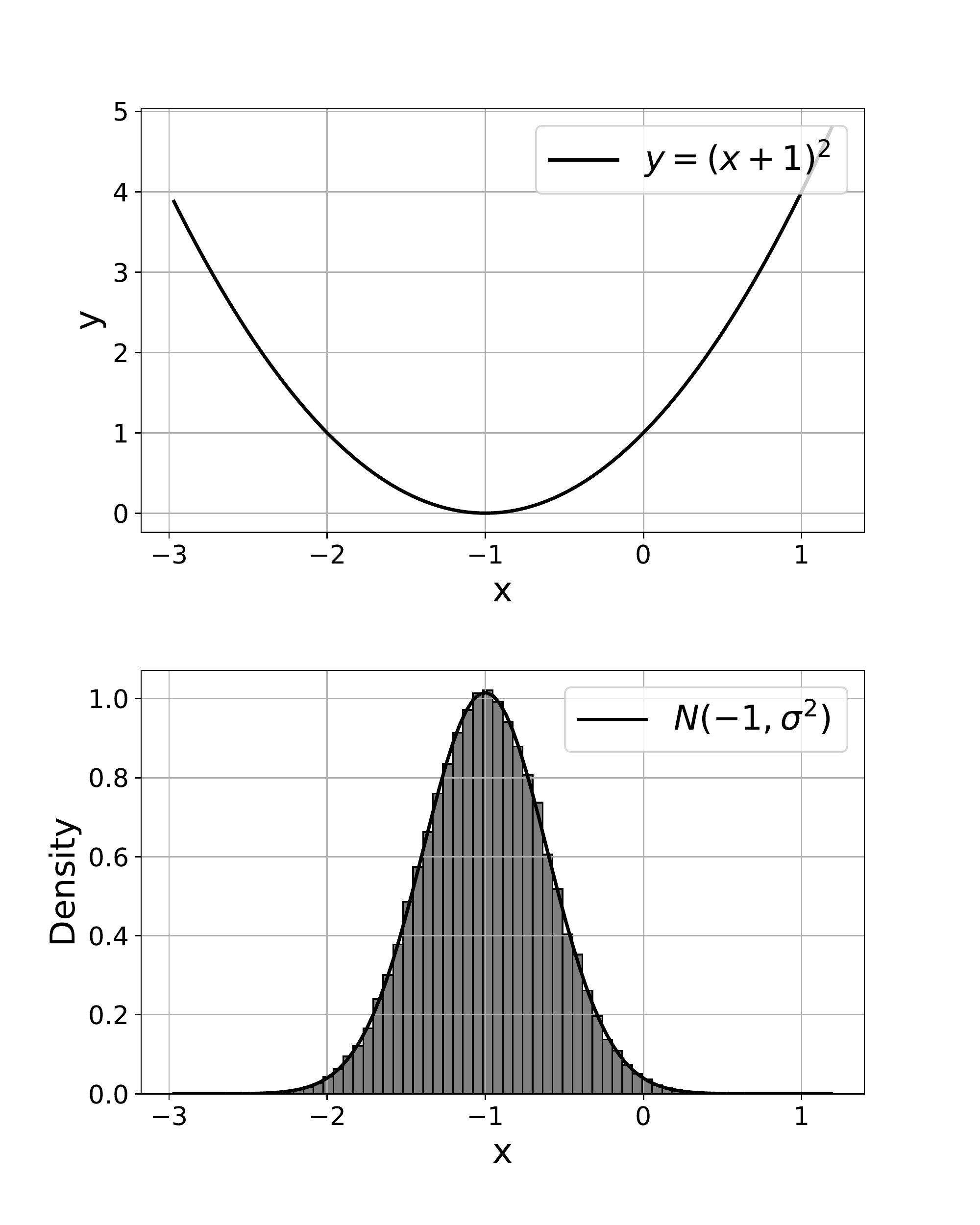}
  \caption{Unconstrained Stochastic Gradient Descent}
  \label{fig:sub1}
\end{subfigure}%
\begin{subfigure}{.5\textwidth}
  \centering
  \includegraphics[width=1.\linewidth]{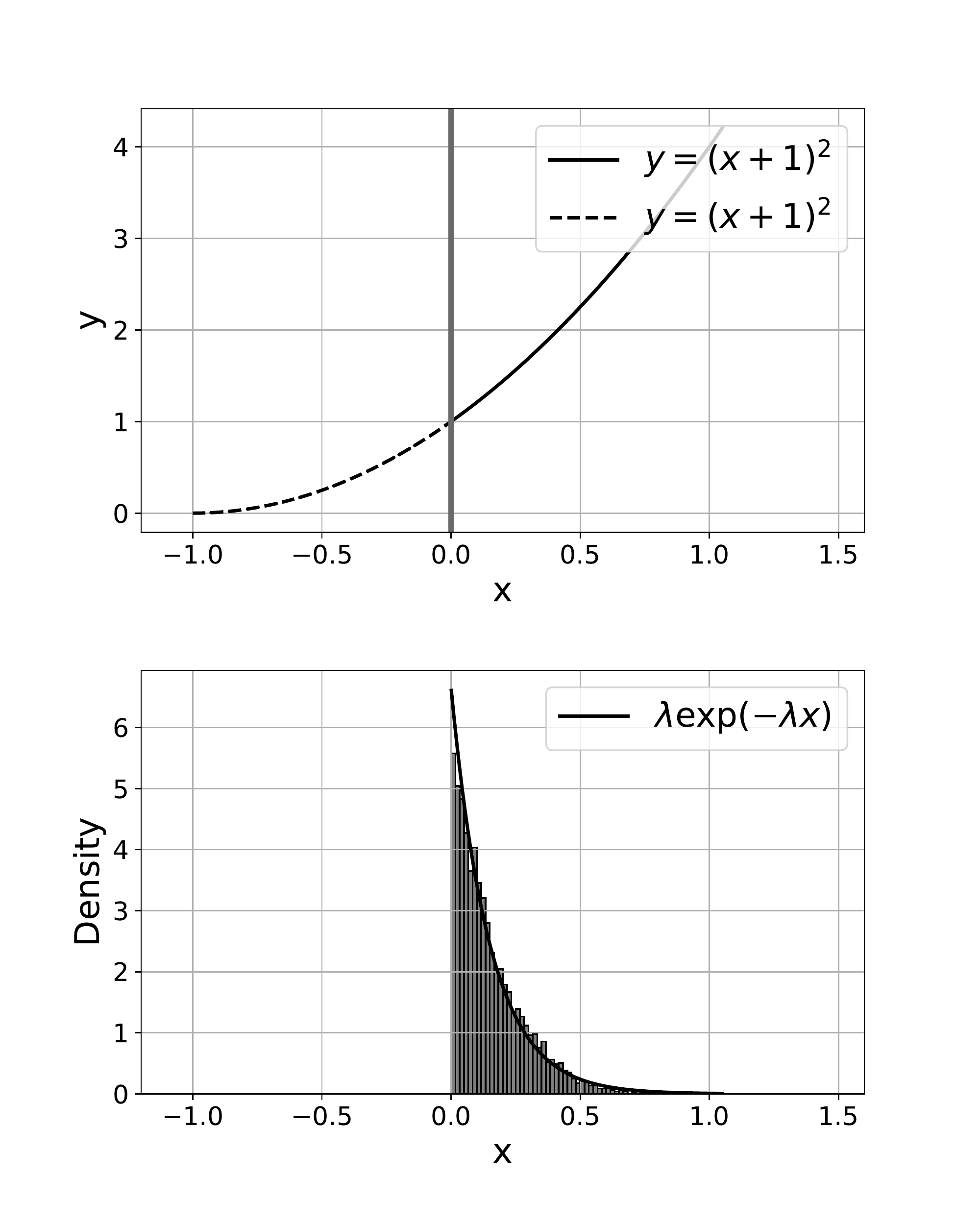}
  \caption{Constrained Stochastic Gradient Descent}
  \label{fig:sub2}
\end{subfigure}
\
\caption{\label{fig:sub}The above plots a simulation of a stochastic gradient descent algorithm with a constant step size on the function $f(x) = (x+1)^2$. Figure \ref{fig:sub1}: When the objective is unconstrained the density of the location of iterates is well approximated by a normal distribution with variance $\sigma^2 = O(\alpha)$, where $\alpha$ is the step size of the algorithm. The distance to the optimum is $O(\alpha^{1/2})$. Figure \ref{fig:sub2}: When the value of $x$ is constrained to the positive orthant the gradient no longer vanishes. The distribution of iterates away from zero now has an exponential decay with rate $\lambda = O(\alpha^{-1})$. So for step size, $\alpha$, the distance to the optimum is $O(\alpha)$. This paper proves that exponential concentration holds more generally for stochastic approximation procedures with non-vanishing gradients. \label{fig:test}}
\end{figure}
With reference to Figure \ref{fig:test}, the high-level intuition for this behavior in a stochastic gradient algorithm is as follows.  
Consider a projected stochastic gradient descent algorithm with a small but fixed learning rate.
When the optimum is in the interior of the constraint set and the objective is smooth, the algorithm's progress will slow as the iterates approach the minimizer in a manner that is roughly proportional to the distance to the optimum. 
In this regime, the process is well approximated by an Ornstein-Uhlenbeck (OU) process, for instance, see Chapter 10 of \cite{kushner2003stochastic}.  
An OU process is known to have a normal distribution as its limiting stationary distribution. 
This stationary distribution determines the rate of convergence to the optimum; see \cite{chen2022stationary}.
If we consider the same iterates, but instead, these are now projected to belong to a constraint set, then the gradient of iterates need not approach zero as we approach the optimum see Figure \ref{fig:test}b). The resulting process behaves in a manner that is approximated by a reflected Brownian motion. 
When the gradient is non-zero on the boundary, it is well known that a reflected Brownian motion with negative drift has an exponential distribution as its stationary distribution, see \cite{harrison1987multidimensional}. We seek to establish bounds that exhibit this exponential, stationary behavior while allowing for time-dependent step sizes. 
This provides intuition for the exponential concentration results found in this paper.

To summarize, \emph{strongly} convex functions are approximately quadratic around their optimum; this leads to the normal approximation. Think of such functions being {\fontfamily{phv}\selectfont U}-shaped. However, for an exponential approximation, the function is {\fontfamily{phv}\selectfont V}-shaped at the optimum. 
Such functions are \emph{sharp} convex functions. (See Remark \ref{rmk:Sharp} for a discussion on sharpness.) 
Here, the gradient does not approach zero as we approach the optimum. We show that when a convex function is sharp, a stochastic gradient descent algorithm has an exponential concentration around its optimum.

\subsection{An Exponential Lyapunov Bound} \label{Lya}

We consider a general stochastic optimization algorithm and show that exponential concentration holds 
when the expected progress towards its objective is proportional to the learning rate $\alpha_t$. 
 This leads to the following condition.

\begin{assumption}[Drift Condition] The sequence $\{ \bm{x}_t \}_{t=0}^\infty$ satisfies
\begin{align}
\mathbb E [ f(\bm{x}_{t+1}) - f(\bm{x}_t) |\mathcal F_t ] \leq - 2 \alpha_t \kappa \label{fcond:1} \tag{C1}
\end{align}
whenever $f(\bm{x}_t) -f(\bm{x}^\star) \geq \alpha_t B$ for some $\kappa>0$ and some $B>0$.     
\end{assumption}

We also assume that noise is sub-exponential.

\begin{darkblue}
\begin{assumption}[{Moment Condition}] 
There exists a constant $\lambda >0$ and a random variable $Y$ such that 
\begin{align}
\big[ | f(\bm x_{t+1}) - f(\bm x_t) | \big|\mathcal F_t \big] \leq \alpha_t Y  \quad \text{and} \quad \mathbb E [ e^{\lambda Y } ] < \infty \, .\label{fcond:2} \tag{C2}
\end{align}   
\end{assumption}
\end{darkblue}

Condition \eqref{fcond:1} states that the stochastic iterates will make progress against its objective when away from the optimum. 
The Condition \eqref{fcond:2} is a mild noise condition. For example, if $f(X)$ is Lipschitz continuous, then it is sufficient that $||\bm c_t ||$ has a sub-exponential tail. (See Lemma \ref{lem:fcond2} in E-companion \ref{sec:subExp} for verification of this claim.)
Shortly we will establish the Conditions \eqref{fcond:1}
and \eqref{fcond:2} when applying projected stochastic gradient descent. However, for now, we leave \eqref{fcond:1} and \eqref{fcond:2} as general conditions that can be satisfied by a stochastic approximation algorithm. 

The main result of this section is as follows.
\begin{restatable}{theorem}{Corone}
\label{Cor:1}
For learning rates of the form 
$
\alpha_t = {a}/{(u+t)^\gamma}
$
with $a,u>0$ and  $\gamma \in [0,1]$, if Conditions \eqref{fcond:1} and \eqref{fcond:2} are satisfied by a stochastic approximation algorithm, then
\begin{align}\label{Cor1:eq1}
\mathbb P( f(\bm x_{t+1}) - f(\bm x^\star) \geq z ) 
\leq
I {e^{- \frac{J}{\alpha_t} z }} \, 
\end{align}
and
\begin{align}\label{Cor1:eq2}
\mathbb E [ f(\bm x_{t+1})  - f(\bm x^\star) ] \leq  {K} \alpha_t\, 
\end{align}
for time independent constants $I$, $J$ and $K$. 
\end{restatable}

What we see in these results is that once the dependence on the initial state of the system has been accounted for then the process $f(\bm x_t)$ has an exponential concentration and will be within a factor of $\alpha_t$ of the optimum.  
This is proved in a more general form in Proposition \ref{Thrm1} in Section \ref{sec:ProofThrm1}.

It is worth remarking that Theorem \ref{Cor:1} (and Proposition \ref{Thrm1}) hold for any algorithm for which the generic Conditions \eqref{fcond:1} and \eqref{fcond:2} hold. Thus, the results are not intended to apply to any particular stochastic optimization, nor do we place specific design restrictions on the algorithm. 
The result emphasizes that a convergence rate may differ depending on the geometry of the problem at hand, and this convergence may well be faster than anticipated. 

\subsection{Projected Stochastic Gradient Descent}
\label{sec:PSGDresult}
In Theorem  \ref{Cor:1}, we did not specify the stochastic approximation procedure used nor did we explore settings where Conditions \eqref{fcond:1} and \eqref{fcond:2} hold. This section provides a standard setting where our results apply.
We consider projected stochastic gradient descent on the Lipschitz continuous function $l : \mathcal \mathcal X \rightarrow \mathbb R$. That is we wish to solve the optimization problem:
	\begin{equation}
\text{minimize} \qquad l(\bm x) \qquad \text{over} \quad \bm x \in \mathcal X \, .
\label{eq:l_obj}	
\end{equation}
We analyze the Project Stochastic Gradient Descent (PSGD) algorithm:
\begin{subequations} \label{eq:PSGD}
\begin{align}
{\bm y}_{t+1} &=\bm x_t - \alpha_t \bm c_t  \, \label{eq:PSGD1} \\ 
\bm x_{t+1} &= \Pi_{\mathcal X} ( {\bm y}_{t+1}) \label{eq:PSGD2}
\end{align}
\end{subequations}
where 
$
\mathbb E \big[ \bm c_t \big| \mathcal F_t \big] = \nabla l(\bm x_t) \, \text{and}\, \alpha_t= {a}/{(u+t)^\gamma }
$
for $a>0$, $u\in \mathbb R_+$, $\gamma \in [0,1]$. Above $\nabla l(\bm x)$ can be either the gradient or a sub-gradient of $l$. We let $\mathcal X^\star = \argmin_{\bm x \in \mathcal X} l(\bm x)$ be the set of optimizers of \eqref{eq:l_obj}.

Previously, we required Conditions \eqref{fcond:1} and \eqref{fcond:2}, which jointly placed assumptions on the iterates and objective. Now that we have specified the iterative procedure, we can decouple to give conditions that only depend on the properties of the objective function.

 \begin{assumption}[Gradient Condition]  There exists a positive constant $\kappa >0$ such that for all $\bm x \in \mathcal X$ 
\begin{align} 
\nabla l( \bm x)^\top  ( \bm x - \bm x^\star) \geq \kappa \| \bm x - \bm x^\star \|, \qquad  \label{cond:D1}\tag{D1}
\end{align}
where $\bm x^\star = \Pi_{\mathcal X^\star}(\bm x)$.  
 \end{assumption}

\begin{assumption}[Sub-Exponential Noise]  There exists a $\lambda >0$ such that 
\begin{align} 
\sup_{t \in \mathbb N} \mathbb E [ e^{ \lambda || \bm c_t ||} | \mathcal F_t ] < \infty \, .   \label{cond:D2} \tag{D2}	
\end{align}
\end{assumption}

\begin{figure}[ht!]
\centering
\begin{subfigure}{.5\textwidth}
  \centering
  \includegraphics[trim={0 0 18cm 0},clip, width=0.8\linewidth]{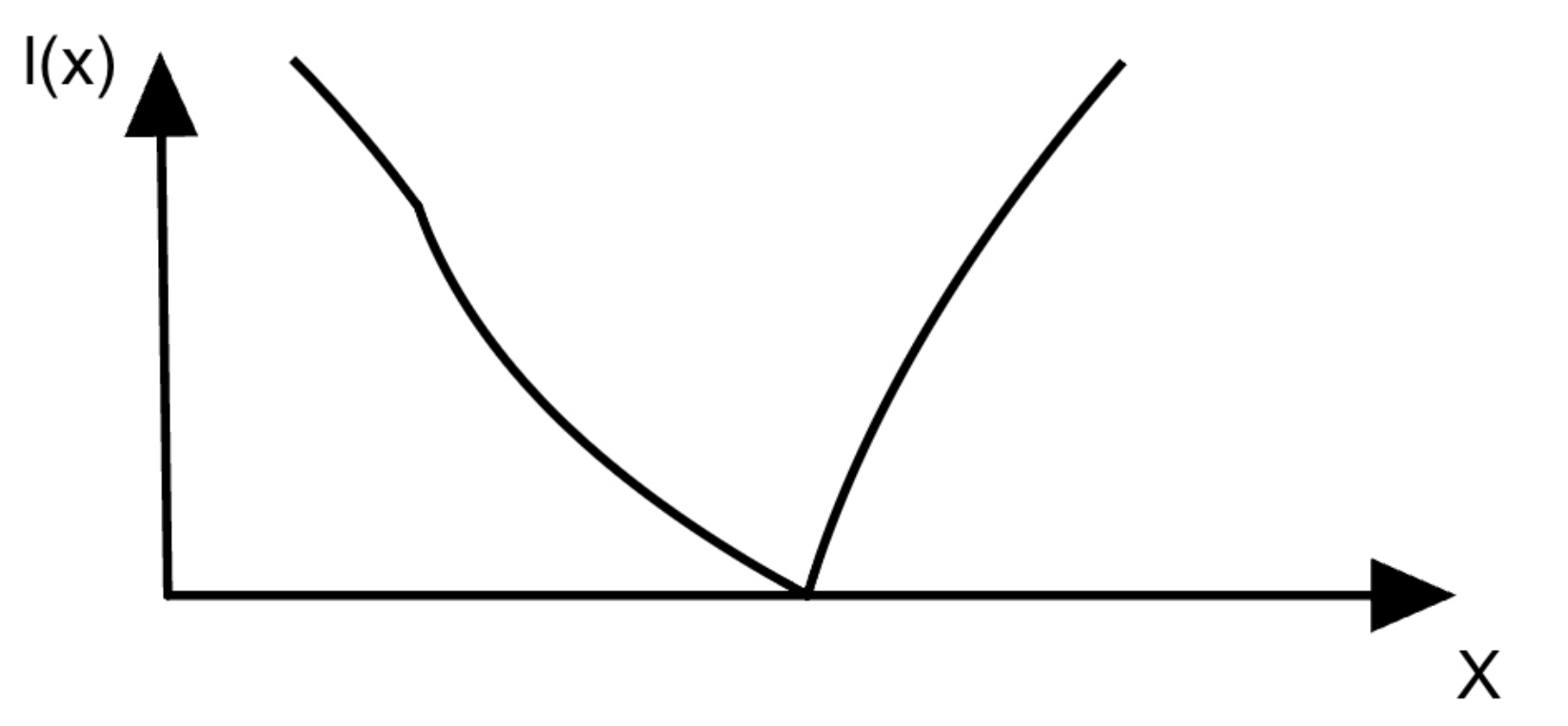}
  \caption{Gradient Condition \eqref{cond:D1}.}
  \label{fig:sub1Con}
\end{subfigure}%
\begin{subfigure}{.5\textwidth}
  \centering
  \includegraphics[trim={18cm 0 0 0},clip, width=0.8\linewidth]{Figures/Grads_Short.pdf}
  \caption{Convexity and Sharpness \eqref{cond:Sharp}}
  \label{fig:sub2Con}
\end{subfigure}
\
\caption{Under the gradient condition \eqref{cond:D1}, the objective need not be convex nor continuously differentiable. We require the derivative in the direction of the optimum to be non-zero. Under convexity and sharpness \eqref{cond:Sharp}, the envelope of the function is bounded below by a cone. Here condition \eqref{cond:D1} is satisfied.
}\label{fig:2}
\end{figure}

 Conditions \eqref{cond:D1} and \eqref{cond:D2} replace Conditions \eqref{fcond:1} and \eqref{fcond:2}. Let's interpret these new conditions. Firstly, \eqref{cond:D1} states that the (unit) directional derivative in the direction from $\bm x$ to $\bm x^\star$ is bounded above by $-\kappa$. (See Figure \ref{fig:2}.) Condition \eqref{cond:D2} assumes that the tail behavior of the gradient estimates is sub-exponential. (See Lemma \ref{lem:fcond2}.) 

 We can prove the following result that holds as a consequence of Theorem \ref{Cor:1}.

\begin{theorem}\label{ThrmNew}
If Condition \eqref{cond:D1} and \eqref{cond:D2} hold and $
\alpha_t = {a}/{(u+t)^\gamma}
$ for $a,u>0$ then PSGD satisfies 
\begin{align*}
\mathbb P
\left(
 \min_{\bm x^\star \in \mathcal X^\star} \left\|
	\bm x_{t+1} - \bm x^\star 
\right\|
\geq 
z
\right)
\leq J e^{-I t^\gamma z}
,\quad 
\mathbb E \left[
	 \min_{\bm x^\star \in \mathcal X^\star} \left\|
	\bm x_{t+1} - \bm x^\star 
\right\|
\right]
\leq \frac{K}{t^\gamma} ,
\quad 
\mathbb E \left[
	l(\bm x_{t+1}) - \min_{\bm x \in \mathcal X} l(\bm x) 
\right] \leq \frac{L}{t^\gamma}\, 
\end{align*}
where above $J, I, K ,L$ are positive constants.
\end{theorem}

For Stochastic Gradient Descent (SGD), the expected distance from $\bm x_t$ to the set of optima $\mathcal X^\star$ is known to converge at rate $\Omega(1/\sqrt{t})$. 
The convergence rate found in Theorem \ref{ThrmNew} is faster than that typically assumed for SGD. 
This is because a tighter exponential concentration occurs around the optimum in such cases. While the Gaussian concentration around the optimum for smooth convex objectives has been known for around 70 years (\cite{Chung,fabian1968asymptotic}), the exponential concentration found here does not appear in prior work on PSGD.

\begin{darkblue}

\begin{remark}[Convexity and Sharpness.]\label{rmk:Sharp}
Sharpness is a condition which, stated informally, ensures that an objected function has a {\fontfamily{phv}\selectfont V}-shape around its optimum. This is considered in the paper of \cite{davis2019stochastic}. A close relationship exists between our a gradient condition \eqref{cond:D1} and the sharpness condition. In particular, the two conditions are equivalent for convex optimization problems.

    A function $l(\bm x)$ is sharp if for all $x \in \mathcal X$
    \begin{equation}
     l(\bm x) - \min_{\bm x^\star \in \mathcal X}l(\bm x^\star) \geq \kappa' \min_{\bm x^\star \in \mathcal X^\star} \| \bm x- \bm x^\star \|. \label{cond:Sharp} \tag{D1$^\prime$}
    \end{equation} 
 The lemma below proves that the gradient condition \eqref{cond:D1} implies sharpness and, for convex functions, the two properties are equivalent:
    
        \begin{restatable}{lemma}{lemsharpe}\label{lem:EqivSharpness}
            If the function $l(\bm x)$ is absolutely continuous then the gradient condition \eqref{cond:D1} implies the function is sharp, \eqref{cond:Sharp}. 
            Moreover, if the function $l(\bm x)$ is convex, 
            then the gradient condition \eqref{cond:D1} is equivalent to the function being sharp \eqref{cond:Sharp}. 
        \end{restatable}
    \noindent We prove Lemma \ref{lem:EqivSharpness} in Section \ref{sec:SharpProof} of the E-companion. The immediate consequence of this lemma is that Theorem \ref{ThrmNew} holds for sharp convex functions.
    So, there is a tighter exponential concentration for sharp convex objectives when compared with the Gaussian concentration bounds found for smooth convex objectives.
    \end{remark}
\end{darkblue}

\begin{darkblue}
    \begin{remark}[Smooth Functional Constraints]
Suppose the optimization \eqref{eq:l_obj} takes the form 
	\begin{align}
\text{minimize} \qquad l(\bm x) \qquad
&\text{subject to} \qquad 
l_i(\bm x) \leq 0, \quad i=1,...,m 
\quad \text{over} \quad \bm x \in \mathbb R^d \, , 
\label{eq:l_obj_constrainted}	
\end{align}
where $l(\bm x)$ and $l_i(\bm x)$ are smooth convex functions defining the bounded constraint set $\mathcal X = \{\bm x \in \mathbb R^d : l_i(\bm x) \leq 0, i=1,...,m \}$.  
It is argued that a stochastic approximation algorithm obeys a central limit theorem if there are $m_0$ active constraints at the optimum with $m_0<d$. See \cite{kushner1978stochastic,shapiro1989asymptotic,duchi2021asymptotic,davis2023asymptotic}.
However, if $m_0 \geq d$, the normal approximation degenerates. In this case, our results can be applied. With the following lemma and Theorem \ref{ThrmNew}, we see that PSGD obeys an exponential concentration bound rather than a normal approximation. 

\begin{restatable}{lemma}{lemfunctional}\label{lem:functional}
    Suppose that at the optimum 
    $
        -\nabla l(\bm x^\star) \in \textrm{relint }
\mathcal N_{\mathcal X}(\bm x^\star) \,
    $, where $\mathcal N_{\mathcal X}(\bm x^\star) := \{ \bm v : \bm v^\top (\bm x^\star - \bm y) \leq 0\, , \forall \bm y \in \mathcal X   \} $ and $\nabla l(\bm x^\star) \neq 0$
    and that there are at least $d$ active constraints at $\bm x^\star$ (w.l.o.g. $i=1,...,d$) and
    \begin{align}
        \{ \nabla l_i(\bm x^\star): i =1,...,d\}\text{ are linearly independent}
    \tag{D1$^{\prime \prime}$}
    \end{align}
    then the function $f$ is sharp at $\bm x^\star$ and Assumption \eqref{cond:D1} holds.
\end{restatable}

\noindent  A proof of Lemma \ref{lem:functional} is given in Section \ref{sec:SharpProof} of the E-companion.
The premise of the above lemma is taken from Assumption B from \cite{duchi2021asymptotic}. However, rather than a Gaussian approximation as found in \cite{duchi2021asymptotic} for $m_0<d$ constraints, a consequence of the above Lemma is that exponential concentration will hold in for PSGD 
if there are $m_0\geq d$ linearly independent, active constraints at the optimum. Because of the degeneracy indicated in prior works; the Gaussian approximation is not asymptotically optimal when the set of active constraints has full rank. Essentially, there is insufficient smoothness at the optimum for a central limit theorem to hold. Large deviation effects will likely determine the asymptotically optimal concentration at the optimum.
When the Gaussian approximation fails, the theory of asymptotic optimality for stochastic optimization with constraints appears to be open.
    \end{remark}
\end{darkblue}

\begin{darkblue}
\begin{remark}[Projection]
Although projection is a common requirement for stochastic gradient descent,  the projection step \eqref{eq:PSGD2} can present computational overhead, so we discuss that here.

There are settings such as sharp objective functions where the optimum belongs to the interior of the constraint set. In this case, a finite number of projections are required. (A proof is given in Proposition \ref{PropProj} in the E-companion.) 

{Some constraints exhibit low complexity projection. It is common to select a set $\mathcal X$ that allows simple projection e.g. a box or disk containing $\mathcal X^\star$. For convex constraints $\{ \bm z : \bm g_j( \bm z) \leq 0, j=1,..,d\}$, the dual of a constraint set is $\mathbb R^d_+ = \{ \bm x : \bm x \geq 0\} $ and thus the dual has simple projection. Low complexity projections exist for single constraint problems such as projection onto the probability simplex (\cite{michelot1986finite,duchi2008efficient}). Chapter 7 of \cite{hazan2016introduction} gives a number of examples of fast projection available with conditional gradient algorithms. See also \cite{bertsekas2015convex} for further examples of low complexity projection.}

There are practical general projection algorithms. 
The cyclic projection algorithm of \cite{bregman1967relaxation} can be used for the intersection of a finite number of convex sets. Here, Bregman also proposes other non-Euclidaen distances that can be used to simplify projection. 
\cite{mandel1984convergence} proves Bregman's algorithm converges linearly for polytope constraints. A number of linear convergent and parallelizable projection algorithms are given in \cite{censor1997parallel}.

Projection is a standard requirement in the analysis of SGD. However, projection is not a requirement of our general result Theorem \ref{Cor:1}; instead, we require iterates to be bounded. For instance, we will apply our results to Stochastic Frank Wolfe as a non-projective alternative to PSGD shortly. Further, the boundedness of $\mathcal X$ can be removed depending on the learning rates applied. For instance, our linear convergence results apply Stochastic Gradient Descent (without projection), and neither projection nor bounded iterates are required for linear convergence.
\end{remark}
\end{darkblue}

\subsection{The Exponential Approximation}
\label{sec:Counter}

The limit distribution of stochastic approximation is a normal distribution when the shape of the objective around the optimum is approximately quadratic, e.g. like $\| \bm x\|^2$. When the curvature behaves as $\| \bm x\|$, our analysis proves exponential tail behavior. 
A natural question is if the limiting distribution of iterates away from the optimum are exponentially distributed under the constant drift condition? That is do we have convergence in distribution:
$$
\frac{x_t - x^\star}{\alpha_t} \xrightarrow[t \rightarrow \infty]{\mathcal D} X, 
$$
 where $X$ is exponentially distributed? The short answer is no. In general, the limit distribution is \emph{not} exponential. 
For a simple counter-example, consider projected stochastic gradient descent where $\mathcal X = \mathbb R_+$ and $c_t$ are i.i.d. random variables with $c_t = -1$ with probability $p$ and $c_t = 1$ with probability $1-p$. If we fix $\alpha$, 
we can already see that an exponential distribution limit is not possible since the process $x_t/\alpha$ belongs to the set $ \mathbb Z_+$. If $p > 1/2$, then the limit distribution is geometrically distributed, not exponential.  For general distributions of $c_t$, the limit distribution is given by an integral equation. (See Lindley's Integral Equation \cite[Corollary 6.6]{asmussen2003applied}.) However,  the resulting distributions all exhibit exponential tail bounds. (See Kingman's Bound c.f. \cite{kingman1964martingale})

Convergence in distribution likely holds. However, the limit $X$ is unlikely to be an exponential distribution,  and it is unlikely to have a simple form. 
We can not aggregate fluctuations in the same manner as found in the normal approximation.
We cannot expect a simple statistic like the Fisher Information to determine the directions of statistical error because the stochastic approximation process is much more concentrated.  A theory of asymptotic optimality is likely to be characterized in terms of exponents rather than distributions. Sharpness is a natural condition for a convex function in the same way smoothness is. However, different techniques are required here because errors and stepsizes are of the same order of magnitude, which is perhaps why exponential tail bounds are rarely seen in prior literature on stochastic approximation. 
However, as we see in this article, we can understand convergence behavior by constructing these exponential concentration bounds.

\begin{darkblue}
\subsection{Kiefer-Wolfowitz Algorithm: Exponential Concentration}\label{sec:KWAlgorithm}
So far, we have applied the concentration bound in Theorem \ref{Thrm1} to
Projected Stochastic Gradient Descent (PSGD). In this section, we consider the Kiefer-Wolfowitz algorithm. 
%
The Kiefer-Wolfowitz is a well-known alternative to Robbins and Monro's Stochastic Gradient Descent algorithm. Here gradient estimates are replaced by noisy finite difference operators.

We consider the Kiefer-Wolfowitz algorithm under the analogous sharpness in noise conditions that we considered for a PSGD. Typically Kiefer-Wolfowitz algorithm has the worst rate of convergence than PSGD. However, under the non-vanishing gradient condition, we show that the Kiefer-Wolfowitz algorithm with an appropriate finite difference estimator will have the same concentration and asymptotic convergence rate as PSGD. 

Consider the optimization:
	\begin{equation}
\text{minimize} \qquad l(\bm x) := \mathbb E_{\hat w} [l(\bm x,\hat w)] \qquad \text{over} \quad x \in \mathcal X \, , 
\label{eq:l_obj_KF}	
\end{equation}
where $\hat w$ is a random variable. For $\nu \in \mathbb R $, we define the vector 
$
\bm l(\bm x + \bm \nu,  w) := (l(\bm x + \nu \bm e_i , w ) : i =1,...,d)
$
where $\bm e_i$ is the $i$th unit vector.
The Kiefer--Wolfowitz (KW) algorithm is as follows:
\begin{subequations} \label{eq:KW}
\begin{align}
{\bm c}_{t} &= \frac{\bm l(\bm x_t + \bm \nu, \hat w^+_t ) - \bm l(\bm x_t - \bm \nu ,\hat w^-_t ) }{2\nu} \, \label{eq:KWdef1} \\ 
{\bm y}_{t+1} &=\bm x_t - \alpha_t \bm c_t  \, \label{eq:KWdef2} \\ 
\bm x_{t+1} &= \Pi_{\mathcal X} ( {\bm y}_{t+1}) \label{eq:KWdef3}
\end{align}
\end{subequations}
where 
$
\alpha_t= \frac{a}{(u+t)^\gamma }
$
for $a>0$, $u\in \mathbb R_+$, $\gamma \in [0,1]$. Above $\hat w_t^+, \hat w^-_t$ are IIDRVs equal in distribution to $\hat w$. 

Notice the main change from the PSGD algorithm is that  $\bm c_t$ is not longer an unbiased estimator of $\nabla l(\bm x_t)$. However, for sufficiently well-behaved functions, there exists a constant $c$ such that for all $x\in\mathcal X$
\begin{align}
\Big\| \nabla l (\bm x) - \frac{\bm l(\bm x + \bm \nu ) - \bm l(\bm x - \bm \nu ) }{2\nu} \Big\| \leq  c \nu^2 \, .
\label{eq:KWconds}   \tag{D3} 
\end{align}
We further assume that the random variable $l(\bm x, \hat w)$ has a variance that is uniformly bounded over $\bm x \in \mathcal X$.
This is a standard assumption for the analysis of the KW algorithm, see \cite{fabian1967stochastic}. We will assume Conditions \ref{cond:D1} and \ref{cond:D2} hold, with $\bm c_t$ as defined in \eqref{eq:KWdef1}. Ordinarily, convergence of the KW algorithm requires $\nu$ to decrease with time. However, we see that this is not necessary for sharp functions. The parameter $\nu$ needs to be below a certain threshold, and once satisfied exponential concentration results hold.

\begin{restatable}{theorem}{kieferwolfowitz}\label{ThrmKW}
If Conditions \eqref{cond:D1}, \eqref{cond:D2}, \eqref{eq:KWconds} hold and if 
\[
 \nu \leq \sqrt{\frac{\kappa}{3c}}
\]
then the Kiefer-Wolfowitz algortihm satisfies 
\begin{align*}
\mathbb P
\left(
 \min_{\bm x \in \mathcal X} \left\|
	\bm x_{t+1} - \bm x 
\right\|
\geq 
z
\right)
\leq \hat J e^{-\hat I t^\gamma z}
,\quad 
\mathbb E \left[
	 \min_{\bm x \in \mathcal X} \left\|
	\bm x_{t+1} - \bm x 
\right\|
\right]
\leq \frac{\hat K}{t^\gamma} ,
\quad 
and
\quad 
\mathbb E \left[
	l(\bm x_{t+1}) - \min_{\bm x \in \mathcal X} l(\bm x) 
\right] \leq \frac{\hat L}{t^\gamma}\, 
\end{align*}
where above $\hat J, \hat I, \hat K ,\hat L$ are positive constants.
\end{restatable}

The proof of Theorem \ref{ThrmKW} is given in Section \ref{sec:KWPRoof} of the E-companion.

If we take $\alpha_t = 1/t$, then we see that the Kiefer-Wolfowitz algorithm has a convergence rate 
$\mathbb E \| x_t - x^\star \| = O(1/t)$. This is faster than the $O(1/t^{1/3})$ rate, which is typically found to be optimal for the KW algorithm. 
Again typically, KW iterations follow a normal approximation. For instance, see \cite{ruppert1982almost}. However, if there is non-vanishing drift, then an exponential concentration is more appropriate than a normal approximation.
It is interesting that a constant finite difference approximation can be used to obtain results. 

\end{darkblue}

\begin{darkblue}
\subsection{Stochastic Frank-Wolfe: a non-Projective Algorithm}
\label{sec:SFW}
We now investigate exponential concentration in stochastic algorithms that do not require projection.
Non-vanishing negative drift is the main requirement for exponential concentration, not projection or boundary effects. 
We emphasize this by proving the exponential concentration for a projection-free algorithm, specifically, the Stochastic Frank-Wolfe algorithm.

The Frank-Wolfe algorithm or Conjugate Gradient algorithm, as it is sometimes called, is proposed as the standard ``{projection-free}" alternative to projected gradient descent algorithms, see \cite{jaggi2013revisiting} and \cite{hazan2012projection}. 
We form an analysis of the Stochastic Frank-Wolfe (SFW) algorithm, as described by \cite{hazan2016variance}. 
This is a standard implementation of Frank-Wolfe with a sample estimate of the gradient. 
As with the PSGD and KW algorithms, we choose a standard stochastic optimization algorithm. Our aim is not algorithm design but instead to emphasize that exponential (rather than normal) concentration can naturally occur in stochastic approximation depending on the geometry of the problem. 

Consider the same setting from Section \ref{sec:PSGDresult}. That is we wish to solve the optimization 
	\begin{equation}
\text{minimize} \qquad l(\bm x) \qquad \text{over} \quad x \in \mathcal X \, . 
\label{eq:l_obj_2}	
\end{equation}
For a sequence of positive integers $(m_t \in \mathbb N : t \in \mathbb N)$
 and a sequence $(\alpha_t \in (0,1) : t \in \mathbb N )$, the Stochastic Frank-Wolfe algorithm is defined as follows 
 \begin{subequations}\label{eq:FrankWolfe}
 \begin{align}
      & \bm c_t = \sum_{i=1}^{m_t} \frac{\bm c^i_t }{ m_t} \label{eq:FrankWolfe0}\\
     & \bm v_t \in \argmin_{x \in \mathcal{X}} \; \bm c_t^\top \bm x \label{eq:FrankWolfe1}
\\
    & \bm x_{t+1} = (1-\alpha_t) \bm x_t + \alpha_t \bm v_t \, . \label{eq:FrankWolfe2}
\end{align}     
 \end{subequations}
Above,  where $\bm c^i_t$ are  random variables, independent (after conditioning on $\bm x_t$) with a uniformly bounded variance such that 
$ \mathbb E [\bm c^i_t | \mathcal F_t] = \nabla l(\bm x_t)$ . 
We continue to assume that $\mathcal X$ is a closed bounded set. In addition to condition \eqref{cond:D1} and \eqref{cond:D2}, we add the following conditions. 
\smallskip

\begin{assumption}[Smooth convex square-error]
For the error $\epsilon(\bm x) := l(\bm x) - l(\bm x^\star)$, 
\begin{align}
\epsilon (\bm x)^2\text{ is a smooth convex function.}     \label{cond:E1} \tag{E1} 
\end{align}
\end{assumption}

\begin{assumption}[Interior Optimum] The set of optima belongs to the interior of the set $\mathcal X$. That is 
\begin{align}
\mathcal X^\star \subset \mathcal X^\circ .    \label{cond:E2}\tag{E2}
\end{align} 
\end{assumption} 

We briefly discuss these two conditions. Analogous to a smoothness convex function having quadratic behavior at the optimum, Condition \eqref{cond:E1} requires that the objective function has a behavior that behaves like the distance function to the optimum. We illustrate this with Lemma \ref{lem:posdef} in the E-companion. Here, we show that Condition \eqref{cond:E1} is satisfied if we take $l(x)$ to be the distance to the desired set of optimal points:
\[
 d_{\mathcal X^\star}(x) = \min_{x^\star \in \mathcal X^\star}
        \|  
            \bm x - \bm x^\star 
        \|_S \, .cvb 
\]
Here $S$ is a positive semi-definite matrix.

Condition \eqref{cond:E2} requires that the optimum is in the interior. When analyzing projected stochastic gradient descent, one case we discussed is when the optimum is on the boundary of the constraint set. Interestingly, the case of the Stochastic Frank-Wolfe algorithm is different: we require the optimum is \emph{not} on the boundary. Since Frank-Wolfe is a non-projective optimization algorithm, the results of this section emphasize that exponential concentration is not the property of projection on a specific boundary type but is really about non-vanishing drift at the optimum. 

The main result of this subsection is given below. It gives sufficient conditions for exponential concentration for the Frank-Wolfe algorithm. 

\begin{restatable}{theorem}{FrankWolfe}\label{thrm:FW}
For learning rates of the form 
$
\alpha_t = {a}/{(u+t)^\gamma}
$
with $a,u>0$ and  $\gamma \in [0,1]$,
    if Conditions \eqref{cond:D1}, \eqref{cond:D2}, \eqref{cond:E1} and \eqref{cond:E2} hold and if $m_t \geq (3 \sigma / \kappa \alpha_t)^2 $
    then 
    \begin{equation*}
        \mathbb P \left( 
            l(\bm x_{t+1} ) - \min_{\bm x \in \mathcal X}l( \bm x ) \geq z 
            \right)\leq I e^{- \frac{J}{\alpha_t} z} \, ,
    \end{equation*}
    for constants $I$,$J$.
\end{restatable}

The proof of Theorem \ref{thrm:FW} can be found in Section \ref{sec:FWProofs} of the E-companion. 

\end{darkblue}

\subsection{Linear Convergence under Exponential Concentration}\label{sec:Linear}

The linear convergence of Projected Stochastic Gradient Descent (PSGD) on convex objectives is established in Theorem 3.2 of \cite{davis2019stochastic}.
	This result relies on a normal approximation concentration result \cite[Theorem 4.1]{Drusvyatskiy}, which is not as tight as the exponential concentration bound in Theorem \ref{ThrmNew} above. 
 Thus, below in Theorem \ref{ThrmLinear}, we provide an improvement to Theorem 3.2 of \cite{davis2019stochastic}.
We give a general version of the linear convergence that can be proven under the conditions of Theorem \ref{Thrm1}. The result does not require the set $\mathcal X$ to be bounded.
From this, extensions of these linear convergence results hold for the Projected Stochastic Gradient Descent, the Kiefer-Wolfowitz algorithm, and the stochastic Frank-Wolfe algorithm.
These results are stated in Section \ref{sec:LinConvAppendix} of the E-companion.
	
We wish to solve the optimization problem \eqref{eq:l_obj}.
We consider a Stochastic Approximation algorithm \eqref{eq:SA} implemented over several stages, $s=1,...,S$. We let $t_s$ be the number of iterations in the $s$th stage. 
The idea is that within each stage $s$, the learning rate $\hat \alpha_s$ is fixed and is chosen so that the error of the stochastic approximation algorithm should be halved by the end of each stage.  
Specifically, we let $\hat{ \bm x}_s$ be the state at the end of stage $s$.
We define $T_s = \sum_{s'=1}^s t_{s'}$ and 
\begin{align}\label{eq:PSGDLinear}
\hat{\bm x}_s &= \bm x_{T_s}, \qquad \text{and} \qquad \alpha_t = \hat{\alpha}_s,\qquad \text{for}\quad T_{s-1} \leq t < T_{s}, \text{ and } s =1,...,S  \,. 
\end{align}

 The following theorem gives choices for $\hat{\alpha}_s$ and $t_s$ to ensure a linear rate of convergence.

\begin{restatable}{theorem}{PSGDLinearThrm} \label{ThrmLinear}
Assume Conditions \eqref{fcond:1} and \eqref{fcond:2} hold for a stochastic approximation procedure with rates given in \eqref{eq:PSGDLinear}:

\noindent a) If, for $\hat \epsilon >0$ and $\hat\delta \in (0,1) $, we set
\begin{align*}
S = \log\left(
	\frac{F}{\hat \epsilon}
\right)
,
\quad 
\hat{\alpha}_s = \frac{2^{-s}F \kappa }{  E \log \Big(\frac{RS}{\hat{\delta}}\Big) }
, \quad \text{and} \quad 
t_s = \left\lceil \frac{2}{\kappa^2} \log\left(
	\frac{RS}{\hat \delta}\right)  \right\rceil
\end{align*}
then with probability greater than $1-\hat \delta$ it holds that
$
\min_{\bm x \in \mathcal X^\star}\left\|
	\hat{\bm x}_S - \bm x 
\right\| \leq \hat \epsilon \, .
$
Moreover, the number of iterations \eqref{eq:PSGDLinear} required to achieve this bound is
\begin{align*}
\Big\lceil \log_2\Big( \frac{F }{\hat \epsilon}\Big) \Big\rceil \left\lceil \frac{2}{\kappa^2} \log \left(
	\frac{R}{\hat \delta}\right)  + \log \Big(\Big\lceil \log_2\Big( \frac{F }{\hat \epsilon}\Big) \Big\rceil\Big)
 \right\rceil \, .%
\end{align*}
(Above $F=f(\bm x_0) -f(\bm x^\star)$ and $R$ and $E$ are time-independent constants that depend on the constants given in Conditions \eqref{fcond:1} and \eqref{fcond:2}.)

b) For 
$
\hat{\alpha}_s = \frac{a}{2^s \log (s+1)}$ \text{and}  $t_s = \log^2 (s+1)
$, 
there exists positive constants $A$ and $M$ such that $\forall \hat \delta \in (0,1)$ if $a \geq A/ \hat \delta$ then 
\begin{align*}
\mathbb P
\left(
	\min_{\bm x \in \mathcal X^\star} \|  \hat{\bm x}_s - \bm x\| \leq 2^{-s} M, \quad \forall s \in \mathbb N
\right)
\geq 1- \hat \delta \, .
\end{align*}
\end{restatable}
The proof of Theorem \ref{ThrmLinear} is given in Section \ref{sec:LinConvAppendix} of the E-companion. We apply an improved exponential concentration bound and make some adjustments; however, other than that, the argument largely follows that of \cite[Theorem 3.2.]{davis2019stochastic}, so we refer the reader to \cite{davis2019stochastic} also. 

The bound in Theorem \ref{ThrmLinear}a) above has an order $O( \log \hat \epsilon^{-1} \big[ \log \log \hat \epsilon^{-1}  + \log \hat \delta^{-1}  \big])$.
The above bound is the same order as the best bound found in \cite{davis2019stochastic}, namely Theorem 3.8, which holds for an ensemble method consisting of three adaptively regularized gradient descent algorithms. The sample complexity is improved due to the improved concentration bound applied.

We see that a bound of the same form holds for PSGD, Kiefer-Wolfowitz, and Frank-Wolfe. The dependence on $\hat \delta$ improves the probabilistic concentration found of PSGD in \cite{davis2019stochastic}. As discussed, the main innovation here is the initial concentration bound. After this, our proof follows an analogous proof to \cite{davis2019stochastic}.

Unlike Theorem \ref{ThrmNew}, Theorem \ref{ThrmLinear}a) suggests that we require a refined understanding to calibrate parameters to improve convergence. 
The implementation of Theorem \ref{ThrmLinear}b) only requires one parameter, $a$, which needs to chosen sufficiently large. (For instance, experiments could increase the parameter $a$ until  convergence is observed.)
So, although there is some cost to the algorithm's complexity, we do not require detailed knowledge of the problem at hand to implement a geometrically convergent algorithm.
Further, Theorem \ref{ThrmLinear}b) holds for a stronger mode of convergence, in that the geometric convergence holds for all time with arbitrarily high probability.

We note that the above bound applies when the function $f$ and constraint sets $\mathcal X$ are unbounded. This is because we apply Lemma \ref{lem:LMGF} under constant step sizes. For this result, the bounded constraint assumption is not required. 

\section{Proofs}\label{sec:Proofs}

This section proves our main result Theorem \ref{Cor:1}.
We then apply this to Projection Stochastic Gradient Descent to prove Theorem \ref{ThrmNew}. 
The proofs of the remaining results are contained in the E-companion.

\subsection{Proof of Theorem \ref{Cor:1}}
\label{sec:ProofThrm1}
{\darkblue
Several constants are introduced in the proof of Theorem \ref{Cor:1}. For later reference, these are listed in Section \ref{sec:constants} of the E-companion.} 

The proof of Theorem \ref{Cor:1} relies on Proposition \ref{Thrm1}, which is a somewhat more general yet more abstract version of Theorem \ref{Cor:1}. 
This result establishes that once the sum $\sum_{s=1}^t \alpha_s$ is sufficiently large, the error of stochastic iterations is of the order of $\alpha_t$. 
Much like Theorem \ref{Cor:1} the key observation is that $\alpha^{-1}_t ( f(\bm x_t) - f(\bm x^\star))$ has an exponential concentration rather than the normal concentration of $\alpha^{-1/2}_t ( f(\bm x_t) - f(\bm x^\star))$. 

For the stochastic iterates described in Section \ref{sec:Notation}, we will assume that $\{\alpha_t \}_{t \geq 0}$ is a deterministic non-increasing sequence such that
\begin{equation}\label{alphacond}
\sum_{t=0}^\infty \alpha_t = \infty, \qquad \liminf_{t \rightarrow \infty }\,\frac{\alpha_{2t}}{\alpha_t} >0  \qquad \text{and } \qquad 
\lim_{t \rightarrow \infty}\frac{\alpha_{t} - \alpha_{t+1} }{ \alpha_t} =0 \, .
\end{equation}
We do not need to assume that $\alpha_t \rightarrow 0$. Later we consider small but constant step sizes.
This condition is satisfied by any sequence of the form $\alpha_t = a/(u+t)^\gamma$ for $a,u>0$ and $\gamma \in [0,1]$.

\begin{restatable}{proposition}{Thrm1}\label{Thrm1} 
    When Conditions \eqref{fcond:1} and \eqref{fcond:2} are satisfied,
there exist positive constants $E,G,H, T_0$ independent of $t$ such that 
\begin{align}\label{Lya:Thrm1eq1}
\mathbb P( f(\bm x_{t+1}) - f(\bm x^\star) \geq z ) 
\leq
e^{
- \frac{\kappa G^n }{2E\alpha_t}(z - \alpha_t B-F - \alpha_0 B+\sum_{s=\lfloor t/2^n \rfloor}^{t}\alpha_s\frac{\kappa}{2} )
} 
+
H {e^{-\frac{\kappa G^n}{2E\alpha_t}(z - \alpha_t B)}} \, .
\end{align}
for any $n$ with $t / 2^n >T_0$. 
Further, 
for any $t$ such that 
$
\sum_{s=\lfloor t/2^n \rfloor}^{t} \alpha_s  \kappa \geq  2 (F + \alpha_0B)
$
 there exists a constant $C$ such that 
\begin{equation}\label{Lya:Thrm1eq2}
\mathbb E [f(\bm x_{t+1}) - f(\bm x^\star)]
\leq  C \alpha_t \, .
\end{equation}
\end{restatable}

\subsubsection{Proof of Proposition \ref{Thrm1}.} \label{sec:Proofs1}

Let's briefly outline the proof of Proposition \ref{Thrm1}. The proof uses Lemma \ref{lem:1}, Lemma \ref{lem:LMGF}, Lemma \ref{LemmaB}, and Proposition \ref{Lya:Cor1}, which are stated below. 
Lemma \ref{lem:1}, although not critical to our analysis, simplifies the drift Condition \eqref{fcond:1} by eliminating some  terms and boundary effects. 
Lemma \ref{lem:LMGF}, on the other hand, is a important component of our proof. It converts the drift Condition \eqref{fcond:1} into an exponential bound which we then iteratively expand. The lemma extends Theorem 2.3 from \cite{hajek1982hitting}  
by allowing for adaptive time-dependent step sizes. 
Proposition \ref{Lya:Cor1} applies standard moment generating function inequalities to the results found in Lemma \ref{lem:LMGF}.
Lemma \ref{LemmaB} is a technical lemma used in the proof of Proposition \ref{Lya:Cor1}.
After Proposition \ref{Lya:Cor1}  is proven, the proof of Proposition \ref{Thrm1} follows.

We now proceed with the steps outlined above. We let \begin{equation} \label{def:L_t}
L_t := f(\bm{x}_t)-f(\bm{x}^\star)- \alpha_t B
\end{equation}
where $\alpha_t$ satisfies \eqref{alphacond}. 
First, we simplify the above Conditions \eqref{fcond:1} and \eqref{fcond:2} to give the Lyapunov conditions \eqref{eq:con1} and \eqref{eq:con2} stated below.
The following is a technical lemma.


\begin{restatable}{lemma}{lemone}
\label{lem:1}
Given Conditions \eqref{fcond:1} and \eqref{fcond:2}  hold, there exists a deterministic constant $T_0$ such that the sequence of random variables $(L_t : t \geq T_0)$ satisfies
\begin{equation}\label{eq:con1}
\mathbb E \big[ L_{t+1}  - L_t   \big| \mathcal F_t \big] \mathbb I [L_t \geq 0]  < - \alpha_t \kappa ,
\end{equation}
and 
\begin{equation}\label{eq:con2}
[| L_{t+1} - L_t | | \mathcal F_t] \leq \alpha_t Z \quad \text{where} \quad D:= \mathbb E [ e^{\lambda Z} ] < \infty \, .
\end{equation}
\end{restatable}
The proof can be found in the E-companion, Section \ref{sec:AppLem}.

%

Given Lemma \ref{lem:1}, we will now assume \eqref{eq:con1} and \eqref{eq:con2} hold in place of \eqref{fcond:1} and \eqref{fcond:2}. We will convert the drift condition \eqref{eq:con1} into an exponential bound and then iterate to give the bound below.

\begin{lemma}\label{lem:LMGF}
For any $t$ and $\hat t$ with $t\geq \hat t \geq T_0$ and for any $\eta>0$ such that $\alpha_{\hat t} \eta \leq \lambda$ then
\begin{equation*}
    \mathbb{E}[e^{\eta L_{t+1}}| \mathcal F_{\hat t}]\leq\mathbb{E}[e^{\eta L_{T_1}} | \mathcal F_{\hat t}]\prod_{k=\hat t}^{t}\rho_t + D\sum_{\tau=\hat t+1}^{t+1}\prod_{k=\tau}^{t}\rho_k \, ,
\end{equation*}
 where $\rho_t = e^{-\alpha_t\eta\kappa + \alpha_t^2 \eta^2 E}$, and $E := \mathbb{E}\left[ ({e^{\lambda Z}-1-\lambda Z})/{\lambda^2}\right]<\infty$.
\end{lemma}
\Beginproof{ of Lemma \ref{lem:LMGF}}
Let $Z_t = (L_{t+1}-L_t)/\alpha_t$. From \eqref{eq:con2}, we have $[|Z_t| |\mathcal F_t] \leq  Z $ where $\mathbb E [e^{\lambda Z}] < \infty$. From \eqref{eq:con1}, we have $\mathbb E [Z_t | \mathcal F_t] \leq -\kappa$ on the event $\{ L_t \geq 0\}$. Thus, on the event $\{ L_t \geq 0 \}$ the following holds:
\begin{align}
\mathbb{E}[e^{\eta (L_{t+1} - L_t)}|\mathcal{F}_t] =     \mathbb{E}[e^{\alpha_t\eta Z_t}|\mathcal{F}_t] &= 1 + \alpha_t\eta\mathbb{E}[Z_t|\mathcal{F}_t] + \alpha_t^2\eta^2\mathbb{E}\left[\frac{e^{\alpha_t \eta Z_t}-1-\alpha_t\eta Z_t}{\alpha_t^2\eta^2}\Big|\mathcal{F}_t\right] \nonumber\\
    &
\leq 1 + \alpha_t\eta\mathbb{E}[Z_t|\mathcal{F}_t] +  \alpha_t^2 \eta^2\sum_{k=2}^{\infty}\frac{1}{k!}\mathbb{E}[|Z_t|^k|\mathcal{F}_t]\eta^{k-2}\alpha_t^{k-2} \nonumber\\
    &\leq 1 - \alpha_t\eta\kappa + \alpha_t^2\eta^2\sum_{k=2}^{\infty}\frac{1}{k!}\mathbb{E}[Z^k]\lambda^{k-2} \nonumber\\
    &= 1 - \alpha_t\eta\kappa + \alpha_t^2\eta^2\mathbb{E}\left[\frac{e^{\lambda Z}-1-\lambda Z}{\lambda^2}\right]\label{eq:bound00}\\
    &= 1 - \alpha_t\eta\kappa + \alpha_t^2\eta^2 E\nonumber\\
    &\leq e^{-\alpha_t\eta\kappa + \alpha_t^2\eta^2E} =: \rho_t\, .\label{eq:bound1}
\end{align}
{\darkblue
We apply a Taylor expansion and the (conditional) Monotone Convergence Theorem in the first inequality above, see \cite[9.7e)]{williams1991probability}.} In the second inequality, we apply \eqref{eq:con1} and \eqref{eq:con2} above, and also recall that $\alpha_t$ is decreasing. In the final inequality, we applied the standard bound $1 + x \leq e^x$. We note that $\rho_t$ as define above satisfies $\rho_t<1$ whenever $\alpha_t < {\kappa}/{\eta E}$.
We note that $E$ is finite since by assumption $\mathbb E [ e^{\lambda Z} ] < \infty$. Also from the expansion given in \eqref{eq:bound00} (which holds by the Monotone Convergence Theorem), it is clear that $E$ is positive.

The bound \eqref{eq:bound1} holds on the event $\{L_t \geq 0\}$.
Now notice
\begin{align*}
    \mathbb{E}[e^{\eta L_{t+1}}|\mathcal{F}_{t}] 
&= 
\mathbb{E}[e^{\eta (L_{t+1}-L_t)}|\mathcal{F}_{t}]e^{\eta L_t}\mathbb{I}[L_t\geq 0] + \mathbb{E}[e^{\eta (L_{t+1}-L_t)}|\mathcal{F}_{t}]e^{\eta L_t}\mathbb{I}[L_t < 0]\\
    &\leq \rho_t e^{\eta L_t}\mathbb{I}[L_t \geq 0] + 
\mathbb E [ e^{\eta \alpha_t Z} ]
e^{\eta L_t}\mathbb{I}[L_t < 0]\\
    &\leq \rho_t e^{\eta L_t}\mathbb{I}[L_t \geq 0] + D \mathbb{I}[L_t< 0]\nonumber 
\\
&\leq 
\, 
 \rho_t e^{\eta L_t} + D\,.
\end{align*}
The first inequality applies the above bound \eqref{eq:bound1} and the second inequality applies the boundedness condition \eqref{eq:con2}. Taking  expectations above gives
\begin{equation*}
    \mathbb{E}[e^{\eta L_{t+1}} | \mathcal F_{\hat t}] \leq \rho_t \mathbb E [e^{\eta L_t} | \mathcal F_{\hat t}] + D.
\end{equation*}
By induction, we have
\begin{equation*}
    \mathbb{E}[e^{\eta L_{t+1}}| \mathcal F_{\hat t}]\leq\mathbb{E}[e^{\eta L_{T_1}}| \mathcal F_{\hat t}]\prod_{k=\hat t}^{t}\rho_t + D\sum_{\tau=\hat t+1}^{t+1}\prod_{k=\tau}^{t}\rho_k \, ,
\end{equation*}
as required.
\Endproof
Note that the above lemma does not require the set of values $\mathcal X$ to be bounded (or convex). This is a point that we will later utilize in the proof of Theorem \ref{ThrmLinear}.
%
%
The following is a technical lemma.

\begin{restatable}{lemma}{LemmaB}
\label{LemmaB}
If $\alpha_t$, $t\in\mathbb{Z}_{+}$, is a decreasing positive sequence, then 
\begin{equation}\label{eq:SumMin}
    \min_{s=\hat t,...,t}\left\{\frac{\sum_{k=s}^{t}\alpha_k}{\sum_{k=s}^{t}\alpha_k^2}\right\} = \frac{\sum_{k=\hat t}^{t}\alpha_k}{\sum_{k=\hat t}^{t}\alpha_k^2} \, .
\end{equation}
Moreover, if $\alpha_t$, $t\in\mathbb{Z}_{+}$ satisfies the learning rate condition \eqref{alphacond} then
\begin{equation}\label{eq:SumMin2}
\frac{1}{\alpha_{\lfloor t/2^n \rfloor}}\geq \frac{G^{{n}}}{\alpha_t} \qquad \text{and} \qquad 
    \min_{s= \lfloor t/2^n \rfloor ,...,t}\left\{\frac{\sum_{k=s}^{t}\alpha_k}{\sum_{k=s}^{t}\alpha_k^2}\right\} \geq  \frac{G^n}{\alpha_t}
\end{equation}
for some constant $G\in(0,1]$ and for $n \in \mathbb N$ such that $t/2^n >1$.
\end{restatable}
\begin{darkblue}
A proof is given in Section \ref{sec:AppLem} of the E-companion.  Looking ahead to the proof of Theorem \ref{Thrm1}, for step sizes of the form $\alpha_t = a/(u+t)^\gamma$, we have $G=1/4^\gamma$ and we will take $n=1$ for $\gamma <1$. For $\gamma = 1$, we need to have to be more careful choosing $n$, which will be a constant depending on $a,u, B$ and $F$. 
\end{darkblue}

With the moment generating function bound in Lemma \ref{lem:LMGF} and the bound in Lemma \ref{LemmaB}, we can bound the tail probabilities and expectation of $L_t$.

\begin{proposition}
\label{Lya:Cor1} 
For any sequence satisfying \eqref{alphacond}, 
there exists a constants $H$ and $Q$ such that
\begin{align}\label{eq:Prop2inq}
\mathbb P ( L_{t+1} \geq z ) 
\leq 
&\
e^{
- \frac{Q G^n }{\alpha_t}(z-F - \alpha_0 B+\sum_{s=\lfloor t/2^n \rfloor}^{t}{\alpha}_s\frac{\kappa}{2} )
} 
+
H {e^{-\frac{Q G^n}{\alpha_t}z}} \ 
\end{align}
for $z\geq 0$ and for $n \in \mathbb N$ such that $t/2^n >T_0$. Further, 
for $t$ is such that $\sum_{s=\lfloor t/2^n \rfloor}^{t} {\alpha}_s\frac{\kappa}{2} \geq F+\alpha_0 B   $, then
\begin{equation*}
    \mathbb{E}[L_{t+1}] \leq  \frac{\left(1 + H\right)}{Q G^n} \alpha_t\, .
\end{equation*}
\end{proposition}
\Beginproof{
 of Proposition \ref{Lya:Cor1}}
By Lemma \ref{LemmaB}, we see that 
\begin{equation*}
\frac{\lambda}{\alpha_{\lfloor t/2^n \rfloor}} \geq \frac{\lambda G^{{n}}}{\alpha_t}
\qquad \text{and }\qquad 
    \min_{\lfloor t/2^n \rfloor\leq s\leq t}\left\{\frac{\sum_{k=s}^{t}\alpha_k\kappa}{2\sum_{k=s}^t\alpha_k^2E}\right\}
\geq \frac{\kappa G^n}{2 \alpha_t E}
\end{equation*}
for a constant $G>0$. So that $\eta$ lower bounds the above two expressions, we take 
\[
\eta = Q\frac{G^n}{\alpha_t}\qquad \text{where}\quad Q= \lambda \wedge ( \kappa / 2E  )\, .
\]
 
We apply Lemma \ref{lem:LMGF} which gives
\begin{align}
    \mathbb{P}(L_{t+1}\geq z) &\leq e^{-\eta z}\mathbb{E}[e^{\eta L_{t+1}}]\nonumber\\
    &\leq e^{-\eta z}\mathbb{E}[e^{\eta L_{\lfloor t/2^n \rfloor}}]\prod_{k=\lfloor t/2^n \rfloor}^t\rho_t + e^{-\eta z}D\sum_{\tau=\lfloor t/2^n \rfloor+1}^{t+1}\prod_{k=\tau}^{t}\rho_k\nonumber\\
    &= e^{-\eta z}\mathbb{E}[e^{\eta L_{\lfloor t/2^n \rfloor}}]e^{\sum_{k=\lfloor t/2^n \rfloor}^{t}-\alpha_k\eta\kappa + \alpha_k^2\eta^2 E} + e^{-\eta z}D \sum_{ \tau = \lfloor t/2^n \rfloor+1 }^{t+1}e^{\sum_{k=\tau}^{t}-\alpha_t\eta\kappa + \alpha_t^2\eta^2 E}\label{eq:bound2}\ .
\end{align}
Notice, for $\eta$ as defined above, it holds that 
\begin{equation*}
    \sum_{k=\tau}^{t}-\alpha_k\eta\kappa + \alpha_k^2\eta^2 E \leq -\frac{1}{2}\sum_{k=\tau}^{t}\alpha_k\eta\kappa
    \leq -\frac{1}{2} ( t - \tau) \alpha_t \eta \kappa
    ,\qquad \forall \tau = \lfloor t/2^n \rfloor,...,t\ .
\end{equation*}
Applying this to \eqref{eq:bound2} gives
\begin{align}
    \mathbb{P}(L_{t+1} \geq z) 
    &\leq e^{-\eta z}\mathbb{E}[e^{\eta L_{\lfloor t/2^n \rfloor}}]e^{\sum_{k=\lfloor t/2^n \rfloor}^{t}-\alpha_k\eta\frac{\kappa}{2}} + e^{-\eta z}D\sum_{\tau=\lfloor t/2^n \rfloor+1}^{t+1}e^{-(t-\tau)\alpha_t\eta\frac{\kappa}{2}} \nonumber\\
    &\leq 
e^{-\eta z}\mathbb{E}[e^{\eta L_{\lfloor t/2^n \rfloor}}]e^{\sum_{k=\lfloor t/2^n \rfloor}^{t}-\alpha_k\eta\frac{\kappa}{2}} 
+ 
e^{-\eta z}D\frac{e^{\alpha_t\eta\frac{\kappa}{2}}}{1-e^{-\alpha_t\eta\frac{\kappa}{2}}} \label{eq:Ltail}
\end{align}
In the 1st inequality above we note that $\alpha_k \geq \alpha_t$ for all $k \leq t$. In the 2nd inequality, we note that the summation over $\tau$ are terms from a geometric series, so we upper bound this by the appropriate infinite sum.

Thus, the bound \eqref{eq:Ltail} becomes
\begin{equation*}
    \mathbb{P}(L_{t+1} \geq z) 
\leq 
\mathbb{E}\Big[e^{\frac{Q G^n L_{\lfloor t/2^n \rfloor}}{\alpha_t}}\Big]
e^{-\frac{Q G^n}{\alpha_t}(z+\sum_{s=\lfloor t/2^n \rfloor}^{t} {\alpha}_s\frac{\kappa}{2})} 
+ 
e^{-\frac{Q G^n}{\alpha_t}z} D \frac{e^{\frac{\kappa Q G^n}{2} }}{1-e^{-\frac{\kappa Q G^n}{2}}} \ .
\end{equation*}
Noting that $L_{\lfloor t/2^n \rfloor} \leq \max_{x\in\mathcal X} f(x) - \min_{x\in\mathcal X} f(x) + \alpha_0 B = F+\alpha_0 B$, by the definition of $F$. We simplify the above expression as follows
\begin{align}
\mathbb P ( L_{t+1} \geq z ) 
\nonumber 
\leq 
&\,
e^{
	\frac{Q G^n }{\alpha_t} 
	[
	F+\alpha_0 B
-z
- \sum_{s=\lfloor t/2^n \rfloor}^{t} {\alpha}_s\frac{\kappa}{2}
	]
}
+ e^{-\frac{Q G^n}{\alpha_t}z} D \frac{e^{\frac{\kappa QG^n}{2 }}}{1-e^{-\frac{\kappa Q G^n}{2}}}
\nonumber 
\\
\leq 
&\
e^{
- \frac{Q G^n }{\alpha_t}(z+\sum_{s=\lfloor t/2^n \rfloor}^{t}{\alpha}_s\frac{\kappa}{2} -F-\alpha_0 B)
} 
+
H {e^{-\frac{Q G^n}{\alpha_t}z}} \ .\label{eq:Ltailpart}
\end{align}
Above we define 
$H:=  D {e^{\frac{\kappa Q G^n}{2}}}/{(1-e^{-\frac{\kappa Q G^n}{2}})}$. 
This gives \eqref{eq:Prop2inq}.

Notice if $t$ is such that $F+\alpha_0 B - \sum_{s=\lfloor t/2^n \rfloor}^{t}{\alpha}_s\frac{\kappa}{2} \leq  0$, then the above inequality \eqref{eq:Ltailpart} can be bounded by 
\begin{equation*}
    \mathbb{P}(L_{t+1} \geq z) \leq \left(1 + H\right)e^{-\frac{Q G^n}{\alpha_t}z}.
\end{equation*}
Thus
\begin{align*}
    \mathbb{E}[L_{t+1}] &\leq  \mathbb E [ L_{t+1} \vee 0 ] = \int_0^\infty \mathbb{P}(L_{t+1} \geq z)dz \leq (1+H)\int_{0}^{\infty}e^{-\frac{Q G^n}{\alpha_t}z}\, dz = \left(1 + H\right)\frac{\alpha_t }{Q G^n} \ ,
\end{align*}
as required.
\Endproof

With Proposition \ref{Lya:Cor1} in place we can prove Proposition \ref{Thrm1}. 

\Beginproof{ of Proposition \ref{Thrm1}}
From Proposition \ref{Lya:Cor1}
\begin{align*}
\mathbb P ( L_{t+1} \geq z' ) 
\leq 
&\
e^{
- \frac{Q G^n }{\alpha_t}(z'-F - \alpha_0 B+\sum_{s=\lfloor t/2^n \rfloor}^{t}\alpha_s\frac{\kappa}{2} )
} 
+
H {e^{-\frac{Q G^n}{\alpha_t}z'}} \ 
\end{align*}
for $z'\geq 0$ where $f(\bm x_{t+1}) = L_{t+1} + \alpha_{t+1} B + f(\bm x^\star)$. Taking $z' = z - \alpha_{t+1} B$, 
gives
\begin{align*}
\mathbb P( f(\bm x_{t+1}) - f(\bm x^\star) \geq z ) 
&=
\mathbb P ( L_{t+1} \geq z - \alpha_{t+1} B) 
\\
&
\leq
e^{
- \frac{Q G^n }{\alpha_t}(z - \alpha_t B-F - \alpha_0 B+\sum_{s=\lfloor t/2^n \rfloor}^{t}\alpha_s\frac{\kappa}{2} )
} 
+
H {e^{-\frac{Q G^n}{\alpha_t}(z - \alpha_t B)}}\, ,
\end{align*}
which gives \eqref{Lya:Thrm1eq1} as required.
Also by Proposition \ref{Lya:Cor1}, we 
thus taking $C=\left[\left(1 + H\right)/{2Q G^n}  +  B \right] $, the required bound \eqref{Lya:Thrm1eq2} holds.
\Endproof

\subsubsection{Proof of Theorem \ref{Cor:1}.}
\label{finalproofthrm1}

We can now prove Theorem \ref{Cor:1}. 

\Beginproof{ of Theorem \ref{Cor:1}}
%
We notice that the bound 
$   
\sum^t_{s=\lfloor t/2^n \rfloor} \alpha_s  \frac{\kappa}{2}\geq  \alpha_0 B + F
$ can be achieved for all $t\geq T_1$ for fixed constants $T_1$ and $n$.
This holds since $\sum^t_{s=\lfloor t/2^n \rfloor} \alpha_s\rightarrow \infty $ as $t \rightarrow \infty$.
(See Lemma \eqref{lem:equationcrunching} in the E-companion for verification of this and a concrete choice of $T_1$ and $n$.)
Thus applying bound \eqref{Lya:Thrm1eq1} from Proposition \ref{Thrm1} with $T_2=\max\{T_0,T_1\}$, we see that 
\begin{align*}
\mathbb P( f(\bm x_{t+1}) - f(\bm x^\star) \geq z ) 
& \leq
e^{
- \frac{Q G^n }{\alpha_t}(z - \alpha_t B-F - \alpha_0 B+\sum_{s=\lfloor t/2^n \rfloor}^{t}\alpha_s\frac{\kappa}{2} )
} 
+
H {e^{-\frac{Q G^n}{\alpha_t}(z - \alpha_t B)}} \\
&\leq 
(1+H) {e^{-\frac{Q G^n}{\alpha_t}(z - \alpha_t B)}} \, && \text{ for } t\geq T_2 \, 
\\
&\leq 
I e^{-\frac{J}{\alpha_t} z } && \text{ for } t\geq 0 \, . 
\end{align*}
Thus we see that \eqref{Cor1:eq1} holds for $t\geq 0$ with suitable choice of $I$ and $J$ (e.g. $I=(1+H)e^{Q G/(F/\alpha_{T_2}-B)}$ and $J=QG^n$).
Integrating the bound \eqref{Cor1:eq1} then gives 
\begin{align*}
\mathbb E [ f(\bm x_{t+1})  - f(\bm x^\star) ] \leq \int_0^\infty I e^{-(J/\alpha_t) z} dz = \frac{I}{J} \alpha_t \, .
\end{align*}
\Endproof

\subsection{Proof of Theorem \ref{ThrmNew}}
\begin{darkblue}
   Typical Stochastic Gradient Descent proofs use $\|\bm x_t - \bm x^\star \|^2$ as a Lyapunov function. 
However, we want use $\|\bm x_t - \bm x^\star \|$ instead.
We start with the standard SGD drift argument and then take a square root to gain our drift condition for $\|\bm x_t - \bm x^\star \|$.
We can then apply Theorem \ref{Cor:1}, which goes through the mechanics of converting a linear Lyapunov drift condition into an exponential Lyapunov function. The idea of converting linear drift into an exponential Lyapunov function is reasonably well-known in the Markov chains analysis but currently not for SGD. Our proof shows how to adapt and apply these ideas. The basic mechanics to apply Theorem \ref{Cor:1} are the same for other SA procedures, for example, Kiefer-Wolfowitz and Stochastic Frank-Wolfe.  
\end{darkblue}


\Beginproof{ of Theorem \ref{ThrmNew}}
In this proof, we will apply Proposition \ref{Thrm1} with the choice
$
f(\bm x) := \min_{\bm x^\star \in \mathcal X^\star} \| \bm x- \bm x^\star \| \, .
$
We also define $\bm x_t^\star := \argmin_{\bm x \in \mathcal X^\star} \|\bm x_t-\bm x \|$.
Now observe that 
\begin{align}
f(\bm x_{t+1})^2
=
\left\|
	\bm x_{t+1} - \bm x_{t+1}^\star 
\right\|^2
\nonumber
\leq \, 
&
\left\|
	\bm x_{t+1} - \bm x_t^\star 
\right\|^2
=\,
\left\|
	\Pi_{\mathcal X} ( \bm x_t - \alpha_t \bm c_t ) - \Pi_{\mathcal X}(\bm x^\star_t) 
\right\|^2 
\nonumber 
\\
\leq \,
&
\left\|
	 \bm x_t - \alpha_t \bm c_t  - \bm x^\star_t 
\right\|^2
=\,
\left\|
		\bm x_t - \bm x_t^\star 
\right\|^2
-
2 \alpha_t \bm c_t^\top 
\left(
	\bm x_t - \bm x_t^\star
\right) 
+ 
\alpha_t^2
\left\|
	 \bm c_{t}
\right\|^2 \, . \label{eq:fsqrd}
\end{align}
Condition \eqref{cond:D2} implies all moments of $\| \bm c_t\|$ are uniformly bounded. In particular, suppose $\sigma$ is such that $\mathbb E [ \| \bm c_t \|^2 | \mathcal F_t ] < \sigma^2 $ for all $t$.
    On the event where
$
\| \bm x_t - \bm x^\star \| \geq \alpha_t \frac{\sigma^2}{\kappa} >0
$
then \eqref{eq:fsqrd} gives
\begin{align*}
f(\bm x_{t+1}) \leq \,
&
\left\|
	\bm x_t - \bm x_t^\star 
\right\|
\sqrt{
1- 2 \alpha_t \bm c_t^\top 
\frac{\left(
	\bm x_t - \bm x_t^\star 
\right)
}{
\left\|
	\bm x_t - \bm x_t^\star 
\right\|^2
}
+ 
\alpha_t^2
\frac{ 
\left\|
	\bm c_t
\right\|^2
}{
\left\|
	\bm x_t - \bm x_t^\star 
\right\|^2
}
}
\nonumber 
\\
\leq \,
&
\left\|
	\bm x_t - \bm x_t^\star 
\right\|
\left(
1-  \alpha_t \bm c_t^\top 
\frac{\left(
	\bm x_t - \bm x_t^\star 
\right)
}{
\left\|
	\bm x_t - \bm x_t^\star 
\right\|^2
}
+ 
\frac{\alpha_t^2}{2}
\frac{ 
\left\|
	\bm c_t
\right\|^2
}{
\left\|
	\bm x_t - \bm x_t^\star 
\right\|^2
}
\right)\, .
\end{align*}
Above the first inequality follows from \eqref{eq:fsqrd} and in the second inequality we note that $\sqrt{1+x} \leq 1+\frac{x}{2}$.
Taking expectations on both sides shows that, on the event $\{ \| \bm x_t - \bm x^\star \| \geq \alpha_t \frac{\sigma^2}{\kappa} \}$, it holds that
\begin{align*}
\mathbb E \left[
	\left\|
	\bm x_{t+1} - \bm x_{t+1}^\star 
\right\|
	| \mathcal F_t
\right]
\leq \,
&
\left\|
	\bm x_t - \bm x_t^\star 
\right\|
-  \alpha_t \frac{\nabla l(\bm x_t)^\top 
\left(
	\bm x_t - \bm x_t^\star 
\right)
}{
\left\|
	\bm x_t - \bm x_t^\star 
\right\|
}
+ 
\frac{\alpha_t^2}{2}
\frac{ 
\mathbb E [ 
		\left\|
	\bm c_t
\right\|^2
| \mathcal F_t
]
}{
\left\|
	\bm x_t - \bm x_t^\star 
\right\|
}
\nonumber 
\leq \,
\left\|
	\bm x_t - \bm x_t^\star 
\right\|
- \alpha_t \kappa + \alpha_t \frac{\kappa}{2}
\end{align*}
Or in other words
$
\mathbb E [ f(\bm{x}_{t+1}) - f(\bm{x}_t) |\mathcal F_t ] \leq -  \alpha_t \frac{\kappa}{2} 
$
whenever $f(\bm{x}_t) -f(\bm{x}^\star) \geq \alpha_t \frac{\sigma^2}{\kappa} $. Thus we see that Condition \eqref{fcond:1} holds. 

We now verify Condition \eqref{fcond:2}. 
Projections reduced distances, specifically, if $\| \bm x_t - \bm x_t^\star \| \leq \| \bm x_{t+1} - \bm x_{t+1}^\star \| $ then
\[ 
f(\bm x_{t+1})
-
f(\bm x_t)
=
\| \bm x_{t+1} - \bm x_{t+1}^\star \|
-
\| \bm x_t - \bm x_t^\star \| 
\leq 
\| \bm x_{t+1} - \bm x_{t}^\star \|
-
\| \bm x_t - \bm x_t^\star \|
\leq \| \bm x_{t+1} - \bm x_{t}   \| 
= \alpha_t \| \bm c_t \|
\]
(The analogous argument follows if $\| \bm x_{t+1} - \bm x_{t+1}^\star \| \leq \| \bm x_{t} - \bm x_{t}^\star \| $.)
As discussed in Section \ref{sec:PSGDresult}, the MGF condition on $\| \bm c_t\|$ now implies \eqref{fcond:2}. (See Lemma \ref{lem:fcond2} for a proof.)



We can now apply Theorem \ref{Cor:1} which gives:\begin{align*}
\mathbb P \Big(
\min_{\bm x \in \mathcal X^\star}
	 	\left\|
	\bm x_{t+1} - \bm x
\right\| \geq z
\Big)
\leq \hat I e^{-\frac{\hat J}{\alpha_t} z }
\quad \text{and}\quad 
\mathbb E 
\left[
\min_{\bm x \in \mathcal X^\star}
	 	\left\|
	\bm x_{t+1} - \bm x
\right\|
\right] 
\leq {\hat K}\alpha_t
\end{align*}
for constants $\hat I$, $\hat J$ and $\hat K$. 
Since we also assume in addition that $l: \mathcal X \rightarrow \mathbb R$ is Lispchitz continuous (with Lipschitz constant $\hat L/\hat K$) we have, as required, 
\begin{align*}
\mathbb E \left[
	l(\bm x_{t+1}) - \min_{\bm x \in \mathcal X} l(\bm x)
\right]
\leq 
\frac{\hat L}{\hat K}
\mathbb E \left[
\min_{\bm x \in \mathcal X^\star}
	 	\left\|
	\bm x_{t+1} - \bm x
\right\|
\right] 
\leq {\hat L}\alpha_t \, .
\end{align*}

\Endproof

\section{Applications and Numerical Examples}\label{sec:LP}
\begin{darkblue}
\textbf{A Circle Constraint.}
We consider minimizing an objective function $l(x) = ||x - x^*||$ for $x^* = (7, 7)$ over a large circle with center $(0, 0)$ and radius $15$. The Frank-Wolfe, PSGD and Kiefer-Wolfowitz algorithms are applied. For the Kiefer-Wolfowitz, we assume the objective function is observed with noise following $N(0, 0.01)$. The optimum is in the interior. No projection is required in our simulation. Figure \ref{fig:Circle} shows the constraint set and the rate $O(1/t)$ with $\gamma=1$ for the three algorithms. 


\begin{figure}[ht]
\centering

\begin{subfigure}{.45\textwidth}
  \centering
  \includegraphics[width=1.0\linewidth]{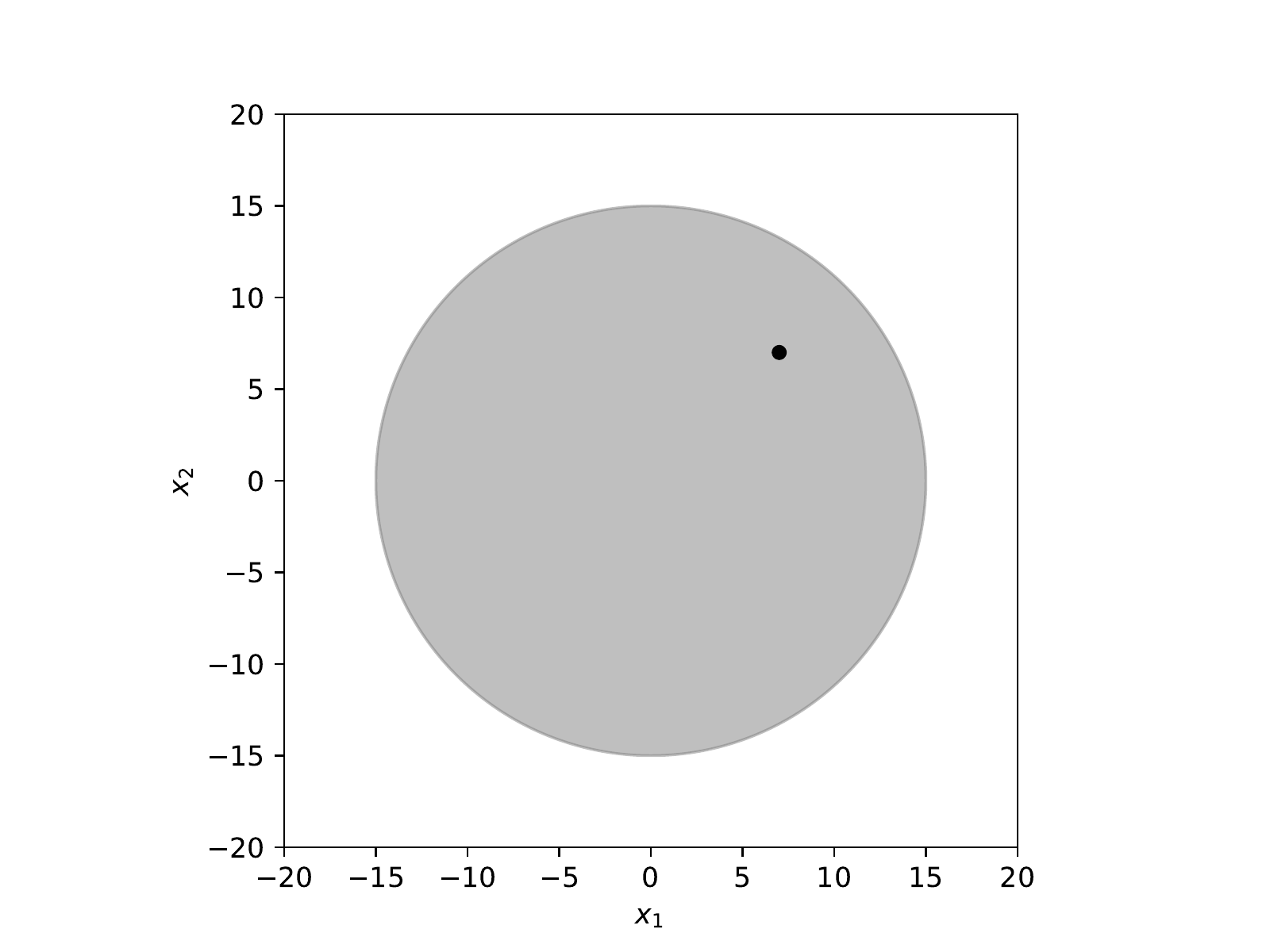}
  \caption{Circle Constraint}
  \label{fig:CircleConstraints}
\end{subfigure}%
\begin{subfigure}{.45\textwidth}
  \centering
  \includegraphics[width=1.0\linewidth]{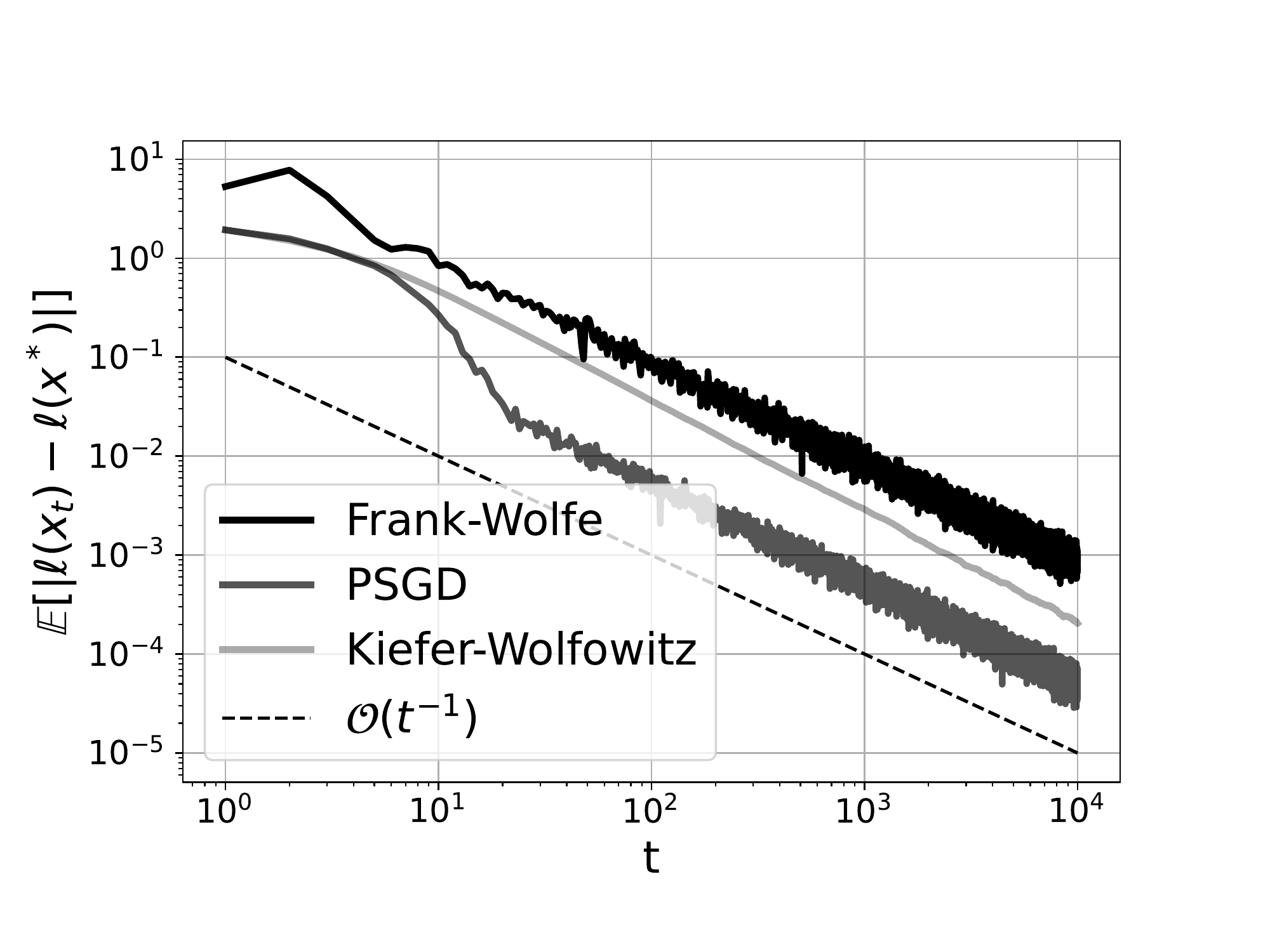} \caption{Convergence of Algorithms}
  \label{fig:CircleEp}
\end{subfigure}
\caption{Convergence of Frank-Wolfe, PSGD, and Kiefer-Wolfowitz algorithms on the circle constraint example. Figure \ref{fig:CircleConstraints}: the black dot is the optimal solution $(7, 7)$. Figure \ref{fig:CircleEp}: The expectation is computed over 20 realizations. The stochastic gradients for Frank-Wolfe and PSGD are computed with batch size $B=10$. The parameter $v=0.8$ is chosen for Kiefer-Wolfowitz. The parameters of step size are chosen as $a=0.9, u=1$ and $\gamma=1$ such that $\alpha_t = 1/(1+t)$.  The fitted slope is $-1.00$, $-1.00$ and $-1.10$ for Frank-Wolfe, PSGD and Kiefer-Wolfowitz.}
\label{fig:Circle}
\end{figure}

\noindent \textbf{Three Spherical Constraints.}
\cite{davis2023asymptotic} consider a normal approximation on PSGD with two spherical constraints. We add a third spherical constraint. Specifically, we minimize the objective $l(x) = -x_1 + \hat{w}_ix_i,$ $ \hat{w}_i \sim N(0, 1)$ for $i=1,2,3$ over the intersection of spheres with center $(1, 0, 0), (-1, 0, 0)$ and $(0, 1, 0)$, and radius $2$ using PSGD and Kiefer-Wolfowitz algorithm. See Figure \ref{fig:ThreeBalls} for the constraint set. The optimal solution is taken at $(0, 0, \sqrt{3})$. Bregman's cyclic algorithm is applied for projection. Rather than $O(1/\sqrt{t})$, Figure \ref{fig:ThreeBallsAlg} shows the rate $O(1/t)$ with $\gamma=1$ for both the PSGD and the Kiefer-Wolfowitz algorithm. Further, we study PSGD and Kiefer-Wolfowitz convergence when halving the step size every $T=20$ iterations. As suggested in Section \ref{sec:Linear}, linear convergence occurs. See Figure \ref{fig:HalfAlpha}. 


\begin{figure}[ht]
\centering

\begin{subfigure}{.45\textwidth}
    
\end{subfigure}

\begin{subfigure}{.45\textwidth}
  \centering
  \includegraphics[width=1.0\linewidth]{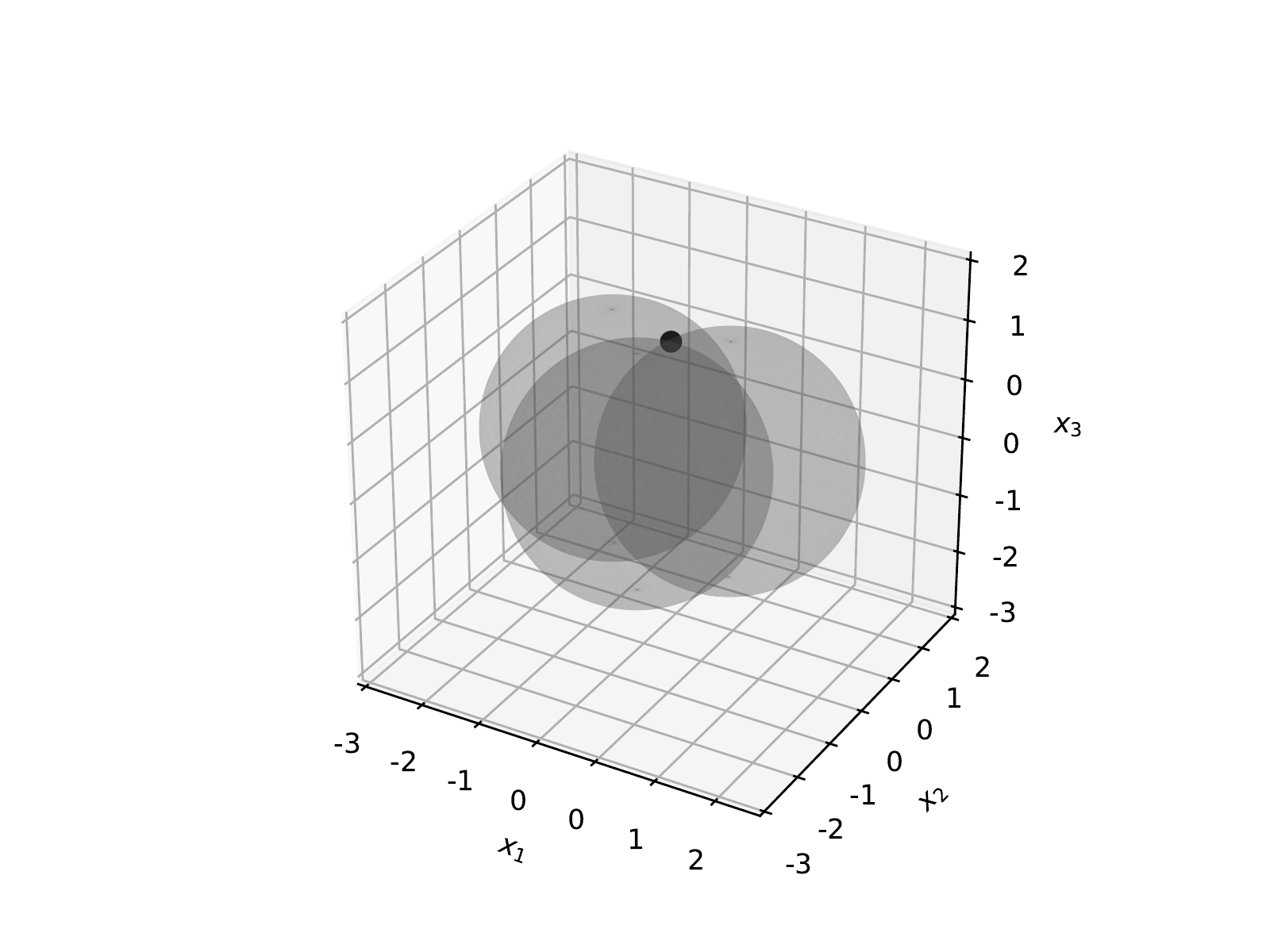}
  \caption{Three Spherical Constraints}
  \label{fig:ThreeBalls}
\end{subfigure}%
\begin{subfigure}{.45\textwidth}
  \centering
  \includegraphics[width=1.0\linewidth]{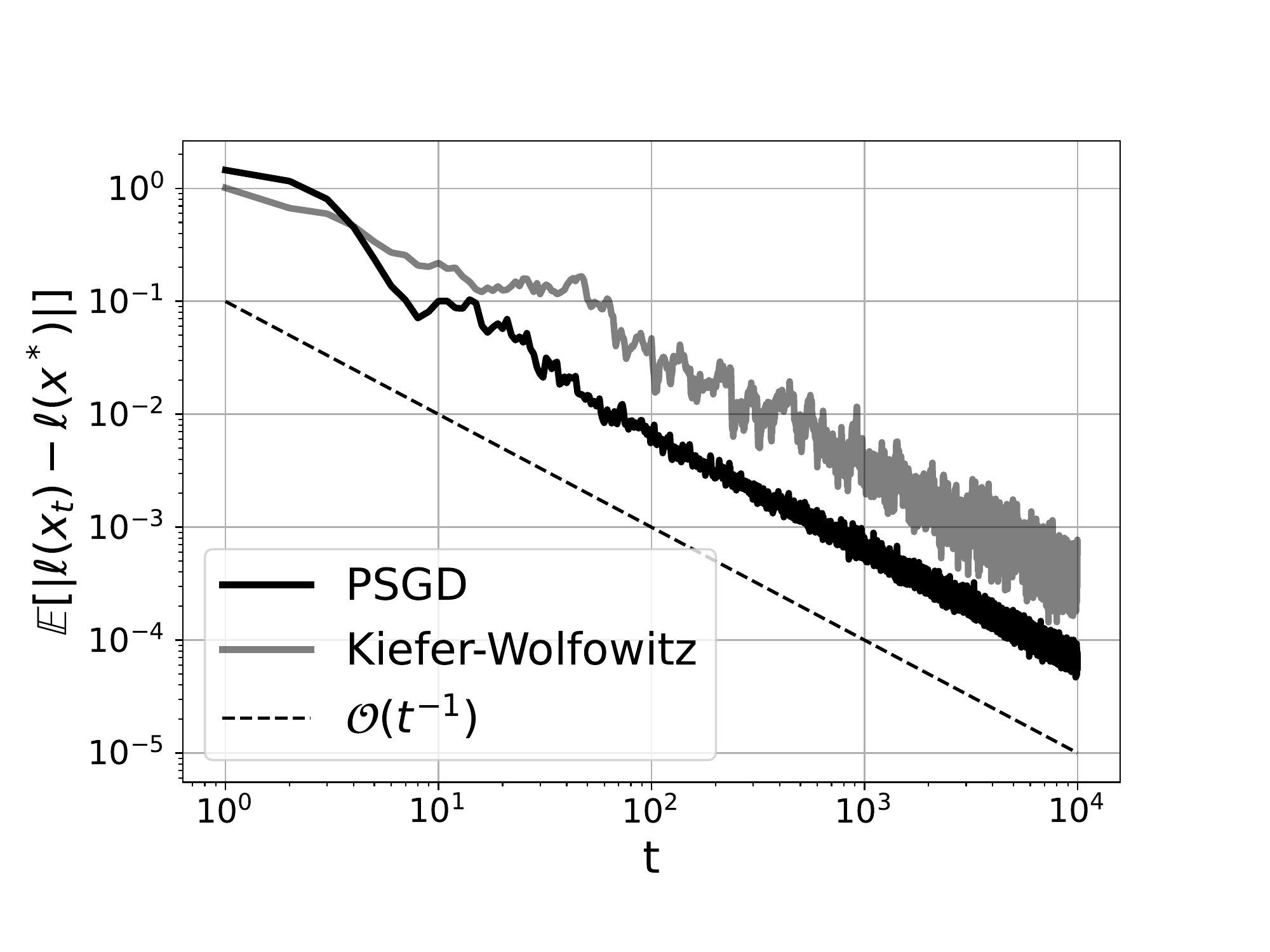} \caption{Convergence of Algorithms}
  \label{fig:ThreeBallsAlg}
\end{subfigure}
\caption{Convergence of PSGD and Kiefer-Wolfowitz algorithms on the three spherical constraints problems. Figure \ref{fig:ThreeBalls}: the black dot is the optimal solution $(0, 0, \sqrt{3})$. Figure \ref{fig:ThreeBallsAlg}: The expectation is computed over 20 realizations. The stochastic gradients for PSGD are computed with $B=10$. The parameter $v = 1$ is chosen for Kiefer-Wolfowitz. The parameters of step size are chosen as $a=1, u=1$ and $\gamma=1$ such that $\alpha_t = 1/(1+t)$. The fitted slope is -1.01 and -1.00 for PSGD and Kiefer-Wolfowitz.}
\label{fig:Balls}
\end{figure}

\begin{figure}[ht]
    \centering
    \includegraphics[width=.5\linewidth]{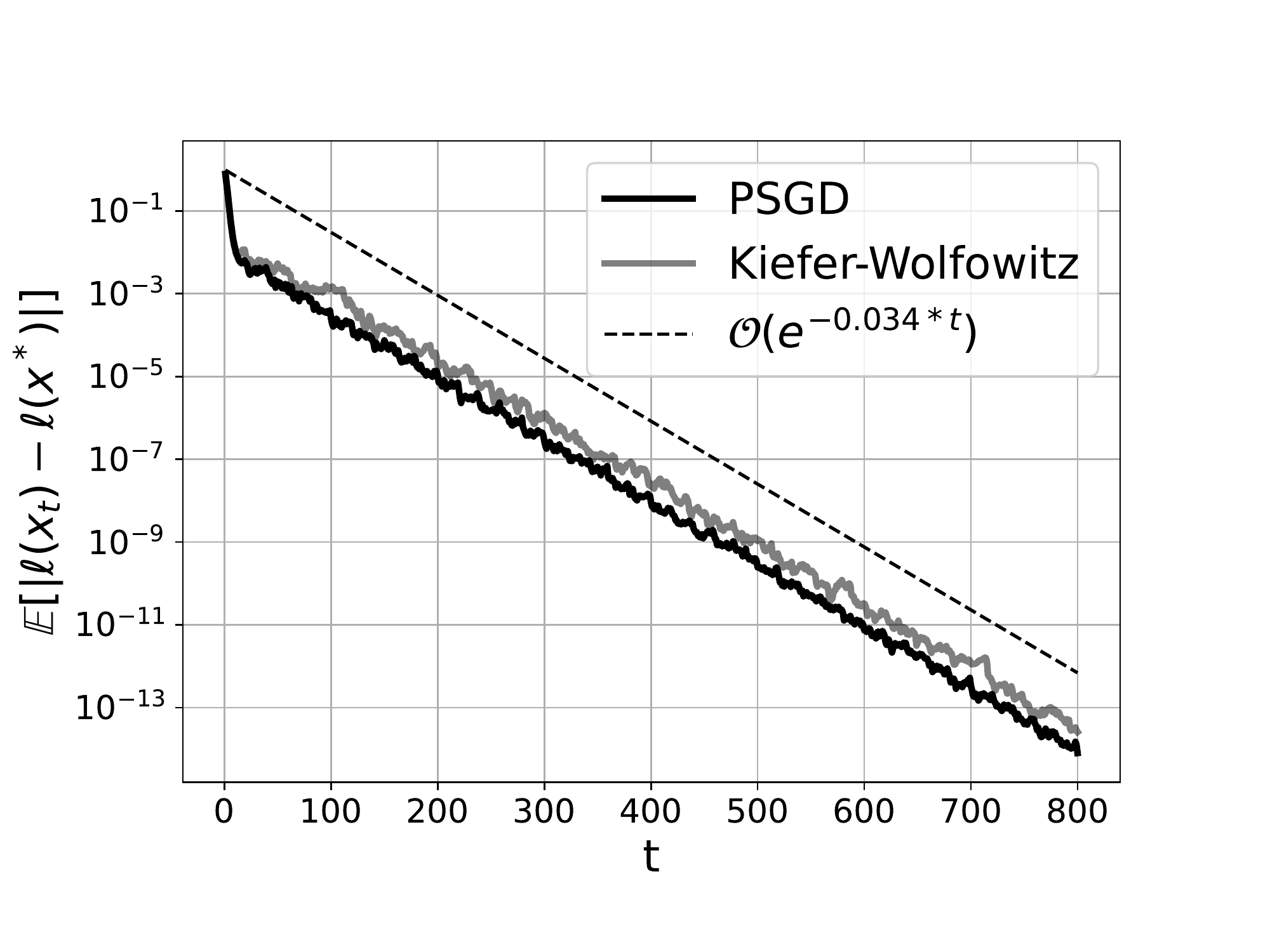}
    \caption{Linear convergence of PSGD and Kiefer-Wolfowitz for the three spherical constraints problem. The expectation is computed over 20 realizations. The stochastic gradients are computed with $B=10$. The parameter $v=10$ is chosen for Kiefer-Wolfowitz. The simulations are conducted with learning rates divided by 2 every 20 steps.}
    \label{fig:HalfAlpha}
\end{figure}

\noindent \textbf{Non-Negative Ridge Regression.} \cite{duchi2021asymptotic} consider the normal approximation with constraints for a non-negative least square problem and for ridge regression. We consider non-negative ridge regression and find that the normal approximation is no longer valid. We apply $l(x) = \frac{1}{2}|a^Tx - b|^2$ as the objective with a constraint set $\{x \in \mathbb{R}_{+}^2: \|x\| \leq \sqrt{0.9}\}$, where $a$ and $b$ are observed with $b_i \sim a_i x_+ + \xi_i$ for $x_+ = (1, -1)$, $a_i \sim N(0, \mathbb{I}_2)$ and $\xi_i \sim N(0,1)$. The optimal solution is taken at $(\sqrt{0.9}, 0).$ Figure \ref{fig:NLS} shows the rate $O(1/t)$ with $\gamma=1$ for the PSGD and Kiefer-Wolfowitz.

\begin{figure}[ht]
    \centering
    \includegraphics[width=.5\linewidth]{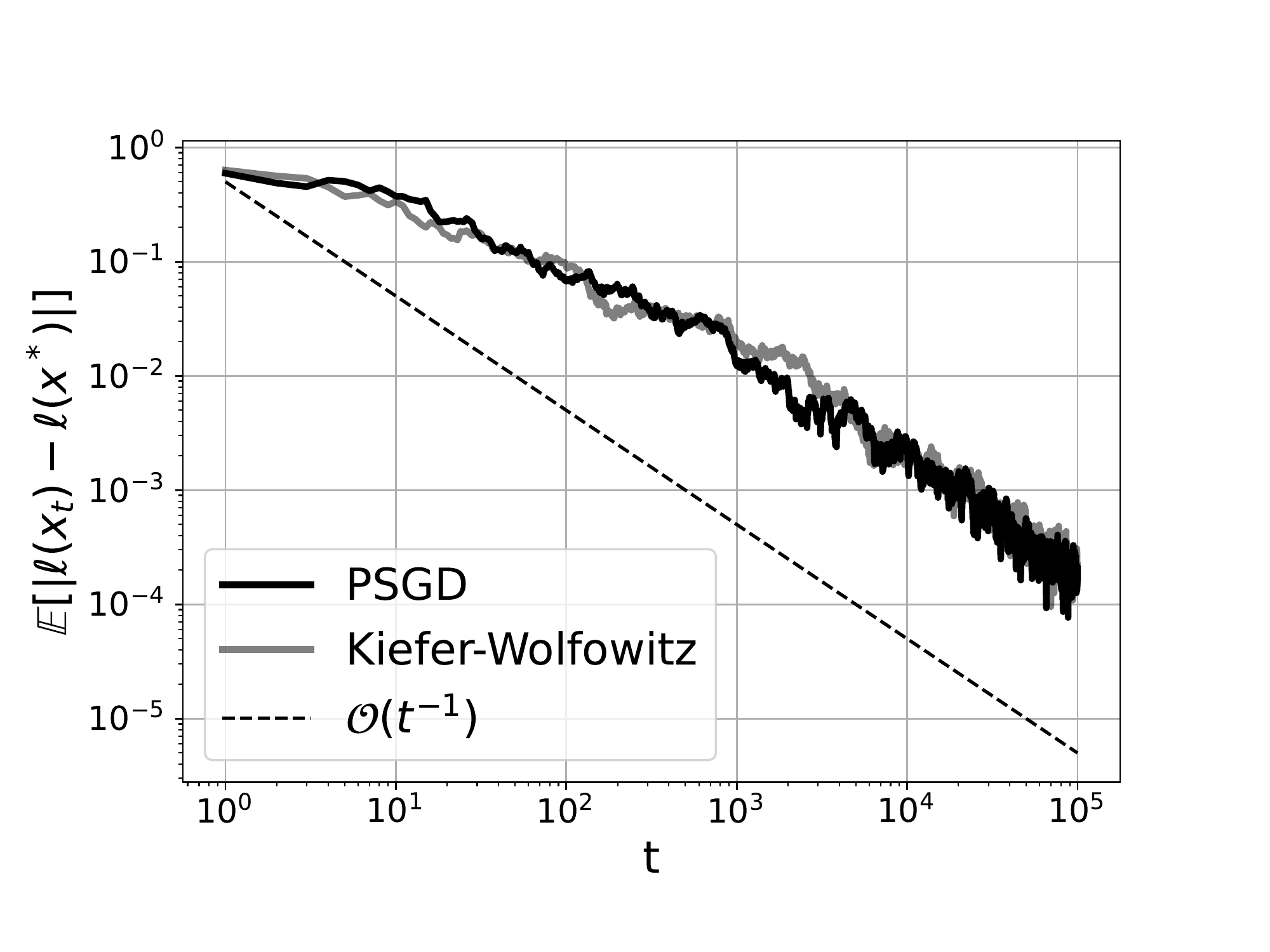}
    \caption{Non-negative least square example. The expectation is computation over 20 realizations. The parameter $v=1$ is chosen for Kiefer-Wolfowitz. The parameters of step size are chosen as $a=1$, $u=1$ and $\gamma=1$ such that $\alpha_t = 1/(1+t).$ The fitted slope is $-1.00$ and $-1.00$ for PSGD and Kiefer-Wolfowitz.\label{fig:NLS}}
\end{figure}

\noindent \textbf{Linear Programs and Markov Decision Processes.} All the theoretical results and simulations so far consider non-linear constraints and objectives. We limit all discussion of linear programs to this paragraph and the E-companion. First, all linear programs are sharp (see Lemma \ref{LemmaA} in the E-companion for proof). Second, linearly convergent (and parallelizable) algorithms can calculate projections onto linear constraints. Hildrecht's Projection algorithm is a first-order algorithm that converges linearity, see \cite{iusem1990convergence}. Further \cite{dos1987parallel} shows the algorithm can be parallelized and adapted to non-linear constraints. Proved in the results of
Section \ref{sec:Linear}, Figure \ref{fig:LinearConLP} demonstrates linear convergence of PSGD on a linear program with unbounded constraints. 
 Markov Decision Processes can be expressed as linear programs, with PSGD on the dual corresponding to a simple policy gradient algorithm. Given this, we solve a general three-state, two-action MDP, and Blackjack (See, \cite{sutton2018reinforcement}). More detailed discussions can be found in Section \ref{sec:APPLP} of the E-companion.

\begin{figure}[ht]
\begin{subfigure}{.45\textwidth}
  \centering
  \includegraphics[width=0.7\linewidth]{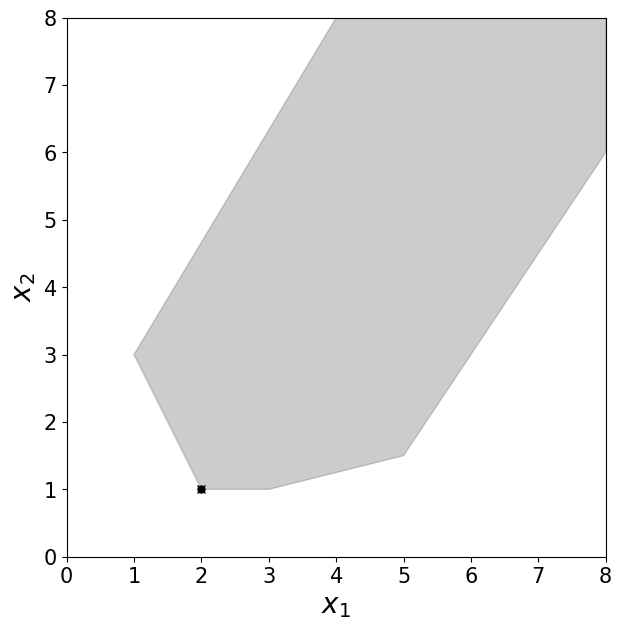}
  \caption{Linear Constraints}
  \label{fig:LP_tope}
\end{subfigure}%
\begin{subfigure}{.45\textwidth}
  \centering
  \includegraphics[width=1.0\linewidth]{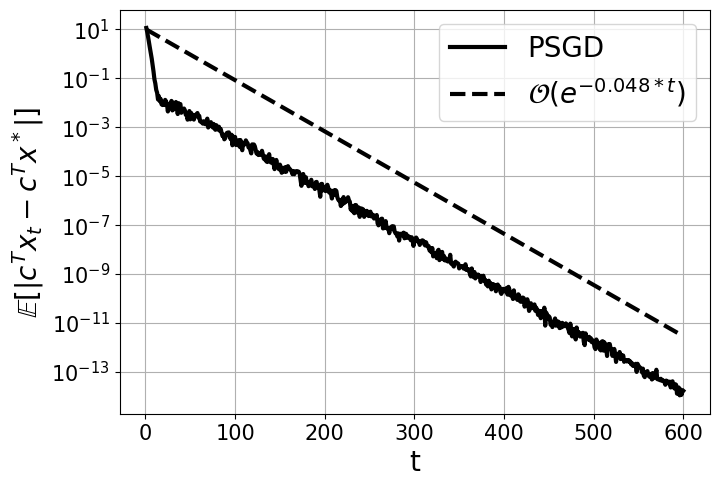}
  \caption{Linear convergence of PSGD}
  \label{fig:LP_Linear}
\end{subfigure}
\caption{Linear convergence of PSGD. Costs are normally distributed with mean $(4,6)$ and covariance $(25, 0; 0, 25)$. The simulation is conducted with learning rates divided by 1.1 every 2 steps.}
\label{fig:LinearConLP}
\end{figure}
    
\end{darkblue}

\section{Conclusions}\label{sec:conclusions}
Motivated by results on the exponential distributions found for queueing networks, we have established convergence rates for constrained stochastic approximation algorithms. Our results extend the findings on Markov chains by \cite{hajek1982hitting} to stochastic approximation. To the best of our knowledge, these techniques from the theory of Markov chains have not been applied in stochastic approximation. 

 Asymptotic normality is classical in stochastic approximation, whereas exponential concentration is not well understood. This paper identifies situations where the asymptotically optimal solution is not Gaussian and provides methods to establish bounds when exponential concentration holds. Our results prove that a potential benefit is faster convergence.

\noindent \textbf{Acknowledgement:} We thank the reviewers for their help in improving this paper. We are particularly grateful to a referee for the reference, \cite{davis2019stochastic}, which lead to Theorem \ref{ThrmLinear}. Also, to Stavros Zenios for advice on projection algorithms, and Matthias Troffaes for help with pycddlib.

\bibliographystyle{informs2014} 
\bibliography{mybib} 
\appendix



\section{Appendix to Introduction: Exponential Limit for Uniform MLE}
\label{sec:uniformMLE}

For $X_1,...,X_n\sim U[0,\theta]$, $\theta>0$, the joint density is 
\begin{equation*}
    \prod_{i=1}^n \frac{1}{\theta} \mathbb I \Big[ 0\leq  x_i \leq \theta \Big] =  \frac{1}{\theta^n} \mathbb I \Big[ 0 \leq \min_{i=1,...,n} x_i \leq \max_{i=1,...,n} x_i \leq \theta \Big]
\end{equation*}
From the above, we see that the maximum likelihood is given by
$
\hat \theta_n =\max_{i=1,...,n} X_i 
$.
For a normal approximation, we would typically analyse $\sqrt{n} ( \hat \theta_n - \theta)$. However, instead we analyse a more concentrated asymptotic $n ( \theta - \hat \theta_n)$. For this observe
\begin{align*}
\mathbb P( n(\theta - \hat \theta_n) \geq z )
& =
\mathbb P\Big( \max_{i=1,...,n} X_i \leq \theta - \frac{z}{n} \Big)
=  \Big( 1- \frac{z}{n\theta}\Big)^n \xrightarrow[n\rightarrow\infty]{} e^{-\frac{z}{\theta}}
\end{align*}
From the above, we see that 
\begin{equation*}
    n(\theta - \hat \theta_n) \xrightarrow[n \rightarrow \infty]{\mathcal D} \exp(\theta^{-1})
\end{equation*}
So the limit here is not normally distributed under a $\sqrt{n}$ normalization but is order $n$ and is exponentially distributed.

\section{Appendix to Section \ref{sec:Main}: Main Results}

\subsection{Appendix to Section \ref{sec:PSGDresult}}
    
\subsubsection{Sub-exponential noise (\ref{cond:D2}) implies Condition   (\ref{fcond:1})}\label{sec:subExp}

\begin{darkblue}
\begin{lemma}\label{lem:fcond2}
    For a  Lispchitz continuous function $f$,  if Condition \eqref{cond:D2} holds, that is
    \begin{equation*}
        \sup_{t\geq 0} \mathbb E\big[e^{\lambda \| c_t\|} | \mathcal F_t \big] < \infty  
    \end{equation*}

    then Condition \eqref{fcond:2} holds that is
        \begin{align} \label{eq:mgfcond}
        \Big[ |f(\bm x_{t+1}) - f(\bm x_t)|  \big| \mathcal F_t \Big]
        \leq \alpha_t Y \, ,\quad \text{with} \quad \mathbb E [ e^{\eta Y } ] < \infty
        \end{align}
for some $\eta>0$.
\end{lemma}

\Beginproof{}
    Since $f(x)$ is Lipschitz
    \begin{align*}
| f(\bm x_{t+1}) - f(\bm x_t) | 
\leq  &
K \| \bm x_{t+1} - \bm x_t \| 
\leq 
\alpha_t K || \bm c_t || \, .
\end{align*}
Since Condition \eqref{cond:D2} holds, we take $M \geq \sup_t \mathbb E \big[
		e^{  \lambda \left\|
	\bm c_t 
\right\|
}
\big|
\mathcal F_t 
\big] $. We let $Y$ be the random variable with CCDF:
$
 \mathbb P ( Y  \geq y ) = 1 \wedge ( Me^{-\frac{\lambda}{K} y} ) .
$
Thus for $y \in \mathbb R_+$
\begin{align*}
\mathbb P
\left(
	\left|
	f(\bm x_{t+1} ) - f (\bm x_t) 
\right|
\geq \alpha_t y 
\Big|
\mathcal F_t 
\right)
\notag 
& \leq 
\mathbb P
\left(
	\| \bm c_t \|
\geq y/K 
\Big|
\mathcal F_t 
\right) 
\\
& \leq \,
\min \big\{ 1, e^{-(\lambda/K) y }\mathbb E \big[
		e^{  \lambda \left\|
	\bm c_t 
\right\|
}
\big|
\mathcal F_t 
\big]\big\} 
\leq \,
\mathbb P (  Y\geq y ) 
\end{align*}
Above, we apply a Chernoff bound.
From this inequality above, we see that Condition \eqref{fcond:2} follows from Condition \eqref{cond:D2}.
\Endproof
\end{darkblue}

\subsubsection{Proof of Lemma \ref{lem:EqivSharpness}: sharpness is equivalent to non-vanishing gradient for convex functions.} \label{sec:SharpProof}
\begin{darkblue}
We now prove Lemma \ref{lem:EqivSharpness}.

    \lemsharpe*

    \Beginproof{}
        First let's assume condition \eqref{cond:D1} holds. 
        Let $\bm x(t) = \bm x^\star + (1-t)( \bm x-\bm x^\star)$.
        Thus we have
            \begin{align*}
                l(\bm x) - l(\bm x^\star) 
                &
                = 
                \int_0^1 \frac{d l(\bm x(t))}{dt} dt
                \\
                &
                =
                \int_0^1 \nabla l(\bm x(t)) ( \bm x(t) - \bm x^\star) dt
                \\
                &
                \geq
                \int_0^1 \kappa \| \bm x(t) - \bm x^\star \| dt 
                \\
                &
                =
                \int_0^1 (1-t)\kappa \| \bm x - \bm x^\star \| dt 
                \\
                &
                =
                \frac{\kappa}{2} \| \bm x - \bm x^\star \| 
            \end{align*}
        The first equality follows since absolute continuity implies the fundamental theorem of calculus holds. The second equality holds by the chain rule. The third equality follows by the gradient condition \eqref{cond:D1}. We then apply the definition of $\bm x(t)$ and integrate. 
        Thus as required, we see that condition \eqref{cond:D1} implies \eqref{cond:Sharp}.

        If we also suppose that the function $l(\bm x)$ is convex and that \eqref{cond:Sharp} holds then
        \[
            l(\bm x^\star) - l(\bm x) \geq \nabla l (\bm x) ( \bm x^\star - \bm x) \,.
        \]
        So 
        \[
           \nabla l(\bm x) ( \bm x - \bm x^\star ) \geq l(\bm x) - l(\bm x^\star) \geq \kappa' \|\bm x - \bm x^\star \|   
        \]
        The first inequality rearranges the convexity definition above. The second inequality applies the Sharp Condition \eqref{cond:Sharp}.
        So we see, as required, for a convex function, the Sharp Condition \eqref{cond:Sharp} implies the gradient condition \eqref{cond:D1}. 
    \Endproof    
\end{darkblue}

\begin{darkblue}
\subsubsection{Proof of Lemma \ref{lem:functional}: with $d$ or more active constraints, SGD is not normally distributed.}

The following Lemma emphasizes that the setting of \cite{duchi2021asymptotic} is designed for smooth problems where the number of active constraints is smaller than the problem dimension. 
Once the number of active constraints exceeds the dimension of the problem, then the normal approximation no longer holds. With the Lemma below, we can say that the limiting distribution has an exponential concentration for PSGD. Calculations can speculate the form of the asymptotic optimality; however, the general theory of asymptotic optimality of stochastic optimization is incomplete, particularly in settings where the normal approximation is invalid. As we indicate, it requires a better understanding of asymptotic optimality in the presence of Sharpness. 

\lemfunctional*
\Beginproof{}
Let $x_\infty$ be such that $\| \bm x \|_\infty < x_\infty$ for all $ x \in \mathcal X$.
Let $\bm c = \nabla l (\bm x^\star)$.
    Let $\bm c_i = \nabla l_i(\bm x^\star)$
    and $b_i = \bm c_i^\top \bm x^\star$.
Let $\mathcal P
= \{ \bm x : \bm c_i^\top \bm x \geq b_i,\, i=1,...,d, \| \bm x \|_\infty \leq x_\infty \}
$. Notice that by convexity 
\begin{equation}\label{lem:1XP}
    \mathcal X \subseteq \mathcal P.
\end{equation} 
Notice by linear independence $\bm x^\star$ is the unique point such that $\bm c_i^\top \bm x^\star = b_i$, $i=1,...,d$. 
That is $\bm x^\star$ is an extreme point of the polytope $\mathcal P$. Since $-\nabla l(\bm x^\star) \in \textrm{relint }
\mathcal N_{\mathcal X}(\bm x^\star)$, $\bm x^\star$ miniminizes $\bm c^\top \bm x$ over $\bm x \in \mathcal P$.
Thus by Lemma \ref{LemmaA}
\begin{equation}\label{lem:2XP}
 \bm c^\top (\bm x - \bm x^\star)  \geq K \|\bm c\| \|\bm x - \bm x^\star \| , \qquad \forall \bm x \in \mathcal P \, .    
\end{equation}
By convexity
\begin{equation}\label{lem:3XP}
    l(\bm x) - l(\bm x^\star) \geq \bm c^\top (\bm x -\bm x^\star) 
\end{equation}
combining \eqref{lem:1XP}, \eqref{lem:2XP} and \eqref{lem:3XP} we see that 
\[
l(\bm x) - l(\bm x^\star) \geq K \|\bm c \| \|\bm x - \bm x^\star \|\, , \qquad \forall \bm x \in \mathcal X\, .
\]
Thus we see that the function $l(\bm x)$ is sharp on $\mathcal X$. Condition \eqref{cond:D1} then follows by Lemma \ref{lem:EqivSharpness} since the function $l(\bm x)$ is convex. 
\Endproof
\end{darkblue}

\subsection{Finite number of Projections for Interior Optimum.}
\begin{darkblue}
We say a projection step is trivial if $x \in \mathcal X$ and thus $\Pi_{\mathcal X}(x) =x $. Otherwise, we say the projection at $x$ is non-trivial. We can show that in instances where the optimum is in the interior only a finite number of non-trivial projections are required.
\begin{proposition}\label{PropProj}
Under the assumptions of Theorem \ref{ThrmNew},
    if $\mathcal X^\star$ belongs to the interior of $\mathcal X$, then the number of (non-trivial) projection steps required by Projected Stochastic Gradient Descent is finite and bounded in expectation.
\end{proposition}
\Beginproof{}
Let the random variable $N$ denote the number of non-trivial projections. By Theorem \ref{ThrmNew}, we have that 
\[
\mathbb P
\left(
 \min_{\bm x^\star \in \mathcal X^\star} \left\|
	\bm x_{t+1} - \bm x^\star 
\right\|
\geq 
z
\right)
\leq J e^{-I t^\gamma z}\, .
\]
We let $\tilde z$ be the distance from the set of optima to the boundary of $\mathcal X$, that is,
\[
\tilde z = \min_{\bm x\in \mathcal X^\star, \bm y \notin \mathcal X} \| \bm x^\star - \bm y \|\, .
\]
Since $\mathcal X^\star$ belongs to the interior of $\mathcal X$, we have that $\tilde z >0$.
Note that if a projection is non-trivial then $\min_{\bm x^\star \in \mathcal X^\star} ||\bm x -\bm x^\star || \geq \tilde z$. Thus
\[
N \leq \sum_{t=0}^\infty \mathbb I \Big[ \min_{\bm x^\star \in \mathcal X^\star} \| \bm x_{t+1} - \bm x^\star \| \geq \tilde z\Big]
\]
and so, as required,
\begin{align*}
    \mathbb E [ N ] 
\leq \sum_{t=0}^\infty 
    \mathbb P \left(
        \min_{\bm x^\star \in \mathcal X^\star} \| \bm x_{t+1} - \bm x^\star \| \geq \tilde z 
        \right) 
        \leq \sum_{t=0}^\infty J e^{-I t^\gamma z} < \infty  \, .
\end{align*}
\Endproof 
\end{darkblue}

\begin{darkblue}
\subsection{Kiefer-Wolfowitz: Proof of Theorem \ref{ThrmKW}} \label{sec:KWPRoof}

We now restate and prove Theorem \ref{ThrmKW}.

\kieferwolfowitz*
\Beginproof{}
The proof here combines the proof ideas for Kiefer-Wolfowitz, see \cite{fabian1967stochastic} (or, more recently, \cite{broadie2011general}), with the proof in Theorem \ref{ThrmNew}. As with the proof of Theorem \ref{ThrmNew} our goal is to verify Conditions \ref{fcond:1} and \ref{fcond:2}, so that we can apply Theorem \ref{Thrm1}.

We can write the KW recursion as 
\begin{align}
    \bm y_{t+1 } 
    & =
    \bm x_t - \alpha_t \nabla l(\bm x_t) 
    + \alpha_t \bm \delta_t 
    + \alpha_t \bm \epsilon_t 
    \\
    \bm x_{t+1}
    & = \Pi_{\mathcal X} ( \bm y_{t+1} )
\end{align}
where 
\begin{align*}
     \bm \delta_t & = 
     \nabla l(\bm x) 
     -
     \frac{\bm l(\bm x_t+\bm \nu_t) - \bm l(\bm x_t - \bm \nu_t)  }{2 \nu_t} \\
     \bm \epsilon_t & = 
     \frac{\bm l(\bm x_t+\bm \nu_t) - \bm l(\bm x_t - \bm \nu_t)  }{2 \nu_t}-
     \frac{\bm l(\bm x_t+\bm \nu_t, \hat w_t^+) - \bm l(\bm x_t - \bm \nu_t, \hat w_t^-)  }{2 \nu_t}\, .
\end{align*}
Letting $\bm x_{t}^\star$ be the projection of $\bm x_t$ onto $\mathcal X^\star$,
then 
\begin{align*}
    \|
        \bm x_{t+1} - \bm x_{t+1}^\star
    \|^2
    & 
    \leq
    \|
        \bm x_{t+1} - \bm x_{t}^\star
    \|^2
    \\
    &
    \leq 
    \| \bm y_{t+1} - \bm x^\star_t \|^2 =
    \| \bm y_{t+1} - \bm x_t + \bm x_t - \bm x^\star_t \|^2 
    \\
    &
    =
    \|\bm x_t - \bm x^\star_t \|^2 
    - 2 \alpha_t \nabla l (\bm x_t)^\top (\bm x_t - \bm x^\star_t )
    + 2 \alpha_t \bm \delta^\top_t (\bm x_t - \bm x^\star_t) 
    + 2 \alpha_t \bm \epsilon^\top_t (\bm x_t - \bm x^\star_t) 
    +\| \bm y_{t+1} - \bm x_t \|^2 \, .
\end{align*}
The first inequality above follows since $\bm x^\star_{t+1}$ is a projection. The second follows since $\bm x_{t+1}$ is a projection. We then expand.

We let $E_t$ be the positive number defined below in \eqref{eq:ETdef}. On the event 
$
\{ \|\bm x_t - \bm x_t^\star \| \geq E_t \,\}  $, we have
\begin{align}
    & \| \bm x_{t+1} - \bm x_{t+1}^\star \| 
    \nonumber \\
    \leq & 
    \| \bm x_{t} - \bm x_{t}^\star \|
    \sqrt{
        1 - 2 \alpha_t \nabla l (\bm x_t)^\top 
        \frac{(\bm x_t - \bm x^\star_t )}{\| \bm x_{t} - \bm x_{t}^\star \|^2}
    + 2 \alpha_t \bm \delta_t^\top \frac{(\bm x_t - \bm x^\star_t )}{\| \bm x_{t} - \bm x_{t}^\star \|^2}
    + 2 \alpha_t \bm \epsilon_t^\top \frac{(\bm x_t - \bm x^\star_t )}{\| \bm x_{t} - \bm x_{t}^\star \|^2}
    +\frac{\| \bm y_{t+1} - \bm x_t \|^2}{\| \bm x_{t} - \bm x_{t}^\star \|^2}
    }
    \nonumber 
    \\
    \leq & \| \bm x_{t} - \bm x_{t}^\star \|     \nonumber 
    \\
     &-  \alpha_t \nabla l (\bm x_t)^\top 
        \frac{(\bm x_t - \bm x^\star_t )}{\| \bm x_{t} - \bm x_{t}^\star \|}
        \label{eq:KW1}
        \\
    &+  \alpha_t \bm \delta_t^\top \frac{(\bm x_t - \bm x^\star_t )}{\| \bm x_{t} - \bm x_{t}^\star \|}
    \label{eq:KW2}
    \\
    &+  \alpha_t \bm \epsilon_t^\top \frac{(\bm x_t - \bm x^\star_t )}{\| \bm x_{t} - \bm x_{t}^\star \|}
    \label{eq:KW3}
    \\
    &+\frac{\| \bm y_{t+1} - \bm x_t \|^2}{2\| \bm x_{t} - \bm x_{t}^\star \|} \, .
    \label{eq:KW4}
\end{align}
 We now analyse the conditional expectation of the four terms above. 

 Term \eqref{eq:KW1} is bounded using to the sharpness condition \eqref{cond:D1}
  \begin{equation}\label{eq:KW-1}
- \nabla l (\bm x_t)^\top \frac{(\bm x_t-\bm x_t^\star)}{\| \bm x - \bm x^\star \| } \leq -\kappa  \,.
 \end{equation}
 Term \eqref{eq:KW2} is bounded by the Taylor approximation condition \eqref{eq:KWconds}. Specifically
 \begin{equation}\label{eq:KW-2}
  \bm \delta_t^\top \frac{(\bm x_t - \bm x_t^\star)}{\|\bm x_t - \bm x_t^\star \|} \leq \| \bm \delta_t \|
= \left\|
\nabla l (\bm x) 
     -
     \frac{\bm l (\bm x_t+\bm \nu) - \bm l (\bm x_t - \bm \nu)  }{2 \nu}
\right\| \leq c \nu^2 \, .   
 \end{equation}
 Term \eqref{eq:KW3} has zero mean
 \begin{equation}\label{eq:KW-3}
      \mathbb E \left[ 
        \bm \epsilon_t^\top \frac{(\bm x_t - \bm x_t^\star)}{\|\bm x_t - \bm x_t^\star \|} \Big| \mathcal F_t
    \right] = 0 \, .   
 \end{equation}
For Term \eqref{eq:KW4}, $\bm y_{t+1} = \bm x_t - \alpha_t \bm c_t$
\begin{align}\label{eq:KW-4} 
    \mathbb E 
    \left[ 
        \frac{\| \bm y_{t+1} - \bm x_t \|^2}{\| \bm x_{t} - \bm x_{t}^\star \|}
        \Big|
        \mathcal F_t
    \right]
    & \leq 
    \frac{\alpha_t^2}{E_t }
    \mathbb E 
    \left[
        || \bm c_t||^2 
        \big|
        \mathcal F_t
    \right]
    \leq
    \frac{\alpha_t^2 \sigma_l^2}{E_t \nu^4}
\end{align}
Since the variance of $l(\bm x, \hat w)$ is bounded (by $\sigma_l^2$), the variance of $|| \bm c_t||$ is bounded. Above, we let $\sigma_l^2/\nu^2 $ define this upper bound. 

Applying bounds \eqref{eq:KW-1}, \eqref{eq:KW-2}, \eqref{eq:KW-3} and \eqref{eq:KW-4} respectively to the terms \eqref{eq:KW1}, \eqref{eq:KW2}, \eqref{eq:KW3}  and \eqref{eq:KW4}  gives
\begin{align} \label{eq:linebound}
    \mathbb E
    \Big[
        \| \bm x_{t+1} - \bm x_{t+1}^\star \| 
        \big|
        \mathcal F_t 
    \Big]
    &
    \leq 
    \| \bm x_{t} - \bm x_{t}^\star \| 
    - \alpha_t \kappa 
    + \alpha_t  c \nu^2 
    +\frac{\alpha_t^2 \sigma_l^2}{E_t \nu^4}\, .
\end{align}
Notice if we choose 
\begin{equation} \label{eq:ETdef}
\nu \leq \sqrt{\frac{\kappa}{3 c}} \qquad \text{and} \qquad E_t = \frac{3 \sigma_l^2}{\nu^4 \kappa} \alpha_t  \, ,
    \end{equation}
then application to \eqref{eq:linebound} gives
\[
    \mathbb E
    \Big[
        \| \bm x_{t+1} - \bm x_{t+1}^\star \| 
        \big|
        \mathcal F_t 
    \Big]
    \leq 
    \| \bm x_{t} - \bm x_{t}^\star \| 
    - \alpha_t \frac{\kappa }{3}
    \qquad \textrm{on the event} \qquad \Big\{ \| \bm x_{t} - \bm x_{t}^\star \| \geq \frac{3  \sigma_l^2}{\nu^4 \kappa} \alpha_t\Big\} \, . 
\] 
This verifies that Condition \eqref{fcond:1} of Theorem \ref{Thrm1} holds. 

We must also verify Condition \eqref{fcond:2}. (The argument that follows is more-or-less identical to the verification of \eqref{fcond:2} in Theorem \ref{ThrmNew}.) For this notice that 
\begin{align*}
    \| \bm x_{t+1} - \bm x_{t+1}^\star \|
    \leq 
    \|\bm x_{t+1} - \bm x_{t}^\star \|
    \leq 
    \| \bm y_{t+1} - \bm x_{t}^\star \|
    \leq 
    \| \bm y_{t+1} - \bm x_{t}  \|
    +
    \| \bm x_{t} - \bm x_{t}^\star  \|
    = \alpha_t \| \bm c_t \| +
    \| \bm x_{t} - \bm x_{t}^\star  \|\, .
\end{align*}
and 
\begin{align*}
     \| \bm x_{t} - \bm x_{t}^\star \|
    \leq 
    \|\bm x_{t} - \bm x_{t+1}^\star \|
    & \leq 
    \| \bm x_{t} - \bm x_{t+1} \|
    +
    \| \bm x_{t+1} - \bm x_{t+1}^\star \|
    \\
    & \leq 
    \| \bm y_{t+1} - \bm x_{t}  \|
    +
    \| \bm x_{t+1} - \bm x_{t+1}^\star  \|
    = \alpha_t \| \bm c_t \| +
    \| \bm x_{t+1} - \bm x_{t+1}^\star  \|\, .
\end{align*}
Thus 
\begin{align}\label{eq:cup}
        \Big| \| \bm x_{t+1} - \bm x_{t+1}^\star \|
        -
        \| \bm x_{t} - \bm x_{t}^\star  \| \Big|
    &\leq 
    \alpha_t \|\bm c_t \| 
\end{align}

Since Condition \eqref{cond:D2} holds, we take $M \geq \sup_t \mathbb E \big[
		e^{  \lambda \left\|
	\bm c_t 
\right\|
}
\big|
\mathcal F_t 
\big] $. We let $Y$ be the random variable with CCDF:
$
 \mathbb P ( Y  \geq y ) = 1 \wedge ( Me^{-\lambda y} ) .
$
Thus for $y \in \mathbb R_+$
\begin{align*}
\mathbb P
\left(
	\left|
	f(\bm x_{t+1} ) - f (\bm x_t) 
\right|
\geq \alpha_t y 
\Big|
\mathcal F_t 
\right)
\notag 
& \leq 
\mathbb P
\left(
	\| \bm c_t \|
\geq y 
\Big|
\mathcal F_t 
\right) 
\\
& \leq \,
\min \big\{ 1, e^{-\lambda y }\mathbb E \big[
		e^{  \lambda \left\|
	\bm c_t 
\right\|
}
\big|
\mathcal F_t 
\big]\big\} 
\leq \,
\mathbb P (  Y\geq y ) 
\end{align*}
Above, we apply \eqref{eq:cup} and a Chernoff bound.
From this inequality above, we see that Condition \eqref{fcond:2} follows from Condition \eqref{cond:D2}.

We can now apply Theorem \ref{Cor:1} which gives:\begin{align*}
\mathbb P \Big(
\min_{\bm x \in \mathcal X^\star}
	 	\left\|
	\bm x_{t+1} - \bm x
\right\| \geq z
\Big)
\leq \hat I e^{-\hat J t z^\gamma }
\quad \text{and}\quad 
\mathbb E 
\left[
\min_{\bm x \in \mathcal X^\star}
	 	\left\|
	\bm x_{t+1} - \bm x
\right\|
\right] 
\leq \frac{\hat K}{t^\gamma}
\end{align*}
for constants $\hat I$, $\hat  J$ and $\hat K$. 
Since we also assume in addition that $l: \mathcal X \rightarrow \mathbb R$ is Lispchitz continuous (with Lipschitz constant $\hat L/\hat K$) we have, as required, 
\begin{align*}
\mathbb E \left[
	l(\bm x_{t+1}) - \min_{\bm x \in \mathcal X} l(\bm x)
\right]
\leq 
\frac{\hat L}{\hat K}
\mathbb E \left[
\min_{\bm x \in \mathcal X^\star}
	 	\left\|
	\bm x_{t+1} - \bm x
\right\|
\right] 
\leq \frac{\hat L}{t^\gamma} \, .
\end{align*}
\Endproof
\end{darkblue}

\begin{darkblue}
    \subsection{Stochastic Frank-Wolfe: Proof of Theorem \ref{thrm:FW}} \label{sec:FWProofs}

    The main aim of this section is to prove Theorem \ref{thrm:FW}. We also show that the distance function satisfies the conditions of our main theorem. This suggests that if the objective function behaves linearly rather than quadratically near the optimum, we should anticipate faster convergence. We also discuss how linear convergence can hold for Stochastic Frank-Wolfe in the same manner that we proved for Projected Stochastic Gradient Descent. 
    
Before proceeding with the proof of Theorem \ref{thrm:FW} we require a couple of lemmas. 

Lemma \ref{lem:FW1} is used to show that there is sufficient negative drift in the Frank-Wolfe algorithm.
\begin{lemma}\label{lem:FW1}
    If Condition \eqref{cond:D1} and Condition \eqref{cond:E1} hold 
    then there exists a $\hat \kappa >0$ such that for all $\bm x \in \mathcal X $ there exists $\bm y \in \mathcal{X} $ such that 
    \[
        (\bm y - \bm x)^\top \nabla l(\bm x)  \leq -\hat \kappa\, .
    \]
\end{lemma} 
\Beginproof{}
By Condition \eqref{cond:E1}, there exists a constant $d>0$ such that
\begin{equation}\label{eq:lemFW1}
\min_{\bm x^\star \in \mathcal{X}^\star , \bm y \in \delta \mathcal{X} } \| \bm y-\bm x^\star \| \geq d \, .
\end{equation}
By Condition \eqref{cond:D1}, for all $\bm x \notin \mathcal X^\star$ there exists $\bm x^\star \in \mathcal X^\star$ 
\begin{equation}\label{eq:lemFW2}
\frac{(\bm x^\star - \bm x)^\top}{\|\bm x^\star - \bm x\|} \nabla l(\bm x) \leq - \kappa \, .    
\end{equation}

We let $\bm y(t)=  \bm x + t(\bm x^\star- \bm x)$ for $t \in \mathbb R$. Notice that 
\begin{equation}\label{eq:lemFW3}
\frac{(\bm y(t) - \bm x)^\top}{\|\bm y(t) - \bm x\|}=\frac{(\bm x^\star - \bm x)^\top}{\|\bm x^\star - \bm x\|}    
\end{equation}
Letting $t^\star = \max\{ t : \bm y(t) \in \mathcal X\}$, we see that 
\begin{equation}\label{eq:lemFW4}
\bm y := \bm y(t^\star) \in \delta \mathcal{X}
\end{equation}
Combining (\ref{eq:lemFW1}-\ref{eq:lemFW4}), we see that 
\begin{equation*}
    (\bm y - \bm x)^\top \nabla l(\bm x) 
    =
    \| \bm y - \bm x \| 
    \frac{(\bm x^\star - \bm x)^\top}{\|\bm x^\star - \bm x\|}
    \nabla l(\bm x) 
    \leq 
    - d \kappa =: -\hat \kappa
\end{equation*}
as required.
\Endproof

We now restate and prove Theorem \ref{thrm:FW}

\FrankWolfe*

\Beginproof{ of Theorem \ref{thrm:FW}}
The proof here combines ideas from Theorem \ref{ThrmNew} with the adjustments for stochastic effects for the Frank-Wolfe algorithm given in Theorem 3 from \cite{hazan2016variance}. 

In the proof we define $D:= \max_{\bm x, \bm v} \| \bm x - \bm v\|$ and we let $\epsilon ( \bm x_t) = l(\bm x_t ) - l(\bm x^\star)$ and we define $\sigma$ such that 
\begin{equation*}
    \mathbb E \left[ 
        \|
            \nabla l(\bm x_t) - \bm c_t^i 
        \|^2
    \right]
    \leq \sigma^2 \,
    \qquad \qquad \forall i,t .
\end{equation*}
(Note that $\sigma$ is finite by the moment generating function condition \eqref{cond:D2})

By Condition \eqref{cond:E1}
\begin{align}
    \frac{\epsilon(\bm x_{t+1})^2}{2} - \frac{\epsilon(\bm x_t)^2}{2}
    &
    \leq 
    \epsilon ( \bm x_t) \big( \bm x_{t+1} - \bm x_t \big)^\top \nabla \epsilon ( \bm x_t)
    +
    \frac{K}{2} \|\bm x_{t+1} - \bm x_t \|^2 
    \nonumber 
    \\
    & 
    =
    \alpha_t \epsilon (\bm x_t) ( \bm v_t - \bm x_t)^\top \bm c_t
    +
    \alpha_t \epsilon (\bm x_t) (\bm v_t - \bm x_t )^\top
    \left[
        \nabla \epsilon(\bm x_t) - \bm c_t
    \right]
    +
    \frac{K}{2} \alpha_t^2 \| \bm v_t - \bm x_t \|^2 \, .
    \label{eq:smconv}
\end{align}
We now consider the event where the following bound holds
\begin{equation}
   \left\{ \epsilon(\bm x_t ) \geq \frac{3\alpha_t K D^2}{2\kappa} \right\}\, .
    \label{eq:EpsIneq}
\end{equation}
Thus 
\begin{align}
    &\epsilon(\bm x_{t+1}) \nonumber
    \\
    \leq &\, 
    \sqrt{
    \epsilon(\bm x_t)^2
    +
    2\alpha_t \epsilon (\bm x_t) ( \bm v_t - \bm x_t)^\top \bm c_t
    +
    2\alpha_t \epsilon (\bm x_t) (\bm v_t - \bm x_t )^\top
    \left[
        \nabla \epsilon(\bm x_t) - \bm c_t
    \right]
    +
    {K} \alpha_t^2 \| \bm v_t - \bm x_t \|^2
    }
    \nonumber 
    \\
    =
    &\,
    \epsilon(\bm x_t)
    \sqrt{
    1
    +
    2\frac{\alpha_t}{ \epsilon (\bm x_t)} ( \bm v_t - \bm x_t)^\top \bm c_t
    +
    2\frac{\alpha_t}{ \epsilon (\bm x_t)} (\bm v_t - \bm x_t )^\top
    \left[
        \nabla \epsilon(\bm x_t) - \bm c_t
    \right]
    +
    {K} \frac{\alpha_t^2}{\epsilon(\bm x_t)^2} \| \bm v_t - \bm x_t \|^2
    }
    \nonumber
    \\
    \leq \, 
    &
    \epsilon(\bm x_t)
    +
    {\alpha_t} ( \bm v_t - \bm x_t)^\top \bm c_t
    +
    {\alpha_t} (\bm v_t - \bm x_t )^\top
    \left[
        \nabla \epsilon(\bm x_t) - \bm c_t
    \right]
    +
    \frac{K}{2} \frac{\alpha_t^2}{\epsilon(\bm x_t)} \| \bm v_t - \bm x_t \|^2  
    \nonumber
    \\
    \leq 
    &
    \epsilon(\bm x_t)
    +
    {\alpha_t} ( \bm y_t - \bm x_t)^\top \bm c_t
    +
    {\alpha_t} \|\bm v_t - \bm x_t \|
    \|
        \nabla \epsilon(\bm x_t) - \bm c_t
    \|
    +
    \frac{\alpha_t \kappa}{3} \nonumber
    \\
    \leq
    &
    \epsilon(\bm x_t)+
    \alpha_t ( \bm y_t - \bm x_t )^\top \bm c_t 
    +
    \alpha_t D 
        \|
            \nabla \epsilon( \bm x_t) 
            - \bm c_t 
        \|  
    +
    \frac{\alpha_t \kappa}{3} 
    \label{eq:epsexpand}
\end{align}
In the first inequality, above we rearrange the expression \eqref{eq:smconv}.
In the second inequality, we apply the inequality $\sqrt{1+z} \leq 1+z/2$.
In the third equality, 
we note that $\bm v_t^\top \bm c_t \leq \bm y_t^\top \bm c_t$, by the definition of $\bm v_t$ \eqref{eq:FrankWolfe1}. Here we let $\bm y_t \in \mathcal X$ be as defined in Lemma \ref{lem:FW1}. 
Also, we apply the Cauchy-Schwarz Inequality and the bound \eqref{eq:EpsIneq}. In the final inequality we note that $\|\bm v_t - \bm x_t \|\geq D$.

Taking the conditional expectation of \eqref{eq:epsexpand}, we see that, on the event \eqref{eq:EpsIneq}, the following holds
\begin{align}
    \mathbb E
    \left[
        l(\bm x_{t+1} ) - l(\bm x_t) | \mathcal F_t 
    \right]
    &
    =
    \mathbb E
    \left[
        \epsilon(\bm x_{t+1} ) - \epsilon(\bm x_t) | \mathcal F_t 
    \right]
    \nonumber
    \\
    &
    \leq
    \alpha_t ( \bm y_t - \bm x_t )^\top \mathbb E [\bm c_t | \mathcal F_t]
    +
    \alpha_t D \mathbb E 
    \left[
        \|
            \nabla \epsilon( \bm x_t) 
            - \bm c_t 
        \| 
        |
        \mathcal F_t
    \right]
    +
    \frac{\alpha_t \kappa}{3} 
    \nonumber
    \\
    &
    \leq 
    - \alpha_t \hat{\kappa} 
    + \alpha_t D \mathbb E
    \left[ 
        \|
            \nabla l ( \bm x_t) - \bm c_t 
        \|
        |
        \mathcal F_t
    \right]
    +
    \frac{\alpha_t \kappa}{3} \, .
    \label{eq:lDiff}
\end{align} 
Notice that, since $m_t \geq (3\sigma D / \hat \kappa \alpha_t)^2$ ,
\[
 \mathbb E 
 \big[
    \|
        \nabla l (\bm x_t ) 
        -
        \bm c_t
    \|
    \, | 
    \mathcal F_t
 \big]
 \leq 
 \sqrt{
  \mathbb E 
 \left[
    \|
        \nabla l (\bm x_t ) 
        -
        \bm c_t
    \|^2
    \, | 
    \mathcal F_t
 \right]
 }
 \leq 
 \frac{\sigma}{\sqrt{m_t}} \
 \leq \frac{\hat \kappa }{3 D}\, .
\]
Now applying this inequality to \eqref{eq:lDiff} gives
\[
\mathbb E \big[ 
    l(\bm x_{t+1} ) - l (\bm x_t) 
    |
    \mathcal F_t
\big]
\leq 
-
\alpha_t \frac{\hat \kappa }{3} 
\]
on the event $l(\bm x_t) - l(\bm x^\star) \geq 3KD / \alpha \hat \kappa $.
Thus Condition \eqref{fcond:1} is met.

For Condition \eqref{fcond:2}, since $l$ is Lipschitz continuous and the set $\mathcal X$ is bounded we have
\[
\| l(\bm x_{t+1}) - l (\bm x_t) \|
\leq 
L \| \bm x_{t+1} - \bm x_t \|
\leq 
\alpha_t L \| \bm v_t - \bm x_t \|
\leq 2 \alpha_t L \max_{\bm x \in\mathcal{X}} \| \bm x\| \, .
\]
Thus we see that Condition \eqref{fcond:2} holds with a constant upperbound $Y=2 L \max_{\bm x \in\mathcal{X}} \| \bm x\|$ .

Here we see that the conditions of Theorem \ref{Cor:1} are met, and thus we have that 
    \begin{equation*}
        \mathbb P \left( 
            l(\bm x_{t+1} ) - l( \bm x_t ) \geq z 
            \right)\leq I e^{-t^{-\gamma} J z} \, ,
    \end{equation*}
    as required. 
\Endproof

\subsubsection{Cones satisfy Condition (\ref{cond:E1})}

Below we recall that we define the matrix norm $\| \cdot \|_S$ for a positive semi-definite matrix $S$ by
\[
 \| \bm x \|_S := \sqrt{\bm x^\top S \bm x }
\] 
\begin{lemma}\label{lem:posdef}
For a symmetric positive definite matrix $S$, the distance function 
    \[
        d_{\mathcal X^\star} (\bm x)
        =
        \min_{x^\star \in \mathcal X^\star}
        \|  
            \bm x - \bm x^\star 
        \|_S
    \]
satisfies Condition \eqref{cond:E1}. 
\end{lemma} 
\Beginproof{}
    We must show that the function 
    \[
        d_{\mathcal X^\star}(\bm x)^2 
        =
                \min_{x^\star \in \mathcal X^\star}
        \|  
            \bm x - \bm x^\star 
        \|_S^2
    \]
    is strongly convex.

    Given $\bm x$, we let $\bm x^\star =\argmin_{\bm x^\star \in \mathcal X^\star}
        \|  
            \bm x - \bm x^\star 
        \|_S  $. By the Envelope Theorem,
    \begin{equation} \label{eq:dXx}
        \nabla d_{\mathcal X}(\bm x)^2 
        = 
        2 S (\bm x-\bm x^\star)         
    \end{equation}
    Also 
    \begin{equation} \label{eq:xyS}
        \| \bm x - \bm y \|_S^2 \leq \lambda_{\max} (S) \| \bm x - \bm y\|^2
    \end{equation}
    where $\lambda_{\max}(S)$ is the maximum eigenvalue of $S$. 

    Now for any $\bm y$ and $\bm x$,
    \begin{align*}
        d_{\mathcal X^\star}(\bm y)^2
        = 
        \min_{\bm y^\star \in \mathcal X^\star}
        \|  
            \bm y - \bm y^\star 
        \|_S^2
        &
        \leq 
        \| 
            \bm y - \bm x^\star
        \|^2
        \\
        &
        =
        \| \bm y - \bm x + \bm x -\bm x^\star \|^2_S
        \\
        &
        =
        \| \bm x - \bm x^\star \|^2_S
        +
        2 (\bm y - \bm x)^\top S (\bm x - \bm x^\star)
        +
        \| \bm y - \bm x \|^2_S
        \\
        &
        \leq 
        d_{\mathcal X^\star}(\bm x)^2 + 
        (\bm y - \bm x) \nabla d_{\mathcal X^\star}^2 (\bm x)
        +
        \lambda_{\max} (S) \| \bm x - \bm y\|^2
\end{align*}
In the first inequality, we apply the sub-optimality of $\bm x^\star$ with respect to the point $\bm y$. 
In the second inequality, we apply \eqref{eq:dXx} and \eqref{eq:xyS}. Thus from the above inequality we see that $d_{\mathcal X^\star}(x)^2$ is a  $\lambda_{\max}(S)$--smoothly convex function, as required.
\Endproof

\end{darkblue}

\subsection{Appendix to Section \ref{sec:Linear} : Linear Convergence Proofs}\label{sec:LinConvAppendix}
As discussed, our proof follows the main argument of Theorem 3.2 of \cite{davis2019stochastic}.
We divide the procedure into $S$ stages. 
We consider PSGD with constant step size within each stage, as defined in \eqref{eq:PSGDLinear}. 
The task of each stage is to half the error with the optimum.
We apply our bound Lemma \ref{lem:unbounded}, which is a stronger concentration bound than, Theorem 4.1, applied in \cite{davis2019stochastic}. This leads to some improvements in the bounds found there.

\subsubsection{Exponential Concentration for constant step-size and unbounded state-space.}
\begin{darkblue}
We state an exponential concentration bound for constant step-sizes below. We note that we do not require the function $f(x)$ or the set $\mathcal X$ to be bounded (or constrained) for this result to hold.
\begin{restatable}{lemma}{unbounded}
\label{lem:unbounded}
    For constant step sizes $\alpha$ 
\[
    \mathbb P( f(\bm x_{t+1}) - f(\bm x^\star) 
    \geq z | \mathcal F_0)
    \leq 
    e^{-\frac{Q}{\alpha} z } 
    \left\{
       e^{\frac{Q}{\alpha} (
        f(\bm x_{0})
        -
        f(\bm x^\star)
        )  }   
     e^{-t\frac{Q}{\alpha}  \kappa / 2 }
    +
    D 
    \frac{     e^{Q \kappa / 2 }}{    1- e^{-Q \kappa / 2 }}e^{Q B} 
    \right\}\, .
\]
\end{restatable}

\Beginproof{}
There are no boundedness assumptions placed in 
Lemma \ref{lem:LMGF}. We restate the conclusion of that result here:
\begin{equation}\label{eq:delivery}
    \mathbb{E}[e^{\eta L_{t+1}} | \mathcal F_{T_0}]\leq\mathbb{E}[e^{\eta L_{T_1}} | \mathcal F_{T_0} ]\prod_{k=T_1}^{t}\rho_t + D\sum_{\tau=T_1+1}^{t+1}\prod_{k=\tau}^{t}\rho_k \, ,
\end{equation}
for $t\geq T_1 \geq T_0$.
If we consider the above terms for constant step sizes  $\alpha= \alpha_t$ then
\begin{align*}
    L_{t+1}
    & =
    f(\bm x_{t+1} ) 
    -f (\bm x^\star) - \alpha B 
    \\
    \rho_t 
    = 
    \rho
    & :=
    e^{-\alpha \eta \kappa
    +\alpha^2 \eta^2 E }
    \leq 
    e^{-\alpha \eta \frac{\kappa}{2} }
    \qquad \text{for} \qquad \alpha \eta \leq Q\\
    T_0
    &
    =
    \min \{ 
        t : \frac{\alpha_t - \alpha_{t+1}}{\alpha_t} \leq \frac{\kappa}{2B}  
    \} = 0\\
    T_1 & =0\, ,
\end{align*}
also
\[
\prod_{k=0}^t \rho_k = \rho^{t+1} \leq 1 
\quad \text{and}\quad
    \sum_{l=1}^{t+1} \prod_{k=l}^t \rho_k 
=
    \sum_{l=1}^{t+1} \rho^{t-l}
\leq 
    \sum_{l=-1}^\infty \rho^l = \frac{\rho^{-1}}{1-\rho} \, . 
\]
with these terms the above expression \eqref{eq:delivery} gives the requied bound
\[
\mathbb E [ e^{\eta (f(\bm x_{t+1} ) 
    -f (\bm x^\star))} | \mathcal F_0 ]
\leq 
e^{\eta (f(\bm x_{0} ) 
    -f (\bm x^\star))} \rho^{t+1} 
+
D\frac{\rho^{-1}}{1-\rho}e^{\alpha\eta B} \, .
\]
Applying Markov's inequality gives
\[
\mathbb P \left(
    f(\bm x_{t+1}) - f(\bm x^\star) \geq z 
\right)
\leq 
e^{-\eta z}
\left\{
e^{\eta (f(\bm x_{0} ) 
    -f (\bm x^\star))} \rho^{t+1} 
+
D\frac{\rho^{-1}}{1-\rho}e^{\alpha\eta B} 
\right\}\, .
\]
Taking $\eta = Q/\alpha$ gives the required bound. 
\Endproof

\subsubsection{Linear Convergence under Exponential Concentration}

We note that while we generally assume that the set $\mathcal X$ is bounded, the above lemma and the linear convergence results of this section apply to unbounded constraint sets. We now prove Theorem \ref{ThrmLinear}.     
\end{darkblue}

\PSGDLinearThrm* 

\Beginproof{ of Theorem \ref{ThrmLinear}}
%
 First, we recall some notation: 
$
f(\hat{\bm x}_s)
:= \min_{\bm x \in \mathcal X^\star} \| \hat{ \bm x}_s - \bm x \|$  \text{and}  
 $F = \max_{x  \in \mathcal X} f(x) \, .
$
The constants $D$ and $E$ are the moment generating function constants as defined in Lemma \ref{lem:1} and Lemma \ref{lem:LMGF}, respectively. 
%

We define the event 
$
\mathcal E_s := \left\{
	f(\hat{ \bm x}_s) \leq 2^{-s} F
\right\} \, .
$
So $\mathbb P(\mathcal E_0 ) = 1$. We inductively analyze $\mathbb P(\mathcal E_s )$. Notice 
\begin{equation}\label{eq:Borel}
\mathbb P ( \mathcal E_s ) 
\geq\,
 \mathbb P( \mathcal E_s | \mathcal E_{s-1} ) 
\mathbb P( \mathcal E_{s-1} ) 
=\,
\left(
	1 - \mathbb P ( \mathcal E_s^c | \mathcal E_{s-1} ) 
\right)
\mathbb P \left(  \mathcal E_{s-1} \right)
\geq\,
\mathbb P ( \mathcal E_{s-1} ) 
-
\mathbb P \left( \mathcal E_{s}^c | \mathcal E_{s-1} \right) \, . 
\end{equation}
By Lemma \ref{lem:unbounded}, for $\bm \hat x_{s-1}$ such that $f(\hat{ \bm x}_{s-1}) \leq 2^{-s+1} F$ we have
\begin{align*}
\mathbb P \left( 
\mathcal E_s^c | \hat{ \bm x}_{s-1}
 \right)
\notag 
=\,
\mathbb P \left( 
f(\hat{ \bm x}_s) \geq 2^{-s} F | \hat{ \bm x}_{s-1}
 \right)
    &
    \leq 
    e^{-\frac{Q}{\hat \alpha_s} z } 
    \left[
       e^{\frac{Q}{\hat \alpha_s} (
        f(\hat{ \bm x}_{s-1})
        -
        f(\bm x^\star)
        )  }   
     e^{-t\frac{Q}{\hat \alpha_s}  \kappa / 2 }
    +
    D 
    \frac{ e^{Q \kappa / 2 }}{    1- e^{-Q \kappa / 2 }}e^{Q B} 
    \right]\, 
    \\
    & 
    \leq 
    e^{-\frac{Q}{\hat \alpha_s} z } 
    \left[
       \exp \left\{\frac{Q}{\hat \alpha_s} 2^{-s+1}F    
     -t\frac{Q}{\hat \alpha_s}  \kappa / 2 \right\}
    +
    D 
    \frac{ e^{Q \kappa / 2 }}{    1- e^{-Q \kappa / 2 }}e^{Q B} \, 
    \right] \, .
\end{align*}
(Here we apply Lemma \ref{lem:unbounded} for times $t=T_{s-1},...,T_{s}-1$ with expectation $\mathbb E [\cdot]$ given by $\mathbb E[\cdot | \hat x_{s-1}]$.)

Notice that the term in curly brackets above is negative iff $ t_s \geq {2^{-s+1}F}/{\kappa \hat{\alpha}_s}$.
If this holds then
\begin{align*}
\mathbb P \left( 
\mathcal E_s^c | \mathcal E_{s-1}
 \right)
\leq \,
&
R e^{- 2^{-s} F \kappa / 2E\hat{\alpha}}
\qquad \text{where} \quad 
R:= 1 +  D 
    \frac{ e^{Q \kappa / 2 }}{    1- e^{-Q \kappa / 2 }}e^{Q B}\, . 
\end{align*}
Applying this to \eqref{eq:Borel},
$
\mathbb P \left( 
\mathcal E_s
 \right) 
\geq\,
\mathbb P \left( 
\mathcal E_{s-1}
 \right)
-
\mathbb P \left( 
\mathcal E_s^c
|
\mathcal E_{s-1}
 \right)
\geq 
\mathbb P \left( 
\mathcal E_{s-1}
 \right)
-
R e^{- 2^{-s} F \kappa / 2E\hat{\alpha}_s}
$
. So we have
\begin{align}\label{eq:Ssum}
\mathbb P \left( \mathcal E_S  \right)
\geq 
1- 
\sum_{s=1}^S 
R e^{- 2^{-s} F \kappa / 2E\hat{\alpha}_s}\, .
\end{align}
The total number of computations/samples required is 
$
\sum_{s=1}^S t_s \geq  \sum_{s=1}^S 2^{-s}F/ \kappa \hat{\alpha}_s  \, .
$

We now prove part a). Given the bounds above, we can optimize the number of samples to achieve a probability $1-\hat{\delta}$. That is we solve
\begin{align*}
\text{minimize} \quad 
\sum_{s=1}^S \frac{2^{-s+1}F}{ \kappa \hat{\alpha}_s }
\quad 
\text{such that}
\quad 
\sum_{s=1}^S 
R e^{- 2^{-s+1} F \kappa / 2E\hat{\alpha}_s} \leq \hat \delta 
\quad \text{over} \quad \hat{\alpha}_s > 0 \, .
\end{align*}
A short calculation shows that this is minimized by 
$
\hat{\alpha}_s = 2^{-s}F \kappa /  E \log (RS/\hat{\delta}) 
$
and thus since  $t_s \geq 2^{-s+1}F/ \kappa \hat{\alpha}_s$ we define
$
  t_s = \left\lceil \frac{2}{\kappa^2} \log\left(
	\frac{RS}{\hat \delta}\right)  \right\rceil
$
and the number of samples required here is $S\times t_s$ which equals
$
 S\left\lceil \frac{2}{\kappa^2} \log \left(
	\frac{RS}{\hat \delta}
\right) 
 \right\rceil .
$
Since for an $\hat \epsilon$ approximation, we require $S$ to be such that $\hat \epsilon \geq 2^{-S} F$, we take $S = \lceil \log_2( F /\hat \epsilon) \rceil$. Thus we see that an $\hat \epsilon$ approximation can be achieved with a probability greater than $1-\hat \delta $ in a number of samples given by
\begin{align*}
\Big\lceil \log_2\Big( \frac{F }{\hat \epsilon}\Big) \Big\rceil \left\lceil \frac{2}{\kappa^2} \log \left(
	\frac{R}{\hat \delta}\right)  + \log \Big(\Big\lceil \log_2\Big( \frac{F }{\hat \epsilon}\Big) \Big\rceil\Big)
 \right\rceil \, .
\end{align*}
This gives the part a) of Theorem \ref{ThrmLinear}.

Notice, we can make the sum \eqref{eq:Ssum} finite for $S=\infty$.
Specifically if we take $\hat{\alpha}_s = a / 2^s \log (s+1)$ and $t_s = (\log (s+1))^2$ then
the Condition \eqref{fcond:1} holds $\forall s \geq s_0$ for $s_0 = \lceil e^{2F/\kappa} \rceil +1 $ and thus
\begin{align*}
\sum_{s=s_0}^\infty
R e^{- 2^{-s+1} F \kappa / 2E\hat{\alpha}_s} 
=
\sum_{s=s_0}^\infty 
R 
\frac{1}{ (s+1)^{aF\kappa / 2E} }
\leq 
R
\int_{s_0}^\infty  
\frac{1}{ s^{aF\kappa / 2E} } ds
\leq 
\frac{2 R E }{a F \kappa } \cdot \frac{1}{s_0^{aF\kappa / 2E}}
\leq  \frac{2 R E }{a F \kappa } .
\end{align*}
The above sum is less than $\hat \delta$ for $a \geq R E /  2 \hat{\delta} F \kappa  $. Letting 
$
A = 2R E / F \kappa $ \text{and}  $M = 2^{s_0} F \, ,
$
we see that 
for $a \geq A/\hat \kappa$ gives
\begin{align*}
\mathbb P \left( 
\exists s \in \mathbb N\text{ s.t. } \min_{x \in \mathcal X^\star}\|x_s - x \| \geq 2^{-s} F
 \right)
& \leq  
\sum_{s=1}^\infty \mathbb P \left(  \mathcal E_s^c \cup \big(\cap_{s' \leq s} \mathcal E_{s'} \big) \right)
\notag 
\leq 
\sum_{s=1}^\infty \mathbb P \left(  \mathcal E_s^c | \mathcal E_{s-1} \right)
\leq 
\sum_{s=1}^\infty
R e^{- \frac{2^s F \kappa }{ 2E\hat{\alpha}}} 
\leq \hat \delta \, .
\end{align*}
Thus 
for learning rates $\hat{\alpha}_s = a / 2^s \log (s+1)$ with $a \geq M/ \hat \delta $ if holds that 
$
\mathbb P 
\left(
	 \forall s , \min_{x \in \mathcal X^\star} \|  x_s - x\| \leq 2^{-s} M
\right)
\geq 1- \hat \delta \, .
$
This gives the 2nd part of Theorem \ref{ThrmLinear}.
\Endproof

\subsubsection{Application to Specific Stochastic Approximation Algorithms.}

The following is the equivalent linear convergence result for Kiefer-Wolfowotiz

\begin{corollary}[Linear Convergence in Projected Stochastic Gradient Descent]
Assume Conditions \eqref{cond:D1} and \eqref{cond:D2} hold for PSGD with rates given in \eqref{eq:PSGDLinear}. If, for $\hat \epsilon >0$ and $\hat\delta \in (0,1) $, we set
\begin{align*}
S = \log\left(
	\frac{F}{\hat \epsilon}
\right)
,
\quad 
\hat{\alpha}_s = \frac{2^{-s}F \kappa }{  E \log \Big(\frac{GS}{\hat{\delta}}\Big) }
, \quad \text{and} \quad 
t_s = \left\lceil \frac{2}{\kappa^2} \log\left(
	\frac{GS}{\hat \delta}\right)  \right\rceil
\end{align*}
then with probability greater than $1-\hat \delta$ it holds that
$
\min_{\bm x \in \mathcal X^\star}\left\|
	\hat{\bm x}_S - \bm x 
\right\| \leq \hat \epsilon \, .
$
Moreover, the number of iterations \eqref{eq:PSGDLinear} required to achieve this bound is
\begin{align*}
\Big\lceil \log_2\Big( \frac{F }{\hat \epsilon}\Big) \Big\rceil \left\lceil \frac{2}{\kappa^2} \log \left(
	\frac{G}{\hat \delta}\right)  + \log \Big(\Big\lceil \log_2\Big( \frac{F }{\hat \epsilon}\Big) \Big\rceil\Big)
 \right\rceil \, .%
\end{align*}
\end{corollary}

\begin{corollary}[Linear Convergence of Kiefer-Wolfowitz]
 \label{KWLinear}
Assume Conditions \eqref{cond:D1}, \eqref{cond:D2}, \eqref{eq:KWconds} hold. For the KW algorithm, \eqref{eq:KW}, with step-sizes given in \eqref{eq:PSGDLinear}:
 If, for $\hat \epsilon >0$ and $\hat\delta \in (0,1) $, we set
\begin{align*}
S = \log\left(
	\frac{F}{\hat \epsilon}
\right)
,
\quad 
\hat{\alpha}_s = \frac{2^{-s}F \kappa }{  E \log \Big(\frac{GS}{\hat{\delta}}\Big) }
,\quad \nu_s = \sqrt{\frac{\kappa}{3c}} \quad \text{and} \quad 
t_s = \left\lceil \frac{2}{\kappa^2} \log\left(
	\frac{GS}{\hat \delta}\right)  \right\rceil
\end{align*}
then with probability greater than $1-\hat \delta$ it holds that
$
\min_{\bm x \in \mathcal X^\star}\left\|
	\hat{\bm x}_S - \bm x 
\right\| \leq \hat \epsilon \, .
$
Moreover, the number of iterations \eqref{eq:PSGDLinear} required to achieve this bound is
\begin{align*}
\Big\lceil \log_2\Big( \frac{F }{\hat \epsilon}\Big) \Big\rceil \left\lceil \frac{2}{\kappa^2} \log \left(
	\frac{G}{\hat \delta}\right)  + \log \Big(\Big\lceil \log_2\Big( \frac{F }{\hat \epsilon}\Big) \Big\rceil\Big)
 \right\rceil \, .%
\end{align*}
\end{corollary}

The proof of Corollary \ref{KWLinear} is identical to the proof of Theorem \ref{ThrmLinear}.

\begin{corollary}[Linear Convergence of Stochastic Frank-Wolfe]
Assume Condition \ref{cond:D1}, \ref{cond:D2}, \ref{cond:E1} and \ref{cond:E2} hold. For the SFW algorithm with step-sizes given by \eqref{eq:PSGDLinear}.
If 
    we set
\begin{align*}
S = \log\left(
	\frac{F}{\hat \epsilon}
\right)
,
\quad 
\hat{\alpha}_s = \frac{2^{-s}F \kappa }{  E \log \Big(\frac{GS}{\hat{\delta}}\Big) },
 \quad 
 m_s := \left\lceil \Big(\frac{3 \sigma}{ \kappa \alpha}\Big)^2 \right\rceil 
\,
, \quad \text{and} \quad 
t_s = \left\lceil \frac{2}{\kappa^2} \log\left(
	\frac{GS}{\hat \delta}\right)  \right\rceil
\end{align*}
then with probability greater than $1-\hat \delta$ it holds that
$
\min_{\bm x \in \mathcal X^\star}\left\|
	\hat{\bm x}_S - \bm x 
\right\| \leq \hat \epsilon \, .
$
Moreover, the number of iterations \eqref{eq:PSGDLinear} required to achieve this bound is
\begin{align*}
\Big\lceil \log_2\Big( \frac{F }{\hat \epsilon}\Big) \Big\rceil \left\lceil \frac{2}{\kappa^2} \log \left(
	\frac{G}{\hat \delta}\right)  + \log \Big(\Big\lceil \log_2\Big( \frac{F }{\hat \epsilon}\Big) \Big\rceil\Big)
 \right\rceil \, .%
\end{align*}
\end{corollary}

\section{Appendix to Theoretical Results}

\begin{darkblue}
\subsection{List of Notations for Theorem \ref{Cor:1}}\label{sec:constants}

There are several time-independent constants (usually denoted with a capital letter) in Theorem \ref{Cor:1}. We list these here.
\begin{align*}
B & && \text{Bound where the drift condition holds. See \eqref{fcond:1}.}
\\
C &= \frac{(1+H)}{2QG^n} + B \quad &&   \text{Constant in Proposition \ref{Thrm1}}\\
\\
D &= \mathbb E [ e^{\lambda Z} ] = \mathbb E [ e^{\lambda [Y+\kappa/2]}] \quad &&     \text{Moment Generating Function, see \eqref{eq:con2}.}
\\
E 
& 
=
\mathbb E \left[ \frac{e^{\lambda Z} - 1- \lambda Z}{\lambda^2} \right]
&& \text{See Lemma \ref{lem:LMGF}}
\\
F
&
= \max_{\bm x\in \mathcal X} f(\bm x) - \min_{\bm x \in \mathcal X} f(\bm x)
\\
G 
&
=
    \frac{1}{4^\gamma}, \quad \text{for } \alpha_t = \frac{a}{(u+t)^\gamma} && \text{See Lemma \ref{LemmaB} and Proof in Section \ref{finalproofthrm1}}
    \\
H
&
=  D {e^{\frac{\kappa Q G^n}{2}}}/{(1-e^{-\frac{\kappa Q G^n}{2}})}
&&
\text{Constant in Proposition \ref{Thrm1} Defined after \eqref{eq:Ltailpart}.}
\\
I
&
=
(1+H)e^{Q G/(F/\alpha_{T_2}-B)}
&&
\text{Constant in Theorem \ref{Cor:1}} 
\\
J
&
= Q G^n 
&&
\text{Exponent in Theorem \ref{Cor:1}} 
\\
K 
&
= \frac{I}{J}  
&&\text{Constant in Theorem \ref{Cor:1}} 
\\
n
&
=
\begin{cases}
     1 &\text{ for } \gamma < 1 \\
     \lceil \frac{\alpha_0 B + F}{ a \log 2} \rceil &\text{ for } \gamma  = 1
\end{cases}
\\
Q
&
= 
\lambda \wedge (\kappa/2E)
\\
T_0 &= \min\Big\{ t \geq 0 : \frac{(\alpha_s - \alpha_{s+1}) }{ \alpha_s} < \kappa / 2B, \forall s \geq t \Big\} && \text{See Lemma \ref{lem:1} and its proof.}
\\
T_1 &=
\begin{cases}
    \frac{\alpha_0 B + F}{ \frac{a}{1-\gamma} \left[ 1 - \frac{1}{2^{1-\gamma}} \right]}\,  & \text{for }\gamma < 1\\
    u2^n\,  & \text{for }
 \gamma =1 
\end{cases}
\\
T_2 &= T_0 \vee T_1 
\\
Y &
&& \text{Sub-exponential Random Variable defined in \ref{fcond:2}}
\\
Z &= Y + \frac{\kappa}{2} 
&&
\text{Random Variable defined in Lemma \ref{lem:1}}
\end{align*}    
\end{darkblue}

\subsection{Technical Lemmas for the Proof of Proposition \ref{Thrm1}}\label{sec:AppLem}

\lemone*
\Beginproof{ of Lemma \ref{lem:1}}
Applying the definition of $L_t$ and the drift Condition \eqref{fcond:1} gives 
\begin{align*}
\mathbb E [ L_{t+1} - L_t |\mathcal F_t ] 
=
& \,
\mathbb E [ f(\bm{x}_{t+1}) - f(\bm{x}_t) | \mathcal F_t ] + (\alpha_{t} - \alpha_{t+1} ) B 
\notag 
\\
\leq 
&
- 2\alpha_t \kappa + (\alpha_{t} - \alpha_{t+1} ) B
\notag 
\\
\leq 
&
-2\alpha_t \kappa \left[ 1 - (\alpha_t - \alpha_{t+1})B / \alpha_t \kappa \right]
\end{align*}
Since $(\alpha_t - \alpha_{t+1}) / \alpha_t \rightarrow 0$, there exists a constant $T_0$ such that $(\alpha_t - \alpha_{t+1}) / \alpha_t < \kappa / 2B$ for all $t \geq T_0$. Specifically we can take $T_0 = \min\{ t \geq 0 : (\alpha_s - \alpha_{s+1}) / \alpha_s < \kappa / 2B, \forall s \geq t \}$.
This gives the first drift condition \eqref{eq:con1}. 

For the second condition, for $t \geq T_0$ with $T_0$ as just defined: 
\begin{align*}
 \big[  |L_{t+1} - L_t |  \big| \mathcal F_t \big]
\leq 
&\,
 \left[
| f(\bm x_{t+1}) - f(\bm x_t) | \big| \mathcal F_t	 
\right]
+
| \alpha_{t+1} - \alpha_t | B
\\
\leq
& \,
\alpha_t Y + \alpha_t \frac{( \alpha_t - \alpha_{t+1} ) }{\alpha_t} B \notag 
\\
\leq 
&\,
\alpha_t ( Y + \kappa/2 )  \, .
\end{align*}
Taking $Z=Y+\kappa/2$, it is clear that condition \eqref{eq:con2} holds for $Z$ as an immediate consequence of the boundedness condition on $Y$ in \eqref{fcond:2}.
\Endproof

%

\LemmaB*

\Beginproof{ of Lemma \ref{LemmaB}}
It is straight-forward to show that for $a,a^{\prime},A,A^{\prime} > 0$
\begin{equation}\label{eq:lemmaB1}
    \frac{a}{a^{\prime}} \leq \frac{A}{A^{\prime}}\quad \text{if and only if}  \quad \frac{a + A}{a^{\prime} + A^{\prime}} \leq \frac{A}{A^{\prime}}\, .
\end{equation}
[Note that both expression above are equivalent to $AA^{\prime} + aA^{\prime} \leq AA^{\prime} + a^{\prime}A$.]

Take positive numbers $a_s, a_s^{\prime},\ s = 1,...,t$. If 
\begin{equation*}
\frac{a_s}{a_s^{\prime}} \leq \frac{a_k}{a_k^{\prime}}
\end{equation*}
for $k = s+1,...,t$, then
\begin{equation}\label{eq:lemmaB2}
   \sum_{k=s+1}^{t}a_{k}^{\prime}a_s \leq \sum_{k=s+1}^{t}a_ka_s^{\prime}.
\end{equation}
Thus,
\begin{equation*}
    \frac{a_s}{a_s^{\prime}} \leq \frac{\sum_{k=s+1}^{t}a_k}{\sum_{k=s+1}^{t}a_k^{\prime}}.
\end{equation*}
Thus applying \eqref{eq:lemmaB1} with $A = \sum_{k=s+1}^{t}a_k$ and $A^{\prime} = \sum_{k=s+1}^{t}a_k^{\prime}$ gives
\begin{equation}\label{eq:lemmaB3}
\frac{\sum_{k=s}^{t}a_k}{\sum_{k=s}^{t}a_k^{\prime}} \leq \frac{\sum_{k=s+1}^{t}a_k}{\sum_{k=s+1}^{t}a_k^{\prime}}. 
\end{equation}
Finally, taking $a_k = \alpha_k$ and $a_k^{\prime} = \alpha_k^2$, we see that \eqref{eq:lemmaB2} holds since $\alpha_t$ is decreasing. Thus, from \eqref{eq:lemmaB3}, we see that the result \eqref{eq:SumMin} holds.

If the condition \eqref{alphacond} holds then $\liminf_{t\rightarrow\infty} \alpha_{2t}/\alpha_t >0$ implies 
\begin{equation}\label{eq:Gcond}
\frac{\alpha_{t}}{\alpha_{\lfloor t/2 \rfloor} } > \sqrt{G}
\end{equation}
for some $1\geq G>0$. Thus
\begin{equation}\label{eq:Gcond2}
\frac{\alpha_{t}}{\alpha_{\lfloor t/2^n \rfloor} }
=
\frac{\alpha_{t}}{\alpha_{\lfloor t/2 \rfloor} } 
\times ... \times
\frac{\alpha_{\lfloor t/2^{n-1} \rfloor} }{\alpha_{\lfloor t/2^n \rfloor} } \geq G^{\frac{n}{2}} \geq G^n  \, . 
\end{equation}

Since the sequence is decreasing and \eqref{eq:Gcond} holds, we have that
\begin{equation*}
\frac{\sum_{k=\lfloor t/2^n\rfloor }^{t}\alpha_k}{\sum_{k=\lfloor t/2^n \rfloor }^{t}\alpha_k^2}
\geq 
\frac{
(t - \lfloor t/2^n \rfloor) \alpha_t
}{
(t - \lfloor t/2^n \rfloor) \alpha_{\lfloor t/2^n \rfloor}^2
}
=
\frac{\alpha^2_{t}}{\alpha^2_{\lfloor t/2^n \rfloor} } \frac{1}{\alpha_t}
=
\frac{\alpha^2_{t}}{\alpha^2_{\lfloor t/2 \rfloor} } 
\times ... \times
\frac{\alpha^2_{\lfloor t/2^{n-1} \rfloor} }{\alpha^2_{\lfloor t/2^n \rfloor} } 
\frac{1}{\alpha_t}
\geq \frac{G^n}{\alpha_t} \, .
\end{equation*}
Applying this to \eqref{eq:SumMin} with $s = \lfloor t/2^n \rfloor $ gives 
\begin{equation*}
    \min_{s= \lfloor t/2^n \rfloor ,...,t}\left\{\frac{\sum_{k=s}^{t}\alpha_k}{\sum_{k=s}^{t}\alpha_k^2}\right\} \geq \frac{G^n}{\alpha_t}\, .
\end{equation*}
Thus the above along with \eqref{eq:Gcond2} proves that \eqref{eq:SumMin2} holds as required.
\Endproof

\begin{darkblue}
\begin{lemma}\label{lem:equationcrunching}
For $\alpha_t= a/(u+t)^\gamma $ with $0\leq \gamma < 1$ Taking 
\[
n=
\begin{cases}
    1, & \text{for } \gamma <1 , \\
    1 + \left\lceil \frac{\alpha_0 B +F}{a \log 2} \right\rceil, & \text{for }
 \gamma = 1,
\end{cases}
\qquad \text{and}
\qquad T_1 =
\begin{cases}
    u+\frac{2^{1+\gamma}}{au^{-\gamma}} [\alpha_0 B+F] \, , & \text{for } \gamma  < 1 , \\
    u2^n , &\text{for } \gamma = 1 , 
\end{cases}
\]
it holds that 
\begin{equation} \label{eq:alphaB}
\sum_{s=\lfloor t/2^n\rfloor}^t \alpha_s \geq \alpha_0 B +F\, , \qquad \forall t\geq T_1 \, .
\end{equation}
\end{lemma}  
\Beginproof{}
We consider the case of $\gamma <1$ separately from the case where $\gamma = 1$.

First we take $\gamma <1$ and $n=1$.
In the following expression, we take $t=xu$ with $x\geq 1$,
\begin{equation}\label{eq:lem:longeq}
    \sum_{s=\lfloor t/2 \rfloor}^t \alpha_s 
    \geq 
    \frac{t}{2} \alpha_t  = \frac{a}{2} \frac{t}{(u+t)^\gamma} 
    = \frac{a}{2} u^{1-\gamma} \frac{x}{(1+x)^\gamma} \geq \frac{a u^{1-\gamma}}{2^{1+\gamma}} x = a \frac{u^{-\gamma}}{2^{1+\gamma}} t
\end{equation}
Thus, $t=ux$ with $x\geq 1$ and right-hand side of \eqref{eq:lem:longeq} is greater than $\alpha_0 B+F$ for 
\[
T_1 = \frac{2^{1+\gamma}}{au^{-\gamma}} [\alpha_0 B+F] +u \, ,
\]
and for any $t$ such that $t\geq T_1$. This completes the proof for $\gamma <1$ 

Second, we take $\gamma =1$. We assume that $t\geq T_1:= 2^n u$ and we will take $n= 1 + \left\lceil \frac{\alpha_0 B +F}{a \log 2}  \right\rceil $.
\[
\sum_{s=\lfloor t/2^n\rfloor}^t \alpha_s \geq \int_{t/2^n}^t \frac{a}{u+s} ds = a \log \left( \frac{u+t}{u+t 2^{-n}} \right) = an \log 2 + a \log \left( \frac{u+t}{u2^n +tk} \right)
\geq an \log 2 + a \log \frac{1}{2} \, .
\]
The last inequality above holds since $t\geq T_1:= 2^n u$.
Notice that 
\[
an \log 2 + a \log \frac{1}{2} = a(n-1) \log 2 \geq \alpha_0 B +F\, , \qquad \text{for }  n = 1 + \left\lceil \frac{\alpha_0 B +F}{a \log 2}  \right\rceil \, .
\]
Thus the required bound \eqref{eq:alphaB} holds for $n$ and $T_1$ as specified for $\gamma =1$. 
\Endproof
\end{darkblue}

\section{Appendix to Applications and Numerical Examples}\label{sec:APPLP}

This section aims to provide a simple application of the main results of Theorem \ref{Cor:1} and Theorem \ref{ThrmNew}. 
Given the importance of Linear Programming (LP) and Markov Decision Processes (MDP) in operations research, we briefly explore these problem settings. 
However, we emphasize that linear objectives are a special case of the results proven in Theorem \ref{Cor:1} and Theorem \ref{ThrmNew}. The results are proved under conditions that apply to non-smooth, non-convex objectives and general convex constraints. 
We refer to 
\cite{birge2011introduction} and \cite{shapiro2021lectures} as standard texts on stochastic linear programming. For the linear programming formulation of MDPs, we refer to \cite{schweitzer1985generalized}.

\subsection{Linear Programming}
Here we consider a linear program in which the cost function that we wish to minimize must be sampled and where the optimization constraints are deterministic. 
We are interested in solving a linear program of the form 
\begin{align}\label{LP:LP}
& \textrm{minimize}\;\; \bar{\bm{c}}^\top \bm{x} \;\;\ \textrm{subject to} \;\; H \bm{x} \leq \bm{b} \;\; \textrm{over} \;\; \bm{x} \in \mathbb R^d \, ,
\end{align}
where $\bar{\bm c} \in \mathbb R^d \backslash \{\bm 0 \}$, $H \in \mathbb R^{p\times d}$ and $\bm{b} \in \mathbb R^p$. We assume $\mathcal X = \{ \bm{x} \in \mathbb R^d : H \bm{x} \leq \bm{b}\}$ 
is a bounded polytope.

We suppose that the constraint set $\mathcal X$ is deterministic and known, however, the cost vector $\bar{\bm{c}}$ is unknown but can be sampled. 

Specifically,  we let $\bm c_t$, $t\in \mathbb Z_+$, be an independent, mean $\bar{\bm{c}}$, sub-exponential random vectors in $\mathbb R^d$. That is 
\begin{align}\label{LP:cond}
	\mathbb E [ \bm c_t | \mathcal F_t ] = \bar{\bm{c}}  
	\qquad 
	\text{and}
	\qquad 
	\sup_{t\in \mathbb Z_+}\mathbb E \left[ e^{ \lambda \|\bm c_t \|} \big| \mathcal F_t 
\right]  < \infty 
\end{align}
for some  $\lambda>0$. We then apply projected stochastic gradient descent \eqref{eq:PSGD}.
Notice that Condition \eqref{cond:D1} is satisfied by \eqref{LP:cond}. Further Condition \eqref{cond:D2} holds for any linear program. This is a consequence of the following technical lemma.

\begin{restatable}{lemma}{LemmaA}
	\label{LemmaA}
If $\mathcal{X}$ is a bounded polytope and 
$
    \mathcal{X}^\star = \text{argmin}_{\bm{x}\in\mathcal{X}} \bar{\bm{c}}^\top \bm{x}.
$, 
then there exists a constant $K>0$ such that
\begin{equation*}
    {\frac{\bar{\bm{c}}^\top(\bm{x}-\bm{x}^\star)}{||\bar{\bm{c}}||||\bm{x}-\bm{x}^\star||}} \geq K,
\end{equation*}
for $\bm{x}^\star$ the projection of $\bm{x}$ onto $\mathcal{X}^\star$. Thus Condition \eqref{cond:D2} holds for PSGD applied to the LP \eqref{LP:LP}.
\end{restatable}


The proof of Lemma \ref{LemmaA} requires some careful bounding between optimal solutions and sub-optimal extreme points. The proof is given below. The result bounds the angle between optimal and sub-optimal points for a polytope. 

\Beginproof{ of Lemma \ref{LemmaA}}
We assume without loss of generality that $\bar{\bm{c}}^\top\bm{x}^\star = 0$ and $||\bar{\bm{c}}||=1$. Let $\mathcal E$ be the extreme points of $\mathcal{X}$. Let $\mathcal E^\star$ be the extreme points in $\mathcal{X}^\star$. Then let $\mathcal E^{\prime} := \mathcal E\setminus \mathcal E^\star$ and $\mathcal X^{\prime}$ is the convex closure of $\mathcal E^{\prime}$. Let $a  := \min_{\bm{x}\in \mathcal X^{\prime}} \bar{\bm{c}}^\top \bm x$ and $D := \max_{\bm{x}^\star\in \mathcal{X}^\star, \bm{x}^{\prime}\in \mathcal{X}^{\prime}}||\bm{x}^\star-\bm{x}^{\prime}||.$
We will show we can take $K := {a}/{D}$. 

For all $\bm{x} \in \mathcal{X}\setminus\mathcal{X}^\star$, $\bm{x}$ must be a convex combination of a point in $\mathcal{X}^\star$ and a point in $\mathcal{X}^{\prime}.$ Specifically, 
\begin{equation}\label{LP:IneqPoly}
\bm{x} = (1-p)\bm{x}_0 + p\bm{x}_1, 
\end{equation}
for $\bm{x}_0\in\mathcal{X}^\star$ and $\bm{x}_1\in\mathcal{X}^{\prime}$ and $p \in (0,1]$. Then, as required,
\begin{equation*}
    \frac{\bar{\bm{c}}^\top\bm{x}}{||\bm{x}-\bm x^\star||} \geq \frac{\bar{\bm{c}}^\top(\bm{x}-\bm{x}_0)}{||\bm{x}-\bm{x}_0||} = \frac{\bar{\bm{c}}^\top(\bm{x}_1-\bm{x}_0)}{||\bm{x}_1-\bm{x}_0||} \geq \frac{a}{D} = K > 0 \, .
\end{equation*}
The first inequality above uses the fact that $\bm x^\star$ is closest to $\bm x$. The equality applies \eqref{LP:IneqPoly}. Then finally, we apply the definitions of $a$, $D$ and $K$.
\Endproof

Thus, we see that both Theorem \ref{ThrmNew} and Theorem \ref{ThrmLinear} hold in the context of linear programming problems with an unknown objective function.

\subsubsection{Polytope Example }\label{LP}
\begin{figure}[ht]
\begin{subfigure}{.45\textwidth}
  \centering
  \includegraphics[width=0.75\linewidth]{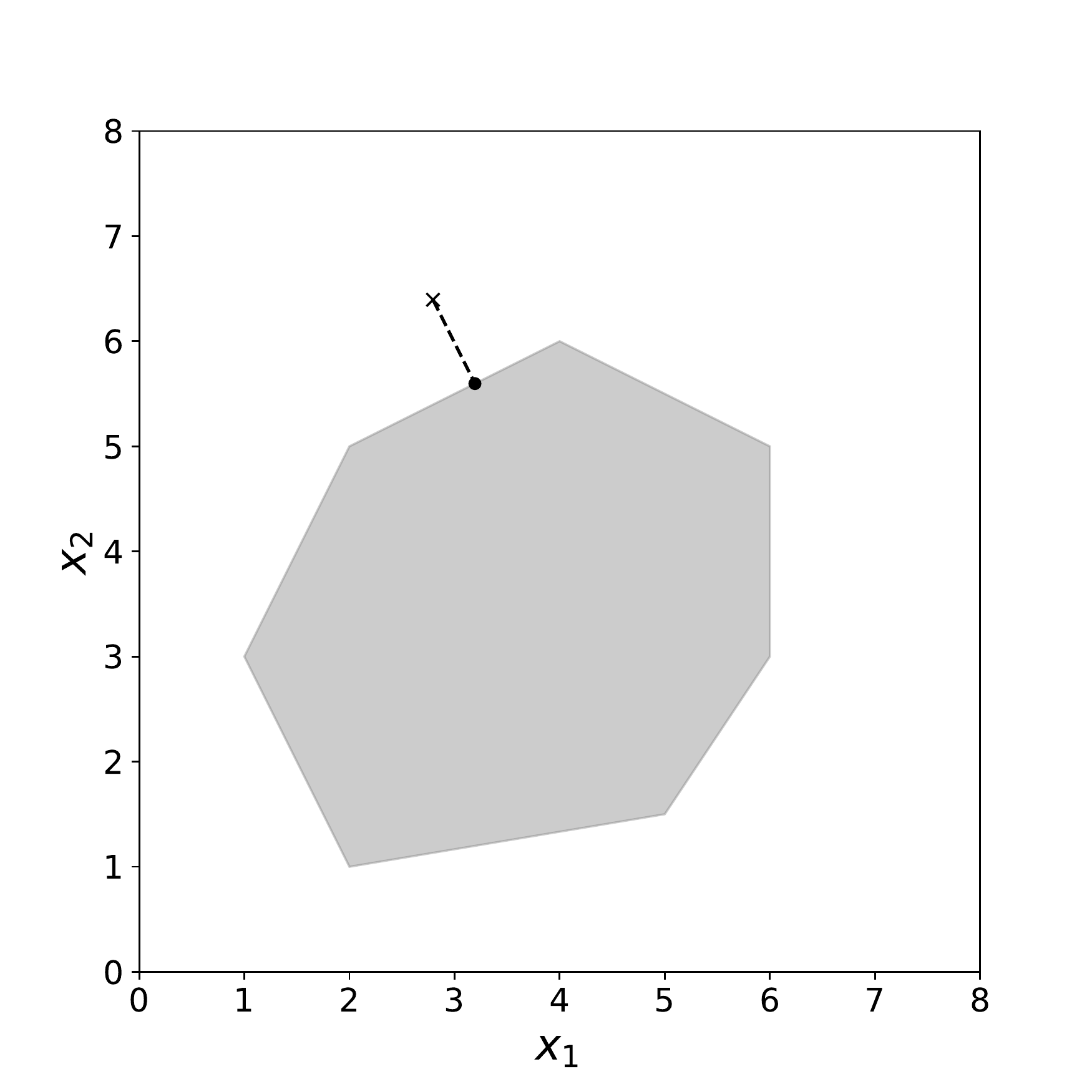}
  \caption{Polytope}
  \label{fig:LPpolytope}
\end{subfigure}%
\begin{subfigure}{.45\textwidth}
  \centering
  \includegraphics[width=1.\linewidth]{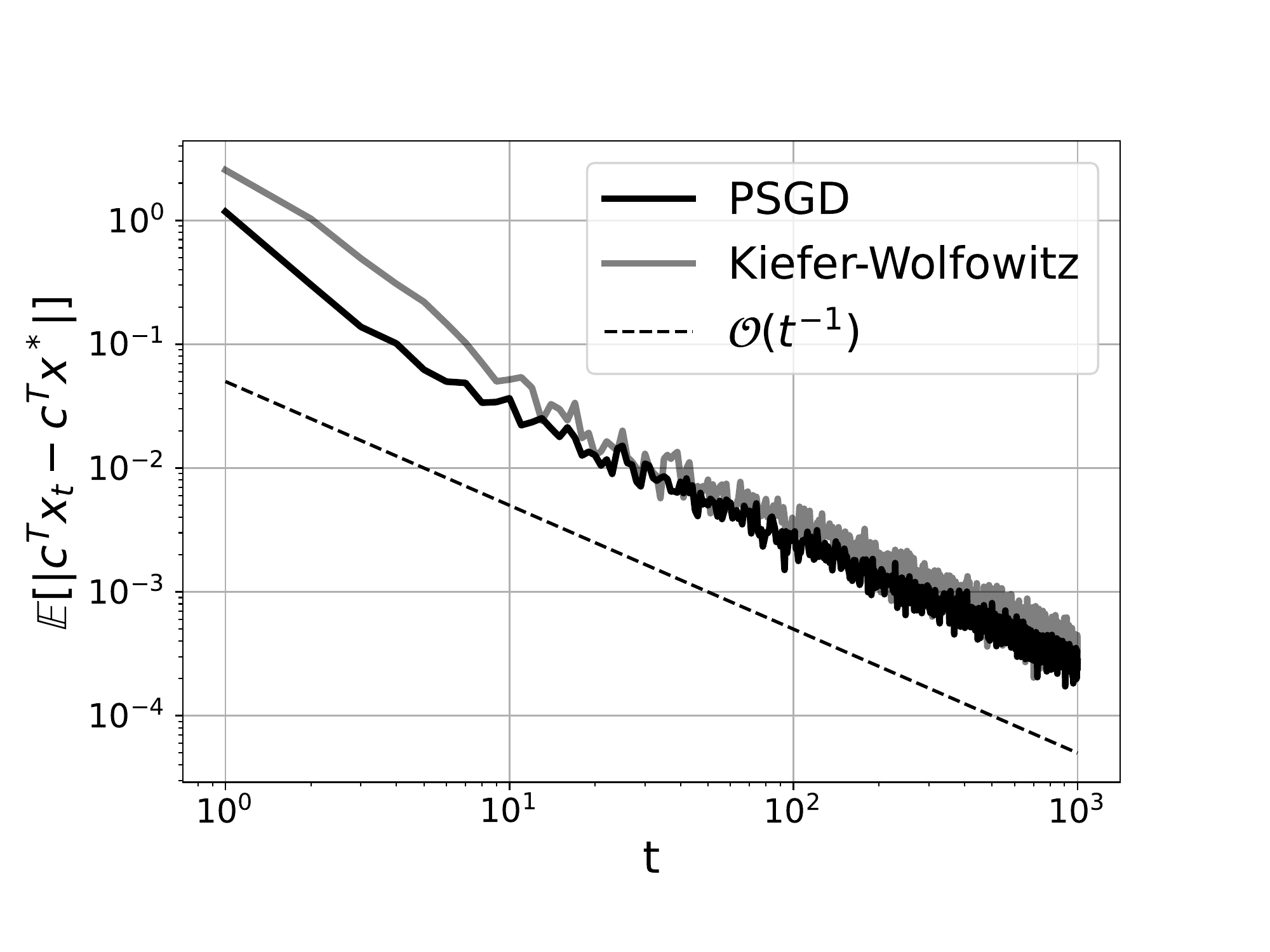}  \caption{Convergence for batch size $B=5$}
  \label{fig:LPB5}
\end{subfigure}
\
\caption{Polytope of the two variables linear programming problem and convergence of projected stochastic gradient descent on the two variables linear programming example.  In Figure \ref{fig:LPpolytope}, the shaded area is the bounded polytope and the cross is one of the points of iterations and the black point is the corresponding projection. In Figure \ref{fig:LPB5}, expectation is computed over 1000 realizations. The parameters of step size are chosen as $a=1, u=1$ and $\gamma=1$ such that $\alpha_t = 1/(1+t)$. The costs $\bm{c}_t$ are computed with batch size  $B=5$. The parameter $v=1$ is chosen for Kiefer-Wolfowitz. The fitted slope is -1.02 and -1.05 for PSGD and Kiefer-Wolfowitz. \label{fig:LP}}
\end{figure}

We consider the problem with two variables 
with the constraints being the polytope in Figure \ref{fig:LPpolytope}. 
We assume that the cost vector $\bm{\bar{c}}=[4,6]^\top$ is unknown but can be sampled from a joint Gaussian distribution of independent random variables with mean vector $\bm{\bar{c}}$ and variance $1$. This problem is analytically tractable. Given the costs, we can calculate the reference solution to be $x^\star = [2,1]^T$. 

The convergence rate for the PSGD and Kiefer-Wolfowitz should be $O(1/t^{\gamma})$ in expectation when the error is measured by the $L^1-$norm. Evidence for the convergence rate is shown in Figure \ref{fig:LPB5}. 
We note that if we increase the batch size above 50. This substantially reduces the noise of sampling the costs, and the algorithm then may perform better than $O(1/t)$. In this case, the algorithm converges reaching the optimum solution after 7 iterations. This occurs because the chance of observing any sample perturbing the stochastic gradient descent algorithm away from the optimal point is a rare event. However, when there is a non-negligible probability of an iteration leaving the optimal point then the $O(1/t)$ is found as anticipated.


\subsubsection{Probability Simplex}
\begin{figure}[ht]
\centering

\begin{subfigure}{.45\textwidth}
  \centering
  \includegraphics[width=1.0\linewidth]{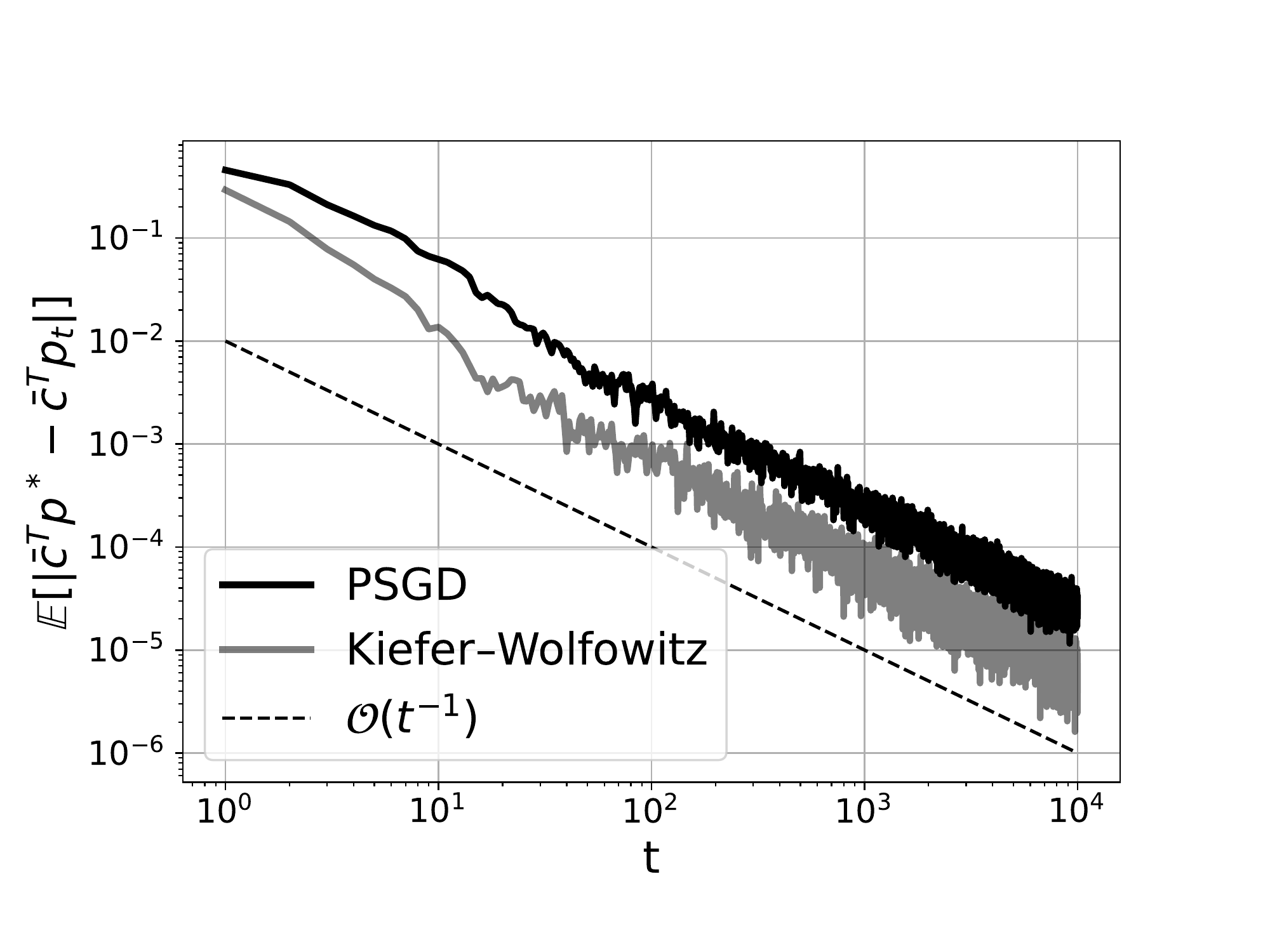}
  \caption{Probability simplex example}
  \label{fig:PS}
\end{subfigure}%
\begin{subfigure}{.45\textwidth}
  \centering
  \includegraphics[width=1.0\linewidth]{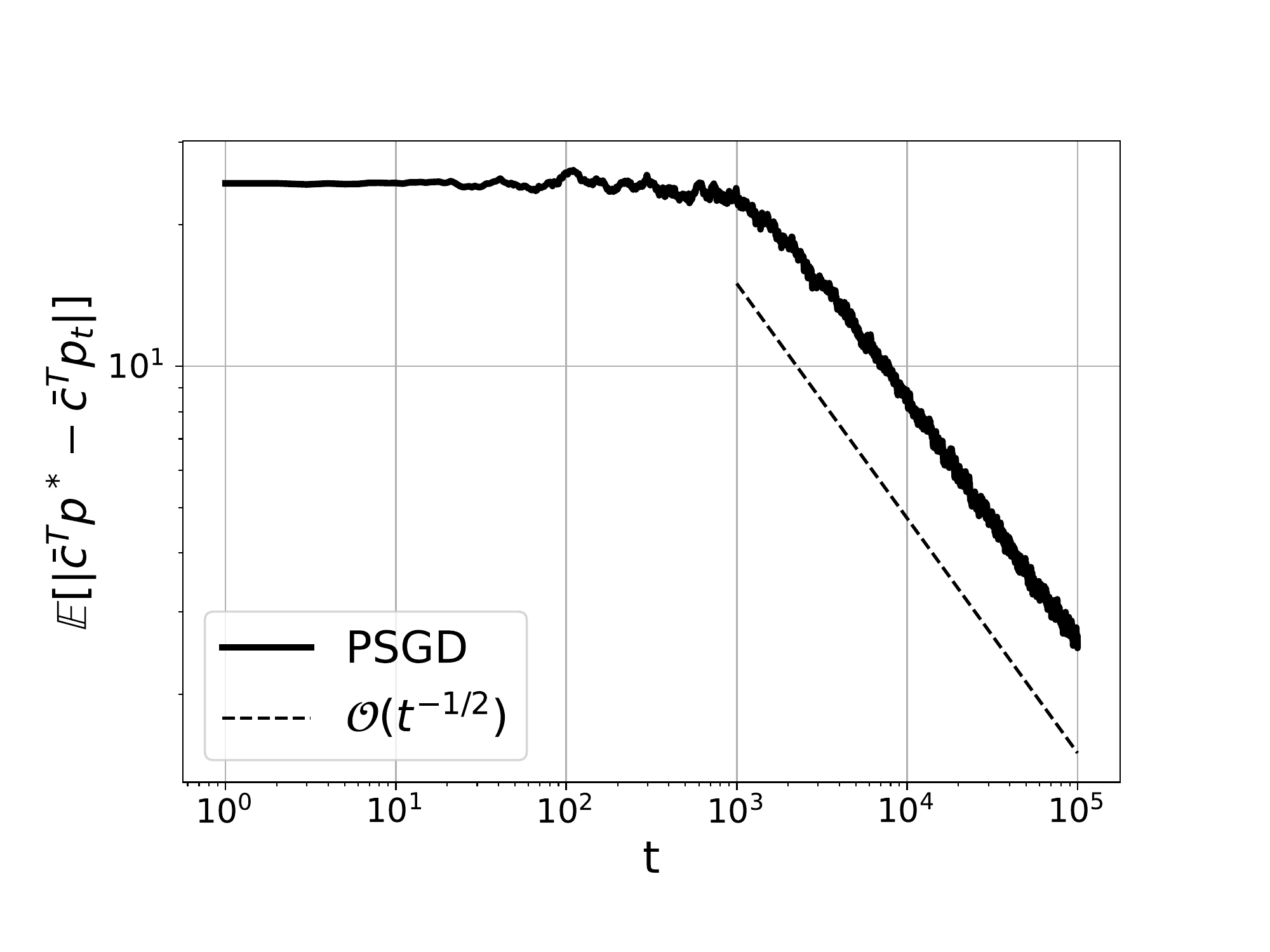}  \caption{Multi-arm bandit problem}
  \label{fig:MAB}
\end{subfigure}
\caption{Convergence of projected stochastic gradient descent on the probability simplex example and multi-arm bandit problem. Expectations are computed over 100 realizations. The parameter $v=1$ is chosen for Kiefer-Wolfowitz. The parameter of step size is chosen as $a = 1$ such that $\alpha_t=1/t$ for both. In Figure \ref{fig:PS}, the fitted slope is -1.01 and -1.01 for PSGD and Kiefer-Wolfowitz. In Figure \ref{fig:MAB}, the fitted slope is -0.475.}
\end{figure}

In this section, we consider a higher dimension for the optimization over the probability simplex as an example. There are simple efficient  algorithms for projection onto the probability simplex \citep{duchi2008efficient}. 
The problem that we solve is formulated as follows
\begin{align*}
    \text{minimize} & \quad p_1 \bar c_1 + p_2 \bar c_2 + ... + p_n \bar c_n = \bm{\bar{c}^T\bm{p}}\quad
    \text{subject to} \quad \sum_{i=1}^{n} p_i = 1\quad
    \text{over} \quad p_i \geq 0,\ \forall i=1,...,n,
\end{align*}
where $\bar c_1 < \bar c_2 < ... < \bar c_n$ and $n=50$. We label the polytope due to the constraint as $\mathcal{P}$ and suppose that the cost vector $\mathbf{\bar{c}}$ is unknown but can be sampled from a normal distribution with a certain mean vector and covariance matrix. In particular, for $t \in \mathbb{Z}_+$, we apply the stochastic gradient descent  {iteration}:
$
    \bm{p}_{t+1} = \Pi_\mathcal{P}(\bm{p}_t - \alpha_t\bm{c_t}),
$
where $\mathbf{c_t} \sim \mathcal{N}(\mathbf{\bar{c}}, \mathbf{1})$ and $\alpha_t = {a}/{t}$ with $a > 0$. According to the special settings  {above}, the minimum of this problem is $\bm p^* = (1,0,...,0)$. We expect that $\mathbb{E}\left[|\bm{\bar{c}^T\bm{p}_t} - \bm{\bar{c}^T}\bm{p}^*|\right] = O\left({1}/{t}\right).$
Figure \ref{fig:PS} confirms that the PSGD and Kiefer-Wolfowitz converge with an order of $-1$.

Although it falls somewhat outside of the scope of the results in this paper. It is also possible to consider the multi-armed bandit variation of this problem. Here the natural generalization of the projected gradient descent algorithm applies importance sampling. Here we sample an index $i_t$ according to the distribution $\bm p_t$
and apply the updated $
p_{i, t+1} = p_{i, t} - \alpha_t \frac{c_{i,t}}{p_{i, t} } \mathbb I[i=i_t]\, 
$ 
for $i=1,...,n$. 
Simulations suggest a rate of convergence of the order of $O(1/\sqrt{t})$, see Figure \ref{fig:MAB}. 



\subsection{Markov Decision Processes}\label{sec:MDP}

We now optimize a discounted Markov decision process (MDP) using the results from the last section. 
Here we use a linear programming approach to give the convergence of a simple policy gradient algorithm for an MDP where the dynamics of the system are known but the costs are unknown. 

An MDP can be formulated as a linear program, where the primal form of this linear program solves for the optimal value function, and the dual form finds the optimal occupancy measure. 
In this linear programming formulation, the dual problem takes the form: 
\begin{align}
	\text{minimize}\qquad & \sum_{s\in\mathcal S}\sum_{a\in\mathcal A}   \bar{c}(s,a) x(s,a)  \label{IDP:Dual} \tag{Dual} \\
\text{subject to} \qquad 
& \sum_{ a \in \mathcal A} x(s', a)
= \xi(s') + \beta \sum_{s\in\mathcal S} \sum_{a\in\mathcal A} x(s,a) P(s' | s,a), \qquad \forall s' \in\mathcal S\nonumber \\
\text{over} \qquad &(x(s,a):s\in\mathcal S, a\in\mathcal A)\in\mathbb R_+^{\mathcal S\times \mathcal A}.\nonumber
\end{align}
Here  $(\xi(s): s \in\mathcal S)$ is a positive vector. 
We assume that the dynamics as given by $(P(s'|s,a): a \in \mathcal A,\, s,s' \in \mathcal S)$ are known but costs   $(\bar{c}(s,a) : a  \in \mathcal A, s \in \mathcal S )$ are unknown and must be sampled, then above we have a linear program with an unknown objective and known constraints. For this reason, we can apply the analysis developed in the last section. 

Here we assume that we can sample costs $\hat{\bm c} = (\hat c(s,a) : s \in \mathcal S, a\in \mathcal A )$ where the states and actions are distributed according to some predetermined probability distribution $\bm \pi = (\pi(s,a): a \in \mathcal A, s \in \mathcal S)$. There are several ways of sampling the cost vector $\bm c_t$ for each $t$. The most straightforward one is as follows.
For each $t$, the cost ${\bm c}_t = ( c_t(s,a) : s \in \mathcal S ,\, a \in \mathcal A)$ is 
{sampled by first taking} IID sample $(s_t,a_t)$, with distribution $\bm \pi = ( \pi ( s, a) : s \in \mathcal S, a \in \mathcal A)$ where $\pi(s,a)>0$ for all $s\in \mathcal S$ and $a \in \mathcal A$, and then defining 
\begin{align}\label{eq:cost}
c_t (s,a) = \frac{ \hat{c}(s_t,a_t)}{ \pi (s_t,a_t)} \mathbb I [  (s_t,a_t) = (s,a)]  \, .
\end{align}
We allow for the possibility of averaging batches of costs of the form \eqref{eq:cost}.
We then consider the projected gradient descent {algorithm}
$
\bm x_{t+1} = \Pi_{\mathcal X} \big( \bm x_t - \alpha_t \bm c_t \big) \, .$
%
Here the projection above is onto the constraint set of the dual problem \eqref{IDP:Dual}.
Our above observation on Linear Programs holds here. Specifically Theorem \ref{ThrmNew} and Theorem \ref{ThrmLinear} hold for this PSGD algorithm.

\subsubsection{Three-state two-action Markov decision process}

\begin{figure}[ht]
\centering

\begin{subfigure}{.45\textwidth}
  \centering
  \includegraphics[width=1.0\linewidth]{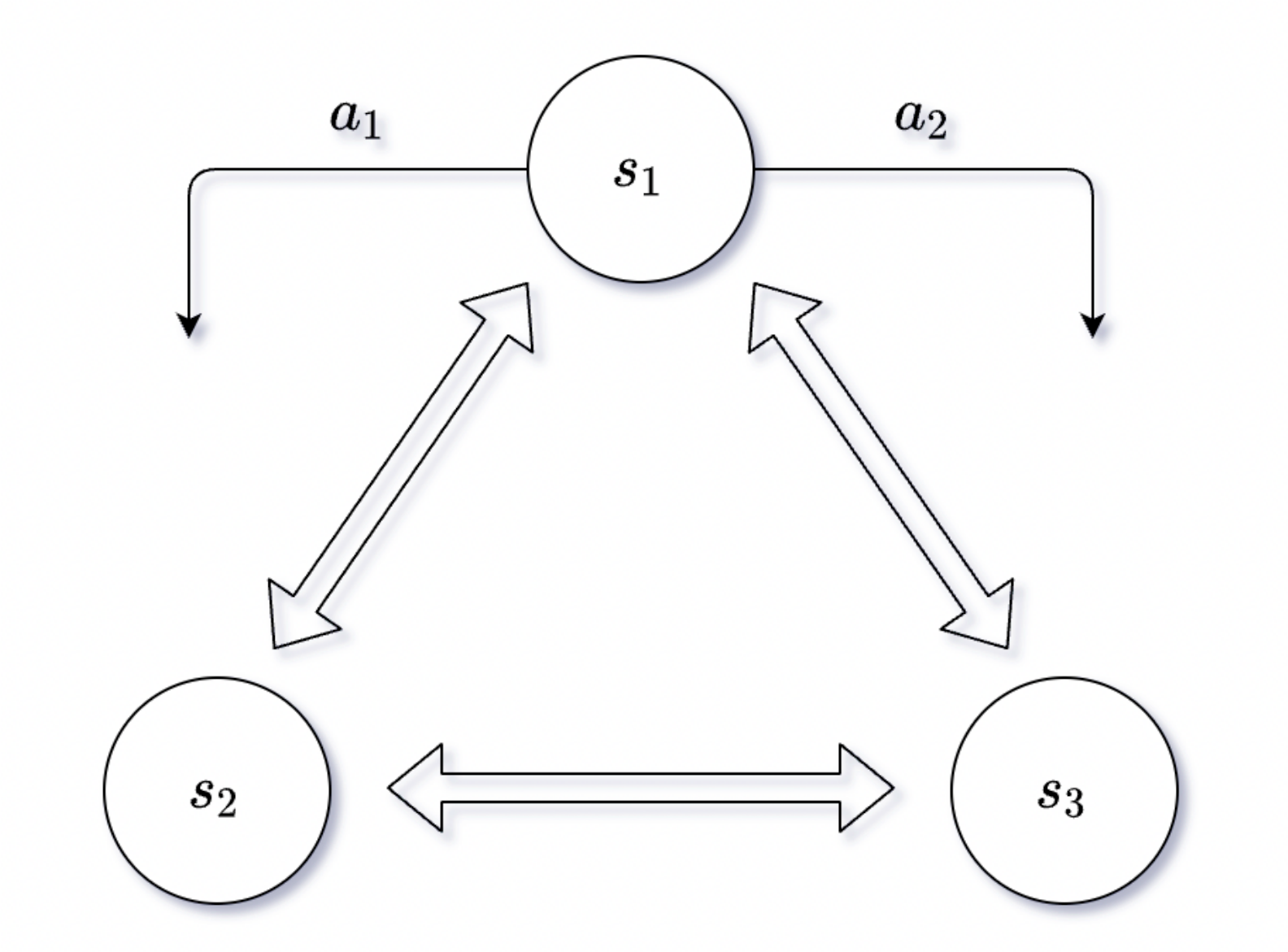}
  \caption{Three-state two-action MDP graph}
  \label{fig:3StatesMDPPlot}
\end{subfigure}%
\begin{subfigure}{.45\textwidth}
  \centering
  \includegraphics[width=1.0\linewidth]{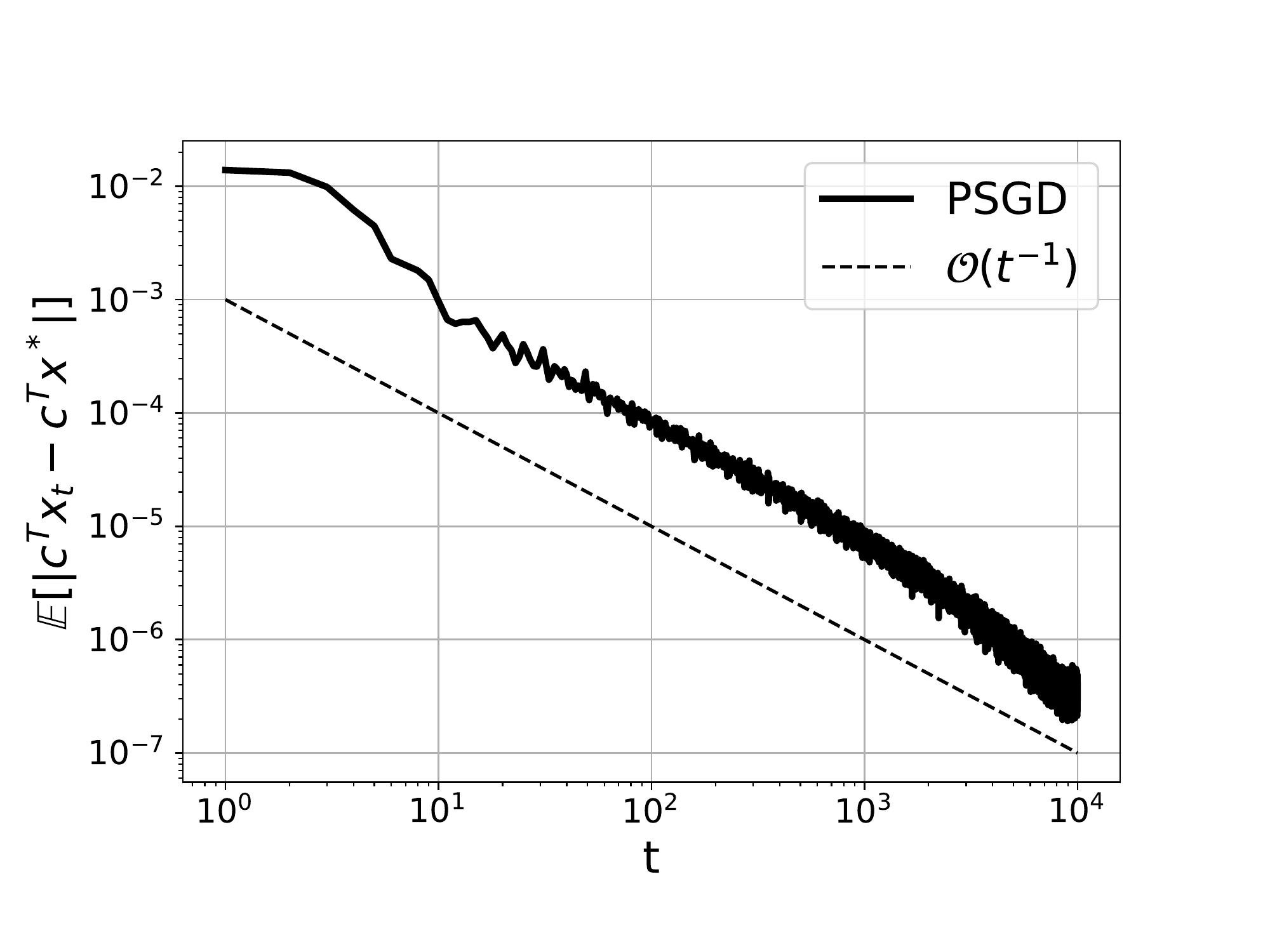} \caption{Three-state two-action MDP example}
  \label{fig:3StatesMDP}
\end{subfigure}
\caption{The three-state two-action MDP graph and convergence of projected stochastic gradient descent on the three-state two-action MDP example. In Figure \ref{fig:3StatesMDP}, the expectation is computed over 20 realizations. The costs $c_t(s,a)$ are computed with batch size $B=200$. The parameters of step size are chosen as $a=0.1, u=1$ and $\gamma=1$ such that $\alpha_t = 0.1/(1+t)$. The fitted slope is -1.29.}
\end{figure}

We now consider the first reinforcement learning application of our results. This is a relatively simple MDP which is described as follows. 
We consider a MDP with three states $\mathcal{S} = \{s_1, s_2, s_3\}$. In each state, there are two actions $\mathcal{A} = \{a_1, a_2\}$, corresponding to move anticlockwise ($a_1$) and clockwise ($a_2$). Figure \ref{fig:3StatesMDPPlot} shows the states and actions. When we choose to take an action, the probability of going to the desired state is $2/3$ otherwise one of the states  {uniformly at random}. 
We assume that the costs $c(s,a)$ are independent normally distributed with $c(s_i,a_j)\sim N(i, 1)$, for $i=1,2,3$. The states and actions are sampled according to the predetermined probability distribution $\bm{\pi} = (\pi(s,a) = 1/6 : s\in\mathcal{S}, a\in\mathcal{A})$. Figure \ref{fig:3StatesMDP} demonstrates the correct rate of convergence as predicted. 

\subsubsection{Blackjack}
We now consider a larger tabular reinforcement learning problem for the game of Blackjack. Blackjack is a simple card game where a player is initially dealt two cards. The player is dealt cards sequentially before deciding to stop. The player must attempt to reach a total that is more than the dealer but not more than 21. The
 problem is described in more detail in \cite{sutton2018reinforcement}. 
The states of the problem depend on three factors which are: the player's current points (4--22); usable ace (with or without); dealer's showing card (1--10), which gives 290 states in total.
%

We label the states in sequence starting with $s_1$ being no usable ace, the player's current points 4 and dealer's showing card 1, and ending with $s_{290}$ being usable ace, the player's current points 21 and dealer's showing card 10. The actions simply consist of hitting ($a_1$) and sticking ($a_0$). Denote the collection of states $\mathcal S = \{s_i: i=1,...,290\}$ and the collection of actions $\mathcal A = \{a_i: i=0,1\}$.

\begin{figure}[ht]
    \centering
    \includegraphics[width=.5\linewidth]{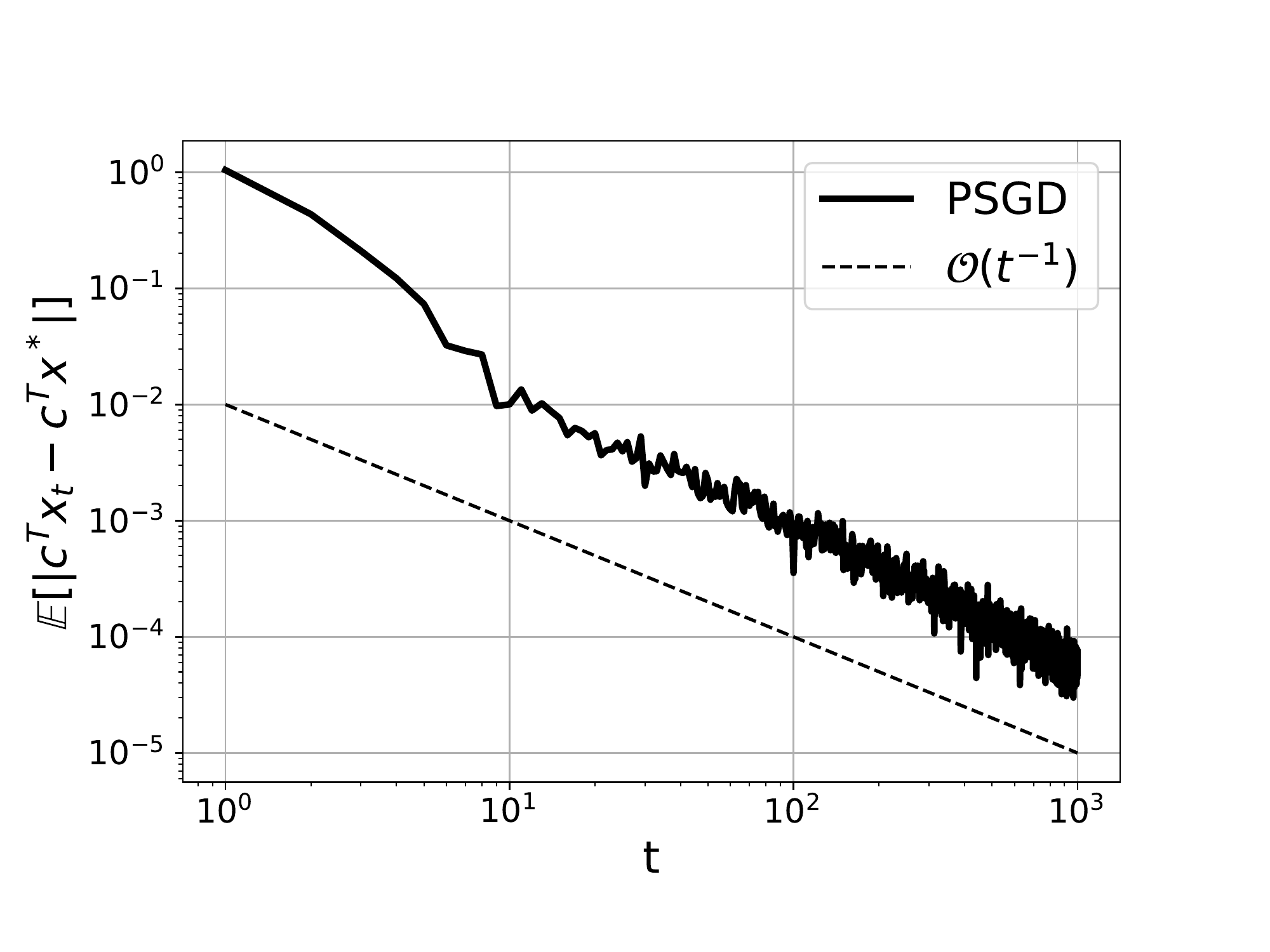}
    \caption{Convergence of projected stochastic gradient descent on the Blackjack example. The expectation is computed over 10 realizations. The costs $c_t(s,a)$ are computed with batch size $B=200$. The parameters of step size are chosen as $a=0.1, u=1$ and $\gamma=1$ such that $\alpha_t = 0.1/(1+t)$. The fitted slope is -1.23.}
    \label{fig:Blackjack}
    
\end{figure}

We assume that the reward $\bm{\bar{r}} = (\bar{r}(s,a):s\in\mathcal{S}, a\in\mathcal{A})$ can be sampled for each iteration of the projected stochastic gradient descent by carrying on the following procedure. We first simulate IID samples $(s_t^i,a_t^i),\ i=1,...,B$ from the distribution $\bm{\pi} = (\pi(s,a) = 1/580: s\in\mathcal{A}, a\in\mathcal{A})$ and then define the cost similar as Equation \eqref{eq:cost} with $\bar{c}(s_t^i,a_t^i) = -\bar{r}(s_t^i, a_t^i).$ In addition, according to the rules, it is reasonable to set the discount factor $\beta=1$.
Applying the PSGD with the learning rate of the form $\alpha_t=a/(b+t)^\gamma$, for $a,b>0$ and $\gamma\in[0,1]$, the projected stochastic gradient descent converges with a rate of $O(1/t^\gamma)$ in expectation. The rate $O(1/t)$ with $\gamma=1$ is shown in Figure \ref{fig:Blackjack}.

\end{document}